\documentclass[reqno]{amsart}

\usepackage{graphicx}
\usepackage{enumitem}
\usepackage{array}
\usepackage{arydshln}
\usepackage[foot]{amsaddr}
\usepackage{ dsfont }

\newtheorem{theorem}{Theorem}[section]

\theoremstyle{definition}
\newtheorem{definition}[theorem]{Definition}

\theoremstyle{remark}
\newtheorem{remark}[theorem]{Remark}

\newenvironment{highlight}{\begin{quote}\itshape}{\end{quote}}
\numberwithin{equation}{section}

\newenvironment{princ}[2][]{%
\refstepcounter{theorem}%
\ifstrempty{#1}%
{\mdfsetup{%
frametitle={%
\tikz[baseline=(current bounding box.east),outer sep=0pt]
\node[anchor=east,rectangle,fill=gray!30]
{\strut Visibility Principle~\arabic{section}.\arabic{theorem}};}}
}%
{\mdfsetup{%
frametitle={%
\tikz[baseline=(current bounding box.east),outer sep=0pt]
\node[anchor=east,rectangle,fill=gray!30]
{\strut Visibility Principle~\arabic{section}.\arabic{theorem}:~#1};}}%
}%
\mdfsetup{innertopmargin=10pt,linecolor=gray!20,roundcorner=2pt,
linewidth=2pt,topline=true,backgroundcolor=gray!10,
frametitleaboveskip=\dimexpr-\ht\strutbox\relax
}
\begin{mdframed}[skipabove=5pt,skipbelow=2ex]\relax%
\label{#2}}{\end{mdframed}}

\newenvironment{alg}[2][]{%
\refstepcounter{theorem}%
\ifstrempty{#1}%
{\mdfsetup{%
frametitle={%
\tikz[baseline=(current bounding box.east),outer sep=0pt]
\node[anchor=east,rectangle,fill=gray!30]
{\strut Algorithm~\arabic{section}.\arabic{theorem}: Learning the Invisible (LtI)};}}
}%
{\mdfsetup{%
frametitle={%
\tikz[baseline=(current bounding box.east),outer sep=0pt]
\node[anchor=east,rectangle,fill=gray!30]
{\strut Visibility Principle~\arabic{section}.\arabic{theorem}:~#1};}}%
}%
\mdfsetup{innertopmargin=10pt,linecolor=gray!20,roundcorner=2pt,
linewidth=2pt,topline=true,backgroundcolor=gray!10,
frametitleaboveskip=\dimexpr-\ht\strutbox\relax
}
\begin{mdframed}[skipabove=5pt,skipbelow=2ex]\relax%
\label{#2}}{\end{mdframed}}

\usepackage[usenames, dvipsnames]{xcolor}
\usepackage[framemethod=tikz]{mdframed}
\usepackage{tikz-3dplot}
\usetikzlibrary{calc,fadings,decorations.pathreplacing}
\usepackage{url}
\usepackage{amsmath}

\usepackage{subcaption}
\DeclareMathOperator{\diag}{\mathrm{diag}}
\newcommand {\mcC}{\mathcal{C}}  
\newcommand {\mcR}{\mathcal{L}}  

\newcommand{\norm}[1]{\left\lVert#1\right\rVert}
\newcommand{\abs}[1]{\left|#1\right|}
\DeclareMathOperator{\supp}{supp}
\DeclareMathOperator{\argmin}{argmin}
\DeclareMathOperator{\R}{\mathbb{R}}
\DeclareMathOperator{\N}{\mathbb{N}}
\DeclareMathOperator{\Z}{\mathbb{Z}}

\DeclareMathOperator{\WF}{WF}
\renewcommand{\vec}[1]{\boldsymbol{#1}} 
\newcommand{\indset}[1]{\mathbb{I}_{#1}} 

\newcommand{\f}{\vec{f}} 
\newcommand{\prop}{\vec{f}_{\texttt{LtI}}} 
\newcommand{\x}{\vec{x}} 
\newcommand{\xo}{\vec{x_0}} 
\newcommand{\xio}{\vec{\xi}} 

\newcommand{\meas}{\vec{y}} 
\newcommand{\noise}{\vec{\eta}} 

\DeclareMathOperator{\Radon}{\mathcal{R}}
\DeclareMathOperator{\RadonD}{\vec{\mathcal{R}}}
\DeclareMathOperator{\RadonLim}{\mathcal{R}_{\phi}}
\DeclareMathOperator{\RadonLimD}{\vec{\mathcal{R}}_{\phi}}

\newcommand{\gen}{\psi} 
\newcommand{\cshear}{\psi_{a,s,\vec{t}}}
\newcommand{\Mas}{M_{as}}
\newcommand{\SH}{\mathcal{SH}}
\DeclareMathOperator{\sh}{SH}
\DeclareMathOperator{\shD}{{\bf SH}}
\newcommand{\shear}{\psi_{j,k,\vec{m}}}

\newcommand{\NNt}{\mathcal{NN}_{\vec{\theta}}}
\newcommand{\NN}{\mathcal{NN}}

\newcommand{\tp}{\vec{t}}
\newcommand{\lp}{\vec{m}}
\newcommand{\solu}{f^*}
\newcommand{\soluD}{\vec{f}^*}
\newcommand{\inv}{\mathcal{I}_{\texttt{inv}}}
\newcommand{\vis}{\mathcal{I}_{\texttt{vis}}}

\newcommand{\ellip}{Ellipses-${50^\circ}$}
\newcommand{\mayo}{Mayo-${60^\circ}$}
\newcommand{\Mayo}{Mayo-${75^\circ}$}
\newcommand{\lotus}{Lotus-${60^\circ}$}
\newcommand{\Lotus}{Lotus-${75^\circ}$}

\newcommand{\fbp}{\vec{f}_{\texttt{FBP}}}
\newcommand{\tv}{\vec{f}_{\texttt{TV}}}
\newcommand{\unser}{\NNt(\vec{f}_{\texttt{FBP}})}
\newcommand{\jcyour}{\NNt(\shD (\vec{f}_{\texttt{FBP}}))}
\newcommand{\jcysol}{\vec{f}_{[31]}}
\newcommand{\methodname}{PhantomNet}

\newcolumntype{T}{>{\centering\arraybackslash} m{1.25cm} }
\newcolumntype{S}{>{\centering\arraybackslash} m{2.0cm} }
\newcolumntype{L}{>{\centering\arraybackslash} m{3.5cm} }
\newcolumntype{Y}{>{\centering\arraybackslash} m{4.5cm} }
\newcolumntype{X}{>{\centering\arraybackslash} m{1.5cm} }

\definecolor{amber}{rgb}{1.0, 0.75, 0.0}
\definecolor{darkblue}{rgb}{0.0, 0.0, 0.55}
\definecolor{aquamarine}{rgb}{0.4, 0.8, 0.66}
\definecolor{purple}{RGB}{186, 121, 246}

\usepackage[textwidth=2.3cm]{todonotes}
\begin{document}


\title[Learning The Invisible]{Learning The Invisible: A Hybrid Deep Learning-Shearlet Framework for Limited Angle Computed Tomography}

\author[T. A. Bubba]{Tatiana A. Bubba}
\email{tatiana.bubba@helsinki.fi}

\author[G. Kutyniok]{Gitta Kutyniok}
\email{kutyniok@math.tu-berlin.de}

\author[M. Lassas]{Matti Lassas}
\email{matti.lassas@helsinki.fi}

\author[M. M\"{a}rz]{Maximilian M\"{a}rz}
\email{maerz@math.tu-berlin.de}

\author[W. Samek]{Wojciech Samek}
\email{wojciech.samek@hhi.fraunhofer.de}

\author[S. Siltanen]{Samuli Siltanen}
\email{samuli.siltanen@helsinki.fi}

\author[V. Srinivasan]{Vignesh Srinivasan}
\email{vignesh.srinivasan@hhi.fraunhofer.de}

\address[T.~A.~Bubba, M.~Lassas, S.~Siltanen]{Department of Mathematics and Statistics, University of Helsinki, 00014 Helsinki, Finnland}
\address[G.~Kutyniok]{Department of Mathematics and Department of Electrical Engineering \& Computer Science, Technische Universit\"{a}t Berlin, 10623 Berlin, Germany}
\address[M.~M\"{a}rz]{Department of Mathematics, Technische Universit\"{a}t Berlin, 10623 Berlin, Germany}
\address[W.~Samek, V.~Srinivasan]{Department of Video Coding and Analytics, Fraunhofer Heinrich Hertz Institute, 10587 Berlin, Germany}

\date{\today}



\begin{abstract}
The high complexity of various inverse problems poses a significant
challenge to model-based reconstruction schemes, which in such situations often
reach their limits. At the same time, we witness an exceptional 
success of data-based methodologies such as deep learning.
However, in the context of inverse problems, deep neural networks mostly act as black box routines, used for instance for a somewhat unspecified removal of artifacts in classical image reconstructions.  In this paper, we will focus
on the severely ill-posed inverse problem of limited angle computed tomography, in which entire boundary sections are not captured in the measurements. We will develop 
a hybrid reconstruction framework that fuses 
model-based sparse regularization with data-driven deep learning. Our method is
\emph{reliable} in the sense that we only learn the part that can provably not be handled 
by model-based methods, while applying the theoretically controllable 
sparse regularization technique to the remaining parts.  Such a decomposition into \emph{visible} and \emph{invisible} segments is achieved by means of the shearlet transform that allows to resolve wavefront sets in the phase space. Furthermore, this split enables us to assign the clear task of inferring unknown shearlet coefficients to the neural network and thereby offering an \emph{interpretation} of its performance in the context of limited angle computed tomography.
Our numerical experiments show that our algorithm significantly surpasses both 
pure model- and more data-based reconstruction methods.
\end{abstract}

\maketitle

\section{Introduction}

Due to increased computational power and advanced mathematical understanding, there is a growing interest in solving severely ill-posed inverse problems. The goal is to recover an unknown quantity from indirect measurements, where typically only few of them are acquired and the reconstruction process is highly sensitive to modelling errors and noise. Traditional inversion methods are based on complementing the insufficient and corrupted measurement data by mathematical models, which impose {\it a priori} information on the solutions. Such methods include Tikhonov regularization, Bayesian inversion, and inversion algorithms based on partial differential equations or applied harmonic analysis. 

However, sometimes the ill-posedness renders it very difficult to robustly recover specific parts of the target. A prominent example is the inverse problem of limited angle computed tomography (limited angle CT), where the missing part of the wavefront set of the target can be read off the measurement geometry \cite{Quinto93,Frikel13a,Natterer01}. 
In some medical applications, it is enough to consider slices of the reconstruction where the stable part of the wavefront set reliably provides the clinically important boundaries of tissues. For example, in \cite{Rantala2006} the spatial position of  microcalcifications in the breast can be recovered, and the slice considered in  \cite{kolehmainen2007bayesian} provides a low-dose X-ray examination for dental implant planning. However, any new method that is able to recover the missing part of the wavefront set more reliably would improve the quality of those reconstructions and lead to unprecedented applications of limited angle tomography.

Currently, we witness a tremendous success of data-based methodologies such as deep neural 
networks for a wide range of problems, for example, speech recognition \cite{hinton2012}, the game of Go \cite{silver2016} or image classification \cite{krizhevsky2012} and many more. The underlying philosophy is agnostic in the sense that no explicit data model is specified, but vast amounts of training data are used to infer an implicit proxy. During the last years, also the area of inverse problems is increasingly impacted by machine learning approaches, in particular, by deep learning (see, e.g.,  \cite{xie2012,burger2012,kang2017,unser2017,adler2017,hauptmann2018}). However, at this time, neural networks are mostly used as black boxes that are for instance trained for an unspecific image enhancement of direct inversions, or for a replacement of iteration steps in optimization algorithms.


In this paper, we develop a framework for solving the inverse problem of limited angle 
CT by combining model-based sparse regularization using shearlets with a data-driven 
deep neural network approach. The key idea of our hybrid method goes back to Quinto's fundamental visibility analysis of limited angle CT based on \emph{microlocal analysis} \cite{Quinto93}. We utilize sparse regularization with shearlets for splitting the (wavefront set of the) data into a \emph{visible part,} recoverable by classical model-based methods, and an \emph{invisible part}, that is provably not contained in the measured data. Precisely this part is sought to be recovered by an inference in the shearlet domain by means of a trained neural network. 
Such an estimation of unknown shearlet coefficients is highly dependent on a faithful model-based reconstruction of its visible counterpart. Therefore, the focus of our work is on a moderate missing wedge, i.e., where at most as much information needs to be inferred as it is available to classical methods on the visible part.




\subsection{Shearlets and Sparse Regularization}

Given an ill-posed inverse problem $y = \Radon f + \eta$, where $\Radon : X \to Y$ with suitable spaces $X$ and $Y$, and $\eta$ models measurement noise, Tikhonov regularization provides an approximate solution $f^\lambda \in X$, $\lambda > 0$, 
by minimizing the functional
\[
J_\lambda(f) := \|\Radon f-y\|^2 + \lambda \cdot {\mathcal{P}}(f),\quad f \in X,
\]
with ${\mathcal{P}}(f)$ being a penalty term, promoting desired properties in the solution $f^\lambda$. Sparse regularization is then based on the common paradigm that for each class of data, there exists a sparsifying representation system \cite{daubechies2004,Candes2006,Donoho2006,Elad10}. In the considered situation, one would assume that there exists a
system $(\psi_\mu)_\mu \subseteq X$ such that the sparsity promoting $\ell^1$-norm of the coefficient vector $(\langle f, \psi_\mu \rangle)_\mu$ is small,
therefore allowing to choose the regularization term as
\[
{\mathcal{P}}(f) = \|(\langle f, \psi_\mu \rangle)_\mu\|_1.
\]

Let us now focus on inverse problems in imaging. In this situation, wavelet systems \cite{mallat09} are suboptimal, since it
is known that -- due to their isotropic nature -- they are not capable of providing optimally sparse
approximations of images under the well-accepted assumption that images are governed by edges, hence
anisotropic features. \emph{Shearlets} \cite{labate2005,Kutyniok2012} are representation systems specifically designed for multivariate data that are optimally adapted to such anisotropic structures. As such they can be seen as a further 
development of \emph{curvelets} \cite{Candes02}, which were the first system that allowed to provide 
optimally sparse approximations of \emph{cartoon-like images} - a mathematical abstraction of real-world 
images. Shearlets build upon the same ideas, however, they additionally offer the benefit of an unified 
treatment of the continuous and discrete situation allowing for faithful implementations \cite{Kutyniok2012}.
Shearlets have already been very successfully applied to various inverse problems, such as denoising \cite{easley2009}, CT \cite{Colonna2010}, phase retrieval \cite{loock2014} or inverse scattering \cite{kutyniok2017}.

To be a bit more precise, the elements of a {\em shearlet system} $\{\shear\}_{(j,k,\lp) \in \Z \times \Z \times \Z^2}$ 
are parametrized by a scale parameter $j$, a directional parameter $k$, and a parameter for the position $\lp$. 
In the sequel, we will denote the associated transform by $\sh_\gen (f)$, i.e., 
\[
\sh_\gen (f) = (\langle f, \shear \rangle)_{(j,k,\lp) \in \Z \times \Z \times \Z^2}
\]
(for more details see Section \ref{sec:shearlets}).
Similarly, a {\em continuous shearlet system} $\{\cshear\}_{(a,s,\tp) \in \R_+\times \R \times \R^2}$ can be defined by considering continuous indexing parameters. 
 One striking property of the continuous version of shearlets -- which will be 
crucial for our approach -- is their ability to resolve the wavefront set of generalized functions \cite{Grohs2011,kutyniok2009}. 
Roughly speaking, a wavefront set consists of the positions of the singular support of a generalized function $f$ 
together with their directions, and is a subset of the so-called {\em phase space} $\R^2 \times \mathbb{P}_1$.
Considering the decay properties of the shearlet coefficients $(\langle
f, \cshear \rangle)_{(a,s,\tp) \in \R_+\times \R \times \R^2}$ as $a \to 0$ yields precisely those 
position-direction pairs $(\tp,s)$ which constitute the wavefront set of $f$. 

\subsection{Neural Networks and Inverse Problems}

Artificial neural networks were originally introduced in 1943 by McCulloch and Pitts as an approach
to develop learning algorithms by mimicking the human brain \cite{McCulloch1943}. Their main goal at that time was
the development of a theoretical approach to artificial intelligence. However, the limited
amount of data and the lack of high performance computers prevented the training of networks
with many layers. 

By now these two obstacles are overcome and we have massive amounts of training data as well
as a tremendously increased computing power available, thereby allowing the training of
deep neural networks. This is one of the reasons why neural networks have recently seen such
a spectacular comeback with impressive performance results in applications such as 
game playing (AlphaGo), image classification, speech recognition, to name a few \cite{silver2016,hinton2012,krizhevsky2012}.
From a mathematical perspective, a deep neural network in an idealized form is a high-dimensional function $\NN : \mathbb{R}^n \to \mathbb{R}^d$ of the form
\begin{equation} \label{eq:DNN}
\NN(\vec{x}) =  W_L(\sigma(W_{L-1}(\sigma(\ldots(\sigma(W_1(\vec{x})))\ldots)))),
\end{equation}
with the $W_j$ being affine-linear functions and $\sigma : \mathbb{R} \to \mathbb{R}$ being
the (non-linear) activation function applied componentwise. 
In a nutshell, the goal of \emph{deep learning} is to approximate an (unknown) structural relation between the input space $\R^n$ and the output space $\R^d$ with $\NN$. This task is achieved by determining the affine-linear functions $W_j$ from the knowledge of training examples $(\vec{x}_i,\vec{y}_i)_{i=1}^N \subseteq \R^n \times \R^d$ following the underlying relation.


One should stress that various special cases exist, with convolutional neural networks (CNNs) \cite{Fukushima1980,LeCun89} being the most prominent architectures in the context of imaging. However, most of the related research is performance driven, while developing a mathematical foundation mostly plays a secondary role. 
Despite the lack of a complete theoretical understanding, deep learning is currently penetrating various areas of applied mathematics. This is in particular true for the area of inverse problems in imaging sciences where
sophisticated model-based approaches, which used to be the previous state of the art, are now outperformed. 

Some approaches train a deep neural network directly for the inversion from noisy measurements $\vec{y} = \RadonD \vec{f} + \vec{\eta}$, based on a collection of training samples $(\vec{y}_i,\vec{f}_i)_{i=1}^N$ following the considered forward model, e.g.,~\cite{Paschalis2004} for CT and \cite{xie2012} for denoising and inpainting\footnote{In order to highlight the difference between continuous objects and their discretizations, we will use boldface letters to denote matrices and finite-dimensional vectors in $\R^n$ ($n\geq 2$).}.  
Other typical works aim at explicitly incorporating knowledge about the forward model $\RadonD$ into the reconstruction process. This is for instance achieved  by preprocessing the measurements $\vec{y}_i$ with a model-based inversion, e.g.,~\cite{unser2017,kang2017,pelt2013} for CT, or \cite{wang2016} in the case of magnetic resonance imaging.
Finally, recent approaches insert deep
networks into iterative reconstruction schemes, for instance by unrolling the steps and casting them as a network \cite{gregor2010learning}, or by replacing some of the proximal operators by a CNN \cite{adler2018,Yang2016,meinhardt17learning}. Still, all these methodologies have in common that the entire reconstructed image has undergone transformations by one or more neural networks, during which control over the applied modifications might have been lost. We refer the interested reader to the reviews \cite{adler2017,McCann2017} for a more detailed discussion on the use of deep neural networks in the context of inverse problems.

\subsection{A Bit of History: Limited Angle Computed Tomography}

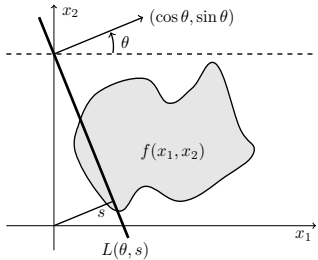
\begin{figure}
\centering
\resizebox{4.3cm}{3.5cm}{
\begin{tikzpicture}
\draw[->] (-2.5,-2.0) -- (-2.5,3.0);
\draw[->] (-3.5,-1.5) -- (3.0,-1.5);
\draw[dashed,thick] (-3.5,2.0) -- (3.0,2.0);
\draw[thick,->] (-2.5,2) -- (-0.625,2.75);
\draw[ultra thick] (-2.8125,2.75) -- (-0.9375,-1.75);
\draw[thick] (-2.5,-1.5) -- (-1.25,-1.0);
\draw[fill=black, fill opacity=0.1,thick]  plot[smooth, tension=.7] coordinates {(-2.0,0.5) (-1.6,1.2) (-0.6,1.5) (-0.1,1) (0.4,1.3) (0.9,1.8) (1.4,0.5) (1.7, 0) (1.1,-0.5) (0.2,-1) (-0.6,-0.7) (-1.1,-1.2) (-1.6,-0.8) (-2.05, 0) (-2.0,0.5)};
\draw[->,domain=-10:23,thick] plot ({-2+0.75*cos(\x)}, {2.15+0.75*sin(\x)});
\draw (0,0) node{$f(x_1,x_2)$};
\draw (2.75,-1.7) node{$x_1$};
\draw (-2.15,2.85) node{$x_2$};
\draw (-1.5,-1.25) node{$s$};
\draw (-1.0,2.25) node{$\theta$};
\draw (-1.0,-2.0) node{$L(\theta,s)$};
\draw (0.4,2.75) node{$(\cos\theta, \sin\theta)$};
\end{tikzpicture}
}
     \caption{Geometric setup of the Radon transform defined in Equation \eqref{eq:Radon}.}
     \label{fig:meas}
\end{figure}

Limited angle tomography appears frequently in practical applications, such as dental tomography \cite{Kolehmainen03}, damage detection in concrete structures \cite{heiskanen1991}, breast tomosynthesis \cite{Zhang06}, or electron tomography \cite{Baumeister99}.  Given the original image $f \in L^1(\R^2)$, a simplified, mathematical model of the general tomographic data acquisition process is given by the \emph{Radon transform}
\begin{equation}
    \label{eq:Radon}
    \Radon f (\theta,s) = \int_{L (\theta, s)} f(\x) dS(\x),
\end{equation}
where $\theta \in [-\pi/2,\pi/2)$, $s \in \R$ and
\[
L (\theta, s) := \left\{ \x \in \R^2 :  x_1 \cos (\theta) + x_2 \sin (\theta) = s \right\}
\]
denotes the line with normal direction $\theta$ and distance $s$ to the origin, with $dS$ being the 
1-dimensional Lebesgue measure along it; e.g.,~\cite{Epstein08,Natterer01}. The underlying geometric setup is sketched in Figure \ref{fig:meas}. 

Due to physical constraints on the measurement device, the Radon transform $\Radon f$ 
is often not known on the entire angular range $\theta \in [-\pi/2,\pi/2)$, but only on a subinterval  
$[-\phi,\phi]$ with $\phi < \pi/2$. We will indicate such a \emph{missing wedge} by the notation
\[
\RadonLim f := \Radon f_{\big|[-\phi,\phi]\times\R}.
\]
The task of \emph{limited angle CT} is to recover an approximation of $f$ from its noisy measurements 
\begin{equation}
    \label{eq:meas}
    y = \RadonLim f + \eta,
\end{equation}
where $\eta$ models deterministic and/or random measurement errors. There exists an abundance of inversion strategies for (limited angle) CT and we will now briefly review some of the most relevant ones for our work. 

Due to the limited angular range, not all features of the measured object $f$ are captured under $\RadonLim$ \cite{Quinto93} and the resulting inverse problem is severely ill-posed \cite{Davidson83}. 
Therefore, classical methods, in particular the filtered backprojection (FBP), are known to yield suboptimal performance, although still being very popular in applications, mostly due to computational performance.

In the case of low-dose CT, it has been demonstrated that sparse regularization methods allow for 
accurate reconstructions from fewer tomographic measurements than usually required by standard methods such as the FBP \cite{Sidky2008,Jorgensen2017,Ritschl2011,Sidky2006,Jorgensen2015}. 
Often total variation (TV), which enforces gradient sparsity, is used as a simple but very 
effective prior, but also wavelets \cite{Loris2006,Rantala2006,Klann2015}, curvelets \cite{Candes2000,Frikel2013b} 
and shearlets \cite{Colonna2010,Bubba2018,Guo2013}  have been successfully applied. However, to the best 
of our knowledge, shearlets have not yet been considered for $\ell^1$-regularization in 
limited angle CT. 
Although advanced sparsity-based variational schemes define effective regularization methods, the amount of missing data in limited angle CT is typically so severe that certain features remain impossible to reconstruct and streaking artifacts appear \cite{Frikel2013b}.

There already exist a few approaches to exploit deep learning for solving the limited angle CT problem. All of the three following  methods have in common that they are essentially based on a direct inversion, followed by a ``denoising'' procedure - as such, one of the most straightforward ways to tackle an inverse problem. In \cite{zhang2016}, a shallow convolutional network is trained to remove artifacts in FBP reconstructions. Additionally to the postprocessing with a variational network, \cite{hammernik2017} makes use of a second neural network for correcting inhomogeneities in the projection domain. Intriguing results are achieved in the work of Gu and Ye \cite{jcy}: similar as in \cite{unser2017}, a so-called U-Net CNN \cite{ronneberger2015u} is trained to improve the FBP reconstruction. However, based on the insight that the artifacts in limited angle tomography posses a directional nature, the CNN processes the directional wavelet coefficients of the FBP image. While all of these three methods yield impressive results given the substantial amount of missing projections, potential drawbacks can be summarized as follows:

\enlargethispage{0.4cm}
\begin{itemize}

\item The remarkable post-processing capabilities of neural networks come with a flavor of \emph{alchemy}: a somewhat unspecified removal of artifacts in the FBP reconstructions can be observed. However, it remains unclear to what extent the resulting image has been modified by a CNN and therefore how reliable the reconstructions are. We regard this as particularly critical for medical applications.  

\item While post-processing FBP data is computationally attractive, it might not be an optimal choice in terms of reconstruction quality, since the FBP solution is heavily contaminated with artifacts and potentially blurry - a flaw that might be amplified by post-processing with a U-Net architecture \cite{han2018}. 

\item The advantage of regularizing with an anisotropic system, which allows for an extraction of visibility information, is not fully exploited. 

\end{itemize}

\subsection{Our Contribution}

The main objective of our approach is to design a reconstruction framework for limited angle CT, where deep learning is solely applied to those parts of the inverse problem that  are provably not contained in the measured data. This will ensure a maximal amount of reliability and interpretability of our results. Interestingly, such an hybrid approach clearly outperforms previous data-based methods. 

Let us now foremost focus on edge information, which in the distributional situation refers to the wavefront set of an image $f$. The fundamental visibility analysis for limited angle CT by Quinto \cite{Quinto93} allows to distinguish which singularities can be accurately reconstructed and which are not contained in the measurements. Put simply, the dividing
criteria is whether an edge is tangent to an acquired line $L (\theta, s)$ or not (see also Theorem \ref{thm:quinto} and Visibility Principle \ref{pr:1}). Singularities belonging to the first type are referred to as {\em visible} while the others are {\em invisible}. 
We can conclude
that some parts of the wavefront set can be robustly recovered -- the {\em visible part} -- and
some parts not -- the {\em invisible part}. Thus, a complete recovery of all singularities of $f$ could be regarded as an inpainting problem on its wavefront set. Our original intention for this work was to \emph{handcraft} a variational prior that promotes the completion of the gaps during the reconstruction. Although such rules are quite intuitive, their mathematical formalization turned out to be surprisingly difficult. Consequently, the goal of our work is to estimate the invisible part by applying a deep neural network that is specifically trained for this task. Such an inference of invisible information from the knowledge of its visible counterpart is feasible, since the wavefront sets of typical images follows similar structural patterns in the phase space. 

As mentioned before, we intend to apply advanced model-based methods for a recovery of the reliable boundary information. Thus, the first step of our algorithm consists in solving a sparse regularization problem, which is conceptually of the following form (see Section \ref{sec:l1Ana} and Algorithm \ref{algo} for more details):

{\noindent\bf{Step 1  - Recover the Visible:}}
\begin{equation*}
  \soluD := \argmin_{\f} \frac{1}{2}\norm{\RadonLimD \f - \meas}_2^2 + \lambda \cdot \norm{\shD_{\gen} (\f)}_{1}
\end{equation*}
Thereby, $\RadonLimD$ and $\shD_{\gen}$ denote finite dimensional approximations of their continuous version $\Radon$ and $\sh_{\gen}$. Promoting sparsity with the $\ell^1$-norm in the shearlet domain allows to characterize the visible parts of the boundaries. Indeed, similar as in \cite{Frikel2013b}, we observe that there exists a partition of the shearlet parameter set $\Lambda = \vis \cup \inv$ approximately satisfying the following:
\begin{itemize}
 \item for $(j,k,\lp) \in \inv$: $\shD_\gen (\soluD)_{(j,k,\lp)} \approx 0$,
 \item for $(j,k,\lp) \in \vis$: $\shD_\gen (\soluD)_{(j,k,\lp)} \approx \shD_\gen (\f)_{(j,k,\lp)}$.
\end{itemize}
The shearlet coefficient tensor $\shD_\gen (\soluD)$ resembles the previously discussed phase space, in which the missing wedge of limited angle CT causes gaps in the wavefront set. Inpainting them can be rephrased as an estimation of the shearlet coefficients associated with $\inv$ - a task that shall be accomplished by an artificial deep neural network.   


The network is trained to generate an estimation of the invisible shearlet coefficients $\shD_\gen (\f)_{\inv}$, when the coefficient tensor $\shD_\gen (\soluD)$ is given as input. 

{\noindent\bf{Step 2  - Learn the Invisible (LtI):}}

\begin{table}[!htbp]
\centering
\begin{tabular}{@{}L@{\!\!\!}T@{\quad}T@{\quad}T@{\quad}S@{}}
$\NN:\; \shD_\gen (\soluD)$
& \begin{tikzpicture} \draw[->,thick] (0,0) -- (1.25,0); \end{tikzpicture} 
& \begin{tikzpicture}[scale=0.18]
\draw[gray,thick] (0.42,-0.3) -- (1.5,-0.75);
\draw[gray,thick] (0.5,0.1) -- (1.5,0.65);
\draw[gray,thick] (0.45,0.25) -- (1.5,2.15);
\draw[gray,thick] (0.35,0.4) -- (1.75,3.3);
\draw[gray,thick] (0.42,1.2) -- (1.55,-0.5);
\draw[gray,thick] (0.5,1.6) -- (1.5,0.85);
\draw[gray,thick] (0.45,1.75) -- (1.55,2.5);
\draw[gray,thick] (0.35,1.9) -- (1.6,3.45);
\draw[gray,thick] (0.42,2.7) -- (1.65,-0.35);
\draw[gray,thick] (0.5,2.9) -- (1.62,1.1);
\draw[gray,thick] (0.45,3.25) -- (1.65,2.65);
\draw[gray,thick] (0.35,3.4) -- (1.5,3.6);

\draw[gray,thick] (2.45,-0.5) -- (3.57,0.45);
\draw[gray,thick] (2.25,-0.3) -- (3.73,1.8);
\draw[gray,thick] (2.25,0.3) -- (3.49,0.65);
\draw[gray,thick] (2.45,1) -- (3.55,2);
\draw[gray,thick] (2.25,1.8) -- (3.5,0.85);
\draw[gray,thick] (2.45,2.5) -- (3.47,2.25);
\draw[gray,thick] (2.25,3.3) -- (3.55,1);
\draw[gray,thick] (2.45,3.5) -- (3.55,2.5);

\fill[black, fill opacity=0.4] (0,0) circle(15pt);
\fill[black, fill opacity=0.4] (0,1.5) circle(15pt);
\fill[black, fill opacity=0.4] (0,3) circle(15pt);

\fill[black, fill opacity=0.4] (2,-0.75) circle(15pt);
\fill[black, fill opacity=0.4] (2,0.75) circle(15pt);
\fill[black, fill opacity=0.4] (2,2.25) circle(15pt);
\fill[black, fill opacity=0.4] (2,3.75) circle(15pt);

\fill[black, fill opacity=0.4] (4,0.75) circle(15pt);
\fill[black, fill opacity=0.4] (4,2.25) circle(15pt);
\end{tikzpicture}
& \begin{tikzpicture} \draw[->,thick] (0,0) -- (1.25,0); \end{tikzpicture} 
& $\vec{F} \; \left(\overset{!}{\approx} \shD_\gen (\f)_{\inv} \right)$ 
\end{tabular}
\end{table}
\noindent The architecture, referred to as \emph{\methodname{}} -- a network that learns  phantom-like\footnote{``\emph{Phantom:} Something apparently seen, heard, or sensed, but having no physical reality"; definition from \cite{collins}.} coefficients; see Section \ref{sec:DLmethod} and Figure \ref{fig:workflow} for details --  is a modified U-Net CNN \cite{ronneberger2015u}, which is a popular choice in the field of inverse problems, e.g.~\cite{kang2017,unser2017}. 

Having an estimation of the invisible part of the wavefront set at hand, the final step consists in fusing both parts and mapping the output back to the image domain via the inverse shearlet transform: 

{\noindent\bf{Step 3 - Combine both Parts:}}
\begin{equation*}
 \prop = \shD_\gen^{-1} \left( \shD_\gen (\soluD)_{\vis} + \vec{F}  \right)
\end{equation*}

\newpage
Concluding, the deep neural network is only used to infer the invisible shearlet coefficients, hence to estimate only the truly invisible boundary information. The visible part is entirely treated by the well understood method of sparse regularization with shearlets, which increases the overall reliability of our reconstructions. By assigning a clear task to the neural network, namely estimating invisible edge information, we gain a deeper understanding of our hybrid reconstruction framework. Furthermore, our network takes a rather accurate reconstruction of the visible coefficients as input, making the estimation of the invisible information easier. Additionally, the central question of how well our results generalize to unrelated testing data is only relevant on the invisible part. These advantages however come with a grain of salt due to the computational complexity of $\ell_1$-minimization, which is dominating the running time of our approach. 

\subsection{Expected Impact}

We anticipate our results to have the following impacts:

\begin{itemize}
\item {\em Limited Angle CT}. We propose a reconstruction framework that allows to complete the gaps in the wavefront set caused by the missing wedge of limited angle CT. We demonstrate  that deep neural networks are capable of inferring the invisible parts in the shearlet domain. 

\item {\em Hybrid Methods}. Our numerical experiments in Section \ref{sec:numerics} show that our hybrid method outperforms both, traditional model-based reconstruction schemes and more data-oriented methods. Thus, it supports the often advocated strategy to ``take the best out of both worlds'', and gives evidence to the potential of such combined approaches. 
\item {\em Interpretable Deep Learning}. Our results reveal a possibility of utilizing deep learning in a more controlled manner by applying it precisely to the part -- here coined the ``invisible'' part -- 
which defies any model-based approach. 
In this sense, the reconstruction method allows for a comprehensible interpretation, where machine learning is only used for inferring lost information. If a theory for inpainting with deep learning became available, our framework might allow for a transfer of these results to limited angle CT. \\
\end{itemize}
One should also stress that this concept might also be applicable to other inverse problems, predominantly those with a substantial amount of missing or distorted data.
\subsection{Outline}

The paper is organized as follows. Section \ref{sec:background} is devoted 
to reviewing the theoretical background of limited angle CT and the shearlet transform. In Section \ref{sec:visibleinvisible}, we detail a
key idea of our approach, namely the decomposition of the data (in the phase space) into a visible 
and an invisible part by means of $\ell^1$-regularization with shearlets. Our algorithmic approach that infers the invisible wavefront information by a deep neural network is introduced in Section \ref{sec:prop}, where we also give some background information on deep learning and discuss our network architecture \methodname{}. Finally, we demonstrate the performance of our 
methodology by a series of numerical experiments (see Section \ref{sec:numerics}).  Concluding remarks and future perspectives are briefly summarized in Section 6.

\section{Theoretical Background}
\label{sec:background}

In this section, we summarize the theoretical concepts that are essential for our proposed recovery framework. We first discuss results from microlocal analysis that explain which edge information is available in the acquired data and then introduce the reader to shearlets, which will be the key ingredient to access the visible information.

\subsection{Visibility of Singularities in Limited Angle CT}

\label{sec:vis}

 There is a body of work based on microlocal analysis that gives a precise description which singularities are visible in the limited data and which singularities cannot be determined \cite{Quinto93,Frikel13a,Nguyen15}. Since these insights will be central for our proposed reconstruction architecture we will briefly summarize some of those in the following. First, the notion of wavefront sets is required, which allows to simultaneously describe the location and direction of a singularity of a function $f$. It is based on a localized correspondence of smoothness and rapid decay in the Fourier domain.

\newpage 

    \begin{definition}[\cite{hormander2003}]
    \label{def:wf}
    Let $f\in L_{\text{loc}}^2(\R^2)$, i.e.,~$f$ is square integrable on every compact subset of $\R^2$. Let $\xo \in \R^2$ and $\xio \in \R^2\backslash \left\{ \vec{0} \right\}$. Then $f$ is said to be \emph{smooth at $\xo$ in the direction $\xio$}, if there exists a smooth cut-off function $\phi \in C_c^{\infty}(\R^2)$ such that $\phi(\xo) \neq 0$ and an open cone $V_{\xio} \subseteq \R^2$ containing $\xio$ such that given $N \in \N$ there exists a $C_N$ with
    \begin{equation*}
        \left| \widehat{\phi \cdot f}(\vec{\xi}) \right| \leq C_N (1 + \norm{\vec{\xi}})^{-N},
    \end{equation*}
    for all $\vec{\xi} \in V_{\xio}$, where $\widehat{\phi \cdot f}$ denotes the Fourier transform of the product $\phi \cdot f$.
    Furthermore, the \emph{wavefront set of $f$} is defined by
    \begin{equation*}
        \WF (f) := \left\{(\xo,\xio) \in \R^2\times \R^2 : \mbox{$f$ is not smooth at $\xo$ in the direction $\xio$} \right\}.
    \end{equation*}
   The wavefront set is a subset of the cotagent space $T^*(\R^2)$.
    \end{definition}

\begin{figure}
     \centering

    \begin{subfigure}[t]{0.45\textwidth}
		\centering
		    \begin{tikzpicture}[scale=0.7]
            \draw[->] (-2,-2.5) -- (-2,3.5);
            \draw[->] (-3.5,-2) -- (3.5,-2);
            \draw[fill=black, fill opacity=0.1]  plot[smooth, tension=.7] coordinates {(-1.4,0.5) (-1,1.2) (0,1.5) (0.5,1) (1,1.3) (1.5,1.8) (2,0.5) (2.3, 0) (1.7,-0.5) (0.8,-1) (0,-0.7) (-0.5,-1.2) (-1,-0.8) (-1.45, 0) (-1.4,0.5)};
            \fill[thick] (1.95,0.6) circle (0.05cm);
            \draw[->,thick] (1.95,0.6) -- (2.75,0.9);
            \draw (1.75,0.4) node{$\xo$};
            \draw (3.25,-2.25) node{$x_1$};
            \draw (-2.3,3.25) node{$x_2$};
            \draw (2.5,1.2) node{$\xio$};
            \end{tikzpicture}
		\caption{}
		\label{fig:wavefront:1}
	\end{subfigure}%
	\quad
	\begin{subfigure}[t]{0.45\textwidth}
		\centering
            \tdplotsetmaincoords{60}{-35}
            \begin{tikzpicture}
            [tdplot_main_coords,
            	cube/.style={very thin, black},scale=0.5]
            	
            \draw[cube] (3,3,0) -- (3,-3,0)  -- (-3,-3,0) -- (-3,3,0) -- cycle;
            \fill[cube,yellow,opacity=0.1] (3,3,0) -- (3,-3,0)  -- (-3,-3,0) -- (-3,3,0) -- cycle;	
            \fill[cube,opacity=0.1] (3,3,0) -- (3,3,5) -- (3,-3,5)  --  (3,-3,0);
            \fill[cube,opacity=0.1] (3,3,0) -- (3,3,5)  -- (-3,3,5) -- (-3,3,0);
            \draw[cube] (3,3,0) -- (3,3,5); 
            \draw[cube] (3,-3,0) -- (3,-3,5); 
            \draw[cube] (-3,3,0) -- (-3,3,5); 

            \draw[cube,->] (-3,-3.5,0)  -- (3,-3.5,0);
            \draw (2.5,-4.25,0) node{$x_1$};
            \draw[cube,->] (-3.5,-3,0) -- (-3.5,3,0);
            \draw (-3.95,2.5,0) node{$x_2$};
            \draw[cube,->] (3.5,-3,0) -- (3.5,-3,5);
            \draw (3.35,-4,4.75) node{$\theta$};


            \draw[gray, thick, dotted] plot[smooth, tension=.7] coordinates {(-1.4,0.5,0) (-1,1.2,0) (0,1.5,0) (0.5,1,0) (1,1.3,0) (1.5,1.8,0) (2,0.5,0) (2.3, 0,0) (1.7,-0.5,0) (0.8,-1,0) (0,-0.7,0) (-0.5,-1.2,0) (-1,-0.8,0) (-1.45, 0,0) (-1.4,0.5,0)};

            \draw[red, ultra thick] plot[smooth, tension=.7] coordinates {(-1.4,0.5,3.75) (-1,1.2,3.5) (0,1.5,3.0) (0.5,1,2.75) (1,1.3,2.5) (1.5,1.8,2.5) (2,0.5,2.75) (2.3, 0,1.75) (1.7,-0.5,1.75) (0.8,-1,1.75) (0,-0.7,1.25) (-0.5,-1.2,0.75) (-1,-0.8,0.5) (-1.45, 0,0.25) (-1.4,0.5,0)};
            \end{tikzpicture}
		\caption{}
		\label{fig:wavefront:2}
	\end{subfigure}%
	
     \caption{ \subref{fig:wavefront:1} Visualization of a point $(\vec{x}_0,\vec{\xi}_0) \in \WF (f)$ where $f = \indset{D}$ for a set $D \subseteq\R^2$ with smooth boundary. 
     Such indicator functions $\indset{D}$ are examples of conormal distributions \cite{hormander2007} of which the wavefront set $\WF (f)$ is contained in the conormal bundle $N^*(S)$ of a smooth surface $S$ (or a curve in the 2-dimensional case). The bundle $N^*(S)$ consists of the points $(\x,\vec{\xi})$ where $\x\in S$ and $\vec{\xi}$ is normal to $S$. For a curve $S\subset \R^2$ the conormal bundle $N^*(S)$ is a 2-dimensional submanifold of the 4-dimensional space $T^*(\R^2) $, that is, the wavefront set of $f$ is contained in a smooth, low dimensional subset of the phase space.}
     \label{fig:wavefront}
 \end{figure}
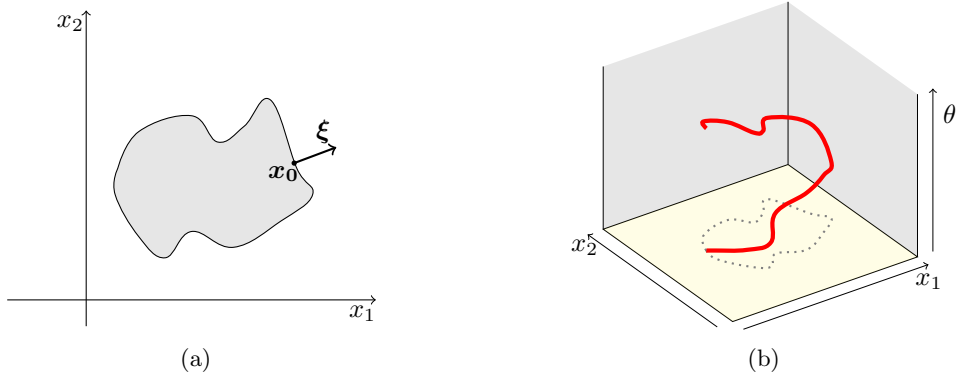

The wavefront set is often visualized in the \emph{phase space}, i.e., in the set of position-orientation pairs $(\xo,\theta)$, where $\xo \in \R^2$ and $\theta$ is in the real projective space $\mathbb{P}_1$ in $\R^2$ (freely identified with $[0,\pi)$). $\WF (f)$ can be seen as a subset of the phase space, encoding the positions and directions in which $f$ is non-smooth, see Figure \ref{fig:wavefront} for a visualization of a simple example.

The previous description of positions and directions of singularities, in the sense of Definition \ref{def:wf}, allows for a characterization of their visibility in computed tomography.
    \begin{theorem}[\cite{Quinto93,Quinto17}] \label{thm:quinto}
    Let $f \in L_{\text{loc}}^2(\R^2)$ and $L_0 = L(\theta_0,s_0)$ be a line in the plane. Let $(\xo,\xio) \in \WF (f)$ such that $\xo \in L_0$ and $\xio \in \R^2$ is a normal vector to $L_0$. Then the following holds.
    \begin{enumerate}
        \item[(i)] The singularity of $f$ at $(\xo,\xio)$ causes a unique singularity in $\WF (\Radon f)$ at $(\theta_0,s_0)$.
        \item[(ii)] Singularities of  $f$ not tangent to $L(\theta_0,s_0)$ do not cause singularities in $\Radon f$ at $(\theta_0,s_0)$.
    \end{enumerate}
    \end{theorem}

The implications of the previous theorem for limited angle CT are colloquially summarized in the following general principle \cite{Quinto17}:

\vspace*{0.25cm}

    \begin{princ}{pr:1}
    \begin{enumerate}
        \item[i)] A singularity of $f$ that is tangent to a line contained in the limited angle data set should be ``easy" to reconstruct. We will refer to such boundaries as \emph{visible}.
        \item[ii)] Singularities of $f$ that are not tangent to a line in the limited angle data set should be impossible to reconstruct. They are referred to as \emph{invisible}.
    \end{enumerate}
    \end{princ}
    
    \vspace*{0.25cm}
    
Let us point out that the information which boundaries are (in-)visible is completely determined by the measurement setup, i.e., by the sampled angular range $[-\phi,\phi]$. Therefore it is known \textit{a priori} and can be used for reconstruction purposes.


\subsection{Shearlets}
\label{sec:shear}

In the following, we will give a brief introduction to shearlet frames and discuss their properties when used as an $\ell_1$-regularizer for limited angle CT. For the sake of readability we aim at conveying the general ideas and refer the interested reader to the cited literature for more details and precise formulations of the described results.

\subsubsection{The Continuous Shearlet Transform}

\begin{figure}
\centering
\begin{subfigure}[t]{0.3\textwidth}
\begin{tikzpicture}[scale=0.6]
\draw[->] (0,-2.75) -- (0,2.75); 
\draw[->] (-2.75,0) -- (2.75,0); 
\draw[->,thick] (-1.0,-0.25) -- (-0.35,-1.0); 
\draw[<->,thick] (-2.25,0.75) -- (0.5,-2.5); 
\draw[<->,thick] (0.75,-2.6) -- (1.15,-2.2); 
\fill[rotate around={-50:(-0.25,-1.75)}, fill opacity=0.2](-1.35,-1.35) ellipse (2.25cm and 0.2 cm);
\draw (2.5,0.35) node{$x_1$};
\draw (0.35,2.5) node{$x_2$};
\draw (1.15,-2.65) node{$a$};
\draw (-1.5,-1.25) node{$\sqrt{a}$};
\draw (-0.2,-0.35) node{$-1/s$};
\end{tikzpicture}
\caption{}
\label{fig:geo_shear:1}
\end{subfigure}%
\quad
\begin{subfigure}[t]{0.3\textwidth}
\begin{tikzpicture}[scale=0.6]
\draw[->] (0,-0.5) -- (0,5); 
\draw[->] (-0.5,0) -- (5,0); 
\draw  plot[smooth, tension=0.65] coordinates {(0.55,0.25) (0.35,1.15) (1.15,1.8) (1.35,3.1) (2,3.85)   (3.4,4.25)};
\fill[fill opacity=0.15] (0.35,0.1) rectangle (0.75,0.5);
\fill[fill opacity=0.15] (0.2,0.45) rectangle (0.6,0.85);
\fill[fill opacity=0.15] (0.15,0.8) rectangle (0.55,1.2);
\fill[fill opacity=0.15] (0.25,1.1) rectangle (0.65,1.5);
\fill[fill opacity=0.15] (0.5,1.35) rectangle (0.9,1.75);
\fill[fill opacity=0.15] (0.8,1.55) rectangle (1.2,1.95);
\fill[fill opacity=0.15] (1.1,1.85) rectangle (1.5,2.25);
\fill[fill opacity=0.15] (1.05,2.2) rectangle (1.45,2.6);
\fill[fill opacity=0.15] (1.15,2.55) rectangle (1.55,2.95);
\fill[fill opacity=0.15] (1.2,2.9) rectangle (1.6,3.3);
\fill[fill opacity=0.15] (1.3,3.25) rectangle (1.7,3.65);
\fill[fill opacity=0.15] (1.5,3.55) rectangle (1.9,3.95);
\fill[fill opacity=0.15] (1.75,3.7) rectangle (2.15,4.1);
\fill[fill opacity=0.15] (2.05,3.85) rectangle (2.45,4.25);
\fill[fill opacity=0.15] (2.4,3.95) rectangle (2.8,4.35);
\fill[fill opacity=0.15] (2.7,4.0) rectangle (3.1,4.4);
\fill[fill opacity=0.15] (3.05,4.05) rectangle (3.45,4.45);
\draw (4.65,0.25) node{$x_1$};
\draw (0.35,4.75) node{$x_2$};
\end{tikzpicture}
\caption{}
\label{fig:geo_shear:2}
\end{subfigure}%
\quad
\begin{subfigure}[t]{0.3\textwidth}
\begin{tikzpicture}[scale=0.6]
\draw[->] (0,-0.5) -- (0,5); 
\draw[->] (-0.5,0) -- (5,0); 
\draw  plot[smooth, tension=0.65] coordinates {(0.55,0.25) (0.35,1.15) (1.15,1.8) (1.35,3.1) (2,3.85)   (3.4,4.25)};
\fill[fill opacity=0.15] (0.35,0.1) -- (0.15,1.1) -- (0.55,1.1) -- (0.75,0.1) -- cycle;
\fill[fill opacity=0.15] (0.2,1.05) -- (0.6,1.05)  -- (1.35,1.85) -- (0.95,1.85) -- cycle;
\fill[fill opacity=0.15] (1.0,1.8) -- (1.4,1.8)  -- (1.55,2.8) -- (1.15,2.8) -- cycle;
\fill[fill opacity=0.15] (1.1,2.75) -- (1.5,2.75)  -- (1.9,3.75) -- (1.5,3.75) -- cycle;
\fill[fill opacity=0.15] (1.5,3.7) -- (2.0,3.7)  -- (2.95,4.1) -- (2.45,4.1) -- cycle;
\fill[fill opacity=0.15] (2.25,4.05) -- (2.85,4.05) -- (4.0,4.35) -- (3.4,4.35) -- cycle;
\draw (4.65,0.25) node{$x_1$};
\draw (0.35,4.75) node{$x_2$};
\end{tikzpicture}
\caption{}
\label{fig:geo_shear:3}
\end{subfigure}%
    \caption{ \subref{fig:geo_shear:1} Visualization of the (asymptotic) orientation of a shearlet $\cshear$ for $a\rightarrow 0$. \subref{fig:geo_shear:2} shows a covering of a boundary section with isotropic wavelet elements \cite{mallat09} and \subref{fig:geo_shear:3} with anisotropic shearlet elements.}
    \label{fig:geo_shear}
\end{figure}
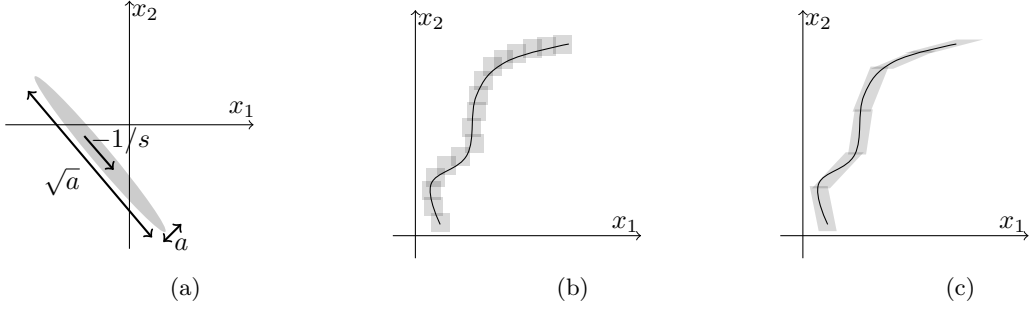

The basic construction of shearlets is based on applying three different operations to a well chosen \emph{generator function} $\gen \in L^2(\R^2)$, obtaining elements of the form
\begin{equation*}
    \cshear = |\det \Mas|^{1/2} \cdot \gen \left( \Mas (\cdot - \tp) \right),
\end{equation*}
where $\tp \in \R^2$ encodes \emph{translations} and $(a,s) \in \R_+\times \R$ controls the \emph{parabolic scaling matrix} $A_a$ and the \emph{shearing matrix} $S_s$ via the composite matrix
\begin{equation*}
    \Mas :=A_a^{-1}S_s^{-1} =   \left( \begin{matrix} a & 0\\ 0 & \sqrt{a}\end{matrix} \right)^{-1} \left( \begin{matrix} 1 & s\\ 0 & 1 \end{matrix} \right)^{-1}.
\end{equation*}
The \emph{continuous shearlet transform} is then defined as the mapping
\begin{equation*}
    L^2(\R^2) \ni f \mapsto \SH_\gen f (a,s,\tp) = \langle f, \cshear \rangle, \quad (a,s,\tp) \in \R_+\times \R \times \R^2.
\end{equation*}
Thus, $\SH_\psi$ analyzes the function $f$ around the location $\tp$ at different resolutions and orientations encoded by the scale and  shearing parameters $a$ and $s$, respectively. The (asymptotic) orientation of such a shearlet $\cshear$ is visualized in Figure \ref{fig:geo_shear:1}. Throughout this work, it is assumed that $\hat \gen$ has compact support such as in the case of a classical shearlet in \cite{kutyniok2009}.

The continuous shearlet transform has become a well studied research object in the last decade: it turns out that it exhibits an unwanted directional bias, which  is circumvented by considering the so-called \emph{cone-adapted} shearlet transform (see also next section). Furthermore, it can be shown that, under mild conditions, the shearlet system forms a continuous frame, implying for instance that a reconstruction of $f$ from its shearlet transform is possible \cite{Grohs2011}.

Of particular importance for our work are the results of \cite{Grohs2011,kutyniok2009}, in which it is shown that the continuous shearlet transform allows resolving the wavefront set of distributions by analyzing the decay properties of the continuous shearlet transform. Due to the cone-adaption, the precise statements are of rather technical nature and we will only give the general principle here: assume that $\xio \in \R^2 \backslash \left\{  \vec{0} \right\}$ with $\xi_2/\xi_1 \in [-1,1]$. It can be shown that $f$ is smooth at $\xo$ in the direction $\xio$, if and only if there is an open neighbourhood $U$ of $(\xi_2/\xi_1,\xo)$ such that
\begin{equation*}
    \abs{\SH_\psi f (a,s,\tp)}  \in \mathcal{O} (a^k), \quad \mbox{as } a\rightarrow 0,
\end{equation*}
for all $k \in \N$, with the $\mathcal{O}(\cdot)$-term uniform over $(s,\tp)$ in $U$. Similar results hold true for other directions $\xio \in \R^2 \backslash \left\{ \vec{0}\right\}$.
Overall we can summarize that:

\vspace*{0.25cm}

\begin{highlight}
    The wavefront set of $f$ is resolved by distinguishing different decay rates of its continuous shearlet transform.
\end{highlight}

\subsubsection{The Discrete Shearlet System}
\label{sec:shearlets}

Starting from the continuous transform, the goal is now to derive a discrete system of functions that allows to encode anisotropic features in the digital realm.
Such a \emph{discrete shearlet system} can be formally obtained by sampling the parameter space $\R_+ \times \R \times \R^2$ on a discrete subset.
The so-called \emph{regular discrete shearlet system} associated with $\gen \in L^2(\R^2)$ is defined by
\begin{equation}\label{eq:sh}
    \sh (\gen) := \left\{  \shear = 2^{(3j)/4} \gen (S_k A_{2^j} \cdot - \lp), \mbox{ for  } j,k \in \Z, \lp \in \Z^2 \right\}.
\end{equation}
Furthermore, the \emph{discrete shearlet transform} is defined as the mapping
\begin{equation*}
    L^2(\R^2) \ni f \mapsto \sh_\gen f (j,k,\lp) = \langle f, \shear \rangle, \quad (j,k,\lp) \in \Z \times \Z \times \Z^2.
\end{equation*}
We point out that the essence of shearlets lies in the fact that the shearing matrix $S_k$ preserves the integer lattice for $k \in \Z$, which is desired for a numerical implementation. Under mild assumptions on the generator $\gen$, it can be shown that the system $\sh (\gen)$ forms a \emph{Parseval frame} \cite{christensen2003} of $L^2(\R^2)$, giving rise to the shearlet representation
\begin{equation}
    \label{eq:shearf}
    f = \sum_{(j,k,\lp) \in \Z \times \Z \times \Z^2} \langle f, \shear \rangle \, \shear = \sh_\gen^T (\sh_\gen (f)).
\end{equation}

\begin{figure}
\centering
\begin{subfigure}[t]{0.3\textwidth}
\begin{tikzpicture}[xscale = 0.5, yscale=0.45]
\draw[white] (-1.8,-6.4) node{{\footnotesize Semi-visible}};

\draw[->] (-5,0) -- (5,0);
\draw[->] (0,-5.5) -- (0,5.5);
\draw (0.3, 0.25) -- (4,4.5) -- (4,-4.5) -- (0.3, -0.25) -- (0.3, 0.25);
\draw (-0.3, 0.25) -- (-4,4.5) -- (-4,-4.5) -- (-0.3, -0.25) -- (-0.3, 0.25);
\draw (4.75,0) node[above]{$\xi_1$};
\draw (0,5.25) node[right]{$\xi_2$};

\draw[dotted,thick] (-0.3,0.3) -- (0.3,0.3); 
\draw[dotted,thick] (-0.3,-0.3) -- (0.3,-0.3); 

\draw (0.3,0.26) - - (0.3,-0.26);
\draw (-0.3,0.26) - - (-0.3,-0.26);
\draw (0.55,0.55) - - (0.55,-0.55);
\draw (-0.55,0.55) - - (-0.55,-0.55);
\draw (1.35,1.47) - - (1.35,-1.47);
\draw (-1.35,1.47) - - (-1.35,-1.47);

\draw (0.3,0.125) - - (0.55,0.27);
\draw (0.3,-0.125) - - (0.55,-0.27);
\draw (-0.3,0.125) - - (-0.55,0.27);
\draw (-0.3,-0.125) - - (-0.55,-0.27);
\draw (0.3,0.26) - - (0.3,0.8) - - (0.55,2.5) - - (0.55,0.55);
\draw (0.3,-0.26) - - (0.3,-0.8) - - (0.55,-2.5) - - (0.55,-0.55);
\draw (-0.3,0.26) - - (-0.3,0.8) - - (-0.55,2.5) - - (-0.55,0.55);
\draw (-0.3,-0.26) - - (-0.3,-0.8) - - (-0.55,-2.5) - - (-0.55,-0.55);
\draw (0.3,0.38) - - (0.55,0.8);
\draw (0.3,0.5) - - (0.55,1.2);
\draw (0.3,0.63) - - (0.55,1.8);
\draw (0.3,-0.38) - - (0.55,-0.8);
\draw (0.3,-0.5) - - (0.55,-1.2);
\draw (0.3,-0.63) - - (0.55,-1.8);
\draw (-0.3,0.38) - - (-0.55,0.8);
\draw (-0.3,0.5) - - (-0.55,1.2);
\draw (-0.3,0.63) - - (-0.55,1.8);
\draw (-0.3,-0.38) - - (-0.55,-0.8);
\draw (-0.3,-0.5) - - (-0.55,-1.2);
\draw (-0.3,-0.63) - - (-0.55,-1.8);

\draw (0.55,0.4) - - (1.35,1.05);
\draw (0.55,0.154) - - (1.35,0.45);
\draw (0.55,-0.154) - - (1.35,-0.45);
\draw (0.55,-0.4) - - (1.35,-1.05);
\draw (-0.55,0.4) - - (-1.35,1.05);
\draw (-0.55,0.154) - - (-1.35,0.45);
\draw (-0.55,-0.154) - - (-1.35,-0.45);
\draw (-0.55,-0.4) - - (-1.35,-1.05);
\draw (1.35,1.47) - - (1.35,4) - - (0.55,1.5) - - (0.55,0.55);
\draw (1.35,-1.47) - - (1.35,-4) - - (0.55,-1.5) - - (0.55,-0.55);
\draw (-1.35,1.47) - - (-1.35,4) - - (-0.55,1.5) - - (-0.55,0.55);
\draw (-1.35,-1.47) - - (-1.35,-4) - - (-0.55,-1.5) - - (-0.55,-0.55);
\draw (0.55,0.7) - - (1.35,2);
\draw (0.55,0.95) - - (1.35,2.7);
\draw (0.55,1.25) - - (1.35,3.3);
\draw (0.55,-0.7) - - (1.35,-2);
\draw (0.55,-0.95) - - (1.35,-2.7);
\draw (0.55,-1.25) - - (1.35,-3.3);
\draw (-0.55,0.7) - - (-1.35,2);
\draw (-0.55,0.95) - - (-1.35,2.7);
\draw (-0.55,1.25) - - (-1.35,3.3);
\draw (-0.55,-0.7) - - (-1.35,-2);
\draw (-0.55,-0.95) - - (-1.35,-2.7);
\draw (-0.55,-1.25) - - (-1.35,-3.3);
\filldraw[fill=gray,fill opacity=0.9] (0.55,0.95) - - (1.35,2.7) - - (1.35,3.3) - - (0.55,1.25) - - (0.55,0.95);
\filldraw[fill=gray,fill opacity=0.9] (-0.55,-0.95) - - (-1.35,-2.7) - - (-1.35,-3.3) - - (-0.55,-1.25) - - (-0.55,-0.95);

\draw (1.35,1.25) - - (4,3.5);
\draw (1.35,0.9) - - (4,2.5);
\draw (1.35,0.55) - - (4,1.5);
\draw (1.35,0.2) - - (4,0.5);
\draw (1.35,-0.2) - - (4,-0.5);
\draw (1.35,-0.55) - - (4,-1.5);
\draw (1.35,-0.9) - - (4,-2.5);
\draw (1.35,-1.25) - - (4,-3.5);
\draw (-1.35,1.25) - - (-4,3.5);
\draw (-1.35,0.9) - - (-4,2.5);
\draw (-1.35,0.55) - - (-4,1.5);
\draw (-1.35,0.2) - - (-4,0.5);
\draw (-1.35,-0.2) - - (-4,-0.5);
\draw (-1.35,-0.55) - - (-4,-1.5);
\draw (-1.35,-0.9) - - (-4,-2.5);
\draw (-1.35,-1.25) - - (-4,-3.5);

\fill (2.5,3.5) circle (1pt);
\fill (2.5,3.7) circle (1pt);
\fill (2.5,3.9) circle (1pt);
\fill (2.5,-3.5) circle (1pt);
\fill (2.5,-3.7) circle (1pt);
\fill (2.5,-3.9) circle (1pt);
\fill (-2.5,3.5) circle (1pt);
\fill (-2.5,3.7) circle (1pt);
\fill (-2.5,3.9) circle (1pt);
\fill (-2.5,-3.5) circle (1pt);
\fill (-2.5,-3.7) circle (1pt);
\fill (-2.5,-3.9) circle (1pt);
\fill (0.85,3.5) circle (1pt);
\fill (0.85,3.7) circle (1pt);
\fill (0.85,3.9) circle (1pt);
\fill (0.85,-3.5) circle (1pt);
\fill (0.85,-3.7) circle (1pt);
\fill (0.85,-3.9) circle (1pt);
\fill (-0.85,3.5) circle (1pt);
\fill (-0.85,3.7) circle (1pt);
\fill (-0.85,3.9) circle (1pt);
\fill (-0.85,-3.5) circle (1pt);
\fill (-0.85,-3.7) circle (1pt);
\fill (-0.85,-3.9) circle (1pt);
\end{tikzpicture}
\caption{}
\label{fig:tiling:1}
\end{subfigure}%
\quad
\begin{subfigure}[t]{0.3\textwidth}
\begin{tikzpicture}[scale=0.45]
\draw[->] (-5.5,0) -- (5.5,0);
\draw[->] (0,-5.5) -- (0,5.5);
\draw (5.25,0) node[above]{$\xi_1$};
\draw (0,5.25) node[right]{$\xi_2$};
\filldraw[fill=black,fill opacity=0.5] (-1,-1) rectangle (1,1);
\filldraw[fill=black,fill opacity=0.3] (4.5,4.5) -- (1,1) -- (1,-1) -- (4.5,-4.5);
\filldraw[fill=black,fill opacity=0.15] (4.5,4.5) -- (1,1) -- (-1,1) -- (-4.5,4.5);
\filldraw[fill=black,fill opacity=0.3] (-4.5,4.5) -- (-1,1) -- (-1,-1) -- (-4.5,-4.5);
\filldraw[fill=black,fill opacity=0.15] (-4.5,-4.5) -- (-1,-1) -- (1,-1) -- (4.5,-4.5);
\draw (-3.2,-1.75) node[above] {$\mcC_{\text{h}}$};
\draw (3.2,1.75) node[above] {$\mcC_{\text{h}}$};
\draw (-1.75,3.2) node[above] {$\mcC_{\text{v}}$};
\draw (1.75,-3.2) node[above] {$\mcC_{\text{v}}$};
\draw (-0.5,0.025) node[above] {$\mcR$};

\draw[white] (-1.8,-6.4) node{{\footnotesize Semi-visible}};
\end{tikzpicture}
\caption{}
\label{fig:tiling:2}
\end{subfigure}%
\quad
\begin{subfigure}[t]{0.3\textwidth}
\begin{tikzpicture}[scale=.5]
\draw[->] (-5,0) - - (5,0);
\draw[->] (0,-5) - - (0,5);
\draw (4.75,0) node[above]{$\xi_1$};
\draw (0,4.75) node[right]{$\xi_2$};
\draw (-4,-4) rectangle (4,4);
\draw (-2.4,-2.4) rectangle (2.4,2.4);
\draw (-1,-1) rectangle (1,1);
\draw[thick] (-0.35,-0.35) rectangle (0.35,0.35);

\draw[thick] (-4,-4) -- (-0.35,-0.35);
\draw[thick] (0.35,0.35)  -- (4,4);
\draw[thick] (-4,4) -- (-0.35,0.35);
\draw[thick] (0.35,-0.35) -- (4,-4);
\draw[dotted, thick] (4,4) -- (4.3,4.3);
\draw[dotted, thick] (-4,4) -- (-4.3,4.3);
\draw[dotted, thick] (4,-4) -- (4.3,-4.3);
\draw[dotted, thick] (-4,-4) -- (-4.3,-4.3);

\draw[ultra thin] (0.1,0.35) -- (0.35,1);
\draw[ultra thin] (-0.1,0.35) -- (-0.35,1);
\draw[ultra thin] (-0.1,-0.35) -- (-0.35,-1);
\draw[ultra thin] (0.1,-0.35) -- (0.35,-1);

\draw[ultra thin] (0.35,0.1) -- (1,0.35);
\draw[ultra thin] (0.35,-0.1) -- (1,-0.35);
\draw[ultra thin] (-0.35,0.1) -- (-1,0.35);
\draw[ultra thin] (-0.35,-0.1) -- (-1,-0.35);

\draw[ultra thin] (2.4,1.86) -- (4,3.1);
\draw[ultra thin] (2.4,1.32) -- (4,2.2);
\draw[ultra thin] (2.4,0.78) -- (4,1.3);
\draw[ultra thin] (2.4,0.24) -- (4,0.4);
\draw[ultra thin] (2.4,-1.86) -- (4,-3.1);
\draw[ultra thin] (2.4,-1.32) -- (4,-2.2);
\draw[ultra thin] (2.4,-0.78) -- (4,-1.3);
\draw[ultra thin] (2.4,-0.24) -- (4,-0.4);
\draw[ultra thin] (-2.4,1.86) -- (-4,3.1);
\draw[ultra thin] (-2.4,1.32) -- (-4,2.2);
\draw[ultra thin] (-2.4,0.78) -- (-4,1.3);
\draw[ultra thin] (-2.4,0.24) -- (-4,0.4);
\draw[ultra thin] (-2.4,-1.86) -- (-4,-3.1);
\draw[ultra thin] (-2.4,-1.32) -- (-4,-2.2);
\draw[ultra thin] (-2.4,-0.78) -- (-4,-1.3);
\draw[ultra thin] (-2.4,-0.24) -- (-4,-0.4);

\draw[ultra thin] (1.86,2.4) -- (3.1,4);
\draw[ultra thin] (1.32,2.4) -- (2.2,4);
\draw[ultra thin] (0.78,2.4) -- (1.3,4);
\draw[ultra thin] (0.24,2.4) -- (0.4,4);
\draw[ultra thin] (-1.86,2.4) -- (-3.1,4);
\draw[ultra thin] (-1.32,2.4) -- (-2.2,4);
\draw[ultra thin] (-0.78,2.4) -- (-1.3,4);
\draw[ultra thin] (-0.24,2.4) -- (-0.4,4);
\draw[ultra thin] (1.86,-2.4) -- (3.1,-4);
\draw[ultra thin] (1.32,-2.4) -- (2.2,-4);
\draw[ultra thin] (0.78,-2.4) -- (1.3,-4);
\draw[ultra thin] (0.24,-2.4) -- (0.4,-4);
\draw[ultra thin] (-1.86,-2.4) -- (-3.1,-4);
\draw[ultra thin] (-1.32,-2.4) -- (-2.2,-4);
\draw[ultra thin] (-0.78,-2.4) -- (-1.3,-4);
\draw[ultra thin] (-0.24,-2.4) -- (-0.4,-4);
\filldraw[fill=gray,fill opacity=0.5,ultra thin] (1.32,2.4) -- (1.86,2.4) -- (3.1,4) -- (2.2,4) -- (1.32,2.4);
\filldraw[fill=gray,fill opacity=0.5,ultra thin] (-1.32,-2.4) -- (-1.86,-2.4) -- (-3.1,-4) -- (-2.2,-4) -- (-1.32,-2.4);

\draw[ultra thin] (1,0.58) -- (2.4,1.4);
\draw[ultra thin] (1,0.19) -- (2.4,0.45);
\draw[ultra thin] (1,-0.58) -- (2.4,-1.4);
\draw[ultra thin] (1,-0.19) -- (2.4,-0.45);
\draw[ultra thin] (-1,0.58) -- (-2.4,1.4);
\draw[ultra thin] (-1,0.19) -- (-2.4,0.45);
\draw[ultra thin] (-1,-0.58) -- (-2.4,-1.4);
\draw[ultra thin] (-1,-0.19) -- (-2.4,-0.45);
\filldraw[fill=gray,fill opacity=0.9,ultra thin] (1,0.58) -- (2.4,1.4) -- (2.4,0.45) -- (1,0.19) -- (1,0.58);
\filldraw[fill=gray,fill opacity=0.9,ultra thin] (-1,-0.58) -- (-2.4,-1.4) -- (-2.4,-0.45) -- (-1,-0.19) -- (-1,-0.58);

\draw[ultra thin] (0.58,1) -- (1.4,2.4);
\draw[ultra thin] (0.19,1) -- (0.45,2.4);
\draw[ultra thin] (-0.58,1) -- (-1.4,2.4);
\draw[ultra thin] (-0.19,1) -- (-0.45,2.4);
\draw[ultra thin] (0.58,-1) -- (1.4,-2.4);
\draw[ultra thin] (0.19,-1) -- (0.45,-2.4);
\draw[ultra thin] (-0.58,-1) -- (-1.4,-2.4);
\draw[ultra thin] (-0.19,-1) -- (-0.45,-2.4);

\filldraw[fill=gray,fill opacity=0.2,ultra thin] (0.19,1) -- (0.45,2.4) -- (-0.45,2.4) -- (-0.19,1) -- (0.19,1);
\filldraw[fill=gray,fill opacity=0.2,ultra thin] (-0.19,-1) -- (-0.45,-2.4) -- (0.45,-2.4) -- (0.19,-1) -- (-0.19,-1);

\draw[thick,red] (2.35,4) -- (-2.35,-4);
\draw[thick,red] (2.35,-4) -- (-2.35,4);
\fill[fill=red,fill opacity=0.07] (0,0) -- (2.35,4) -- (4,4) -- (4,-4) -- (2.35,-4) -- (0,0);
\fill[fill=red,fill opacity=0.07] (0,0) -- (-2.35,4) -- (-4,4) -- (-4,-4) -- (-2.35,-4) -- (0,0);
\draw[red] (3.35,-1.45) node{$W_{\phi}$};

\fill[black,fill opacity = 0.1] (-4,-5.15) rectangle (-3.75,-4.9);
\fill[black,fill opacity = 0.3] (-4,-5.85) rectangle (-3.75,-5.6);
\fill[black,fill opacity = 0.65] (0.5,-5.15) rectangle (0.75,-4.9);
\fill[red,fill opacity = 0.2] (0.5,-5.85) rectangle (0.75,-5.6);

\draw[black] (-2.35,-5.0) node{{\footnotesize Invisible}};
\draw[black] (-1.8,-5.7) node{{\footnotesize Semi-visible}};
\draw[black] (1.85,-5.0) node{{\footnotesize Visible}};
\draw[red] (2.9,-5.75) node{{\footnotesize Visible Wedge}};
\end{tikzpicture}
\caption{}
\label{fig:tiling:3}
\end{subfigure}

\caption{Illustration of tilings in the Fourier domain. \subref{fig:tiling:1} shows the frequency support of elements in the regular shearlet system for different values of $j$ and $k$ and thereby reveals a directional bias. \subref{fig:tiling:2} indicates how the Fourier domain is separated in two conic regions and a low frequency part. In \subref{fig:tiling:3}, the tiling of the cone-adapted discrete shearlet system is visualized. Additionally, the visible wedge $W_{\phi}$ of Definition \ref{def:vis} is shown in red, together with examples for (in-)visible shearlets (see Section~\ref{sec:visShear}).}
    \label{fig:tiling}
\end{figure}
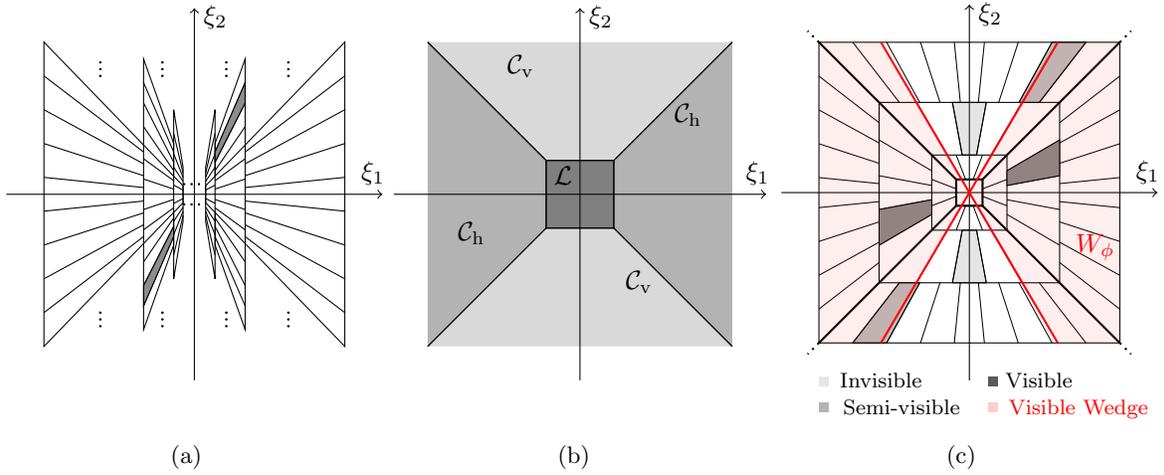

Similar as in the case of the continuous shearlet transform, the discrete system also exhibits an unwanted \emph{directional bias}. This side effect is visualized in Figure \ref{fig:tiling:1}, where the Fourier domain support of various elements in $\sh (\gen)$ corresponding to different values of $j$ and $k$ is shown. By partitioning the Fourier space into vertical and horizontal conic regions, denoted by $\mcC_{\text{v}}$ and $\mcC_{\text{h}}$ respectively, together with a separate low frequency part $\mcR$, a more uniform tiling is achieved, see Figure \ref{fig:tiling:2}.  This is reflected in the following basic definition of \emph{cone-adapted discrete shearlet systems}. 


\begin{definition}
\label{def:ShearletSystem}
Let $\phi, \psi \in L^2(\R^2)$. Then the \emph{cone-adapted discrete shearlet
system} is defined by
\begin{equation*}
\sh (\phi, \psi) = \Phi(\phi) \cup \Psi(\psi) \cup \tilde{\Psi}(\tilde{\psi}),
\end{equation*}
where
\begin{alignat*}{2}
\Phi(\phi) & :=  \Big\{ \psi_{0,0,\lp,0}  && := \phi(\cdot - \lp) :\lp \in \Z^2  \Big\},\\
\Psi(\psi) & := \Big\{ \psi_{j,k,\lp,1}  && := 2^{(3j)/4}\psi(S_k A_{j}\cdot - \lp ): j\in \N_0, k \in \Z, |k| \leq   2^{\left\lceil j/2 \right\rceil}, \lp \in \Z^2 \Big\},\\
\tilde{\Psi}(\tilde{\psi}) & :=\Big\{ \psi_{j,k,\lp,-1}  && := 2^{(3j)/4}\tilde{\psi}(S_k^T \tilde{A}_{j}\cdot -  \lp ): j\in \N_0, k \in \Z, |k| \leq
2^{\left\lceil j/2 \right\rceil}, \lp \in \Z^2 \Big\},
\end{alignat*}
with $\tilde{\psi}(x_1,x_2) := \psi(x_2,x_1)$ and $\tilde{A}_{j} = \diag(2^{j/2},2^{j}) \in \R^{2 \times 2}$. For ease of notation we introduce the index set 
\begin{align*}
\Lambda := \{(j,k,\lp,\iota): j \in \N_0, k \in \Z, |\iota| j \geq j\geq 0, |k|\leq |\iota| 2^{\left\lceil j/2 \right\rceil}, \lp \in \Z^2, \iota \in \{1,0,-1\} \}.
\end{align*}
The \emph{cone-adapted discrete shearlet transform} is then defined as the mapping
\begin{equation*}
    L^2(\R^2) \ni  f \mapsto \sh_{\gen,\phi} f (j,k,\lp,\iota) = \left( \langle f,\gen_{j,k,\lp,\iota} \rangle \right)_{(j,k,\lp,\iota) \in \Lambda}.
\end{equation*}
\end{definition}

In the previous definition, the function $\psi$ is referred to as \emph{shearlet generator}. Its corresponding systems $\Psi(\psi) $ and $\tilde{\Psi}(\tilde{\psi})$ essentially differ in the reversed roles of the input variables and therefore correspond to the horizontal and vertical conic region, respectively. Note that by restricting the range for the shearing variable $k$ on each cone, the orientations of the resulting functions are distributed more equally. This can be seen in the Fourier tiling of the cone-adapted system, which is depicted in Figure \ref{fig:tiling:3}. Finally, $\phi$ is referred to as the \emph{shearlet scaling function} and it is associated to the low frequency part $\mcR$, since it is chosen to have compact frequency support near the origin.

There are various extensions and refinements of this basic definition available in the literature, notably for instance the construction of compactly supported shearlets. We refrain from giving further details and refer the interested reader to \cite{Kutyniok2012} and references therein.
Instead, we conclude this excursion by stating a stylized approximation theorem for discrete shearlet frames, which shows that shearlets are an optimal sparsifying transform for a particular class of natural images.

\newpage

\begin{theorem}[\cite{Guo07}]
Let $\sh (\phi, \gen)$ be a variant of the cone-adapted shearlet system  (see \cite{Guo07} for precise definition). Let $f \in L^2(\R^2)$ be a \emph{cartoon-like} function, i.e., $f = f_0 + f_1 \cdot \indset{B}$, where $B \subseteq [0,1]^2$ is a set with $\partial B$ being a closed $C^2$-curve with bounded curvature and $f_i \in C^2(\R^2)$ with $\supp f_i \subseteq [0,1]^2$  and $\norm{f_i}_{C^2} \leq 1$.

Let $f_N$ be a nonlinear $N$-term approximation obtained by summing over the $N$ largest shearlet coefficients $\langle f, \gen_{j,k,\lp,\iota} \rangle$ in \eqref{eq:shearf}. Then there exists a constant $C>0$, independent of $f$ and $N$, such that
\begin{equation*}
    \norm{f - f_N}_2^2 \leq C N^{-2} \log^3 N, \quad \mbox{ as } N \longrightarrow \infty.
\end{equation*}
\end{theorem}

Ignoring the additional log-factors, the previous theorem reveals that shearlets allow for a $O(N^{-2})$ approximation rate of cartoon-like functions, which is proven to be the optimal rate \cite{Donoho2001}. It is also achieved by other anisotropic systems such as curvelets \cite{Candes02}, but out of reach for isotropic wavelet systems \cite{mallat09}. An intuitive explanation of this fact can be found in Figures \ref{fig:geo_shear:2} and \ref{fig:geo_shear:3}: the anisotropic scaling and the shearing allow to capture geometric features more efficiently than isotropic wavelet systems.



\section{The Concept of Visible and Invisible Coefficients}
\label{sec:visibleinvisible}

While the concepts of the previous section are infinite-dimensional and of rather abstract nature, a central question is how shearlets can help to access the visible part of the wavefront set in a practical manner. We will argue in this section that, at least heuristically, $\ell^1$-minimization allows us to distinguish between \emph{visible} and \emph{invisible shearlet coefficients.}

 \subsection{$\ell^1$-Analysis Minimization}
 \label{sec:l1Ana}
 Motivated by the observation that shearlets define an efficient sparsifying transform for images with anisotropic features, they are becoming an increasingly popular choice for sparsity based regularization of inverse problems \cite{Kutyniok2012,kutyniok2017,loock2014,easley2009}. In particular the sparsifying effect of $\ell^1$-regularization has been an active field of research over the last two decades, often leading to state of the art results for the inversion from limited data. Under the label \emph{compressed sensing} many powerful recovery guarantees for subsampled random measurements have been derived \cite{Candes2006,Donoho2006,Foucart2013}.

 In this work, we propose to use the shearlet system in an \emph{analysis based variational prior} for reconstructing reliable visible shearlet coefficients. This means that for obtaining an approximation $\solu$ from the measurements $y$ in \eqref{eq:meas}, we solve the convex optimization problem
\begin{equation}
\label{eq:l1-ana}
    \solu \in \argmin_{f\geq 0} \norm{\sh_{\gen,\phi} (f)}_{1,w}  + \frac{1}{2} \norm{\RadonLim f - y}_2^2,
\end{equation}
where $\norm{x}_{1,w} = \sum_j w_j |x_j|$ denotes the weighted $\ell^1$-norm of $x \in \ell^1(\Lambda)$ with weights $w \in \ell^2(\Lambda,\R_+)$. Such a weight vector balances the influence of the shearlet regularizer and the $\ell^2$-data fidelity term, subsuming the usual regularization parameter. In most tomographic problems, it is known a priorily that the desired image $f$ is non-negative and including this constraint into \eqref{eq:l1-ana} leads to superior reconstruction results.

\subsection{Visibility of Shearlets}
\label{sec:visShear}

We now discuss the implications of the visibility principle of Section \ref{sec:vis} on the obtained shearlet coefficients $\sh_{\gen,\phi} (\solu)$. 
From to the Visibility Principle \ref{pr:1}  we infer that the variational approach of \eqref{eq:l1-ana} should only be able to reconstruct boundaries which are visible in the limited angle data set. In terms of shearlet coefficients this means that coefficients corresponding to shearlets aligned with invisible boundaries of a solution $\solu$ are negligible. Note that the visible boundaries are completely determined by the measured angular range $[-\phi,\phi]$, which can be conveniently expressed in the frequency domain via the Fourier slice theorem \cite{Natterer01}. This motivates the following definition, which certainly only makes sense for bandlimited shearlet constructions, i.e., when $\hat  \gen$ has compact support.

\begin{definition}
\label{def:vis}
Let $\sh (\phi, \psi)$ be a bandlimited, cone-adapted discrete shearlet system and $\phi \in (0,\pi/2)$.
Then, the \emph{visible wedge} is defined by
\begin{equation*}
    W_{\phi} := \left\{ \xi \in \R^2 : \xi = r \cdot (\cos \omega, \sin \omega)^T, r\in \R, |\omega| \leq \phi \right\}.
\end{equation*}
Furthermore, we define the \emph{invisible shearlet indices} by
\begin{equation}
\label{eq:vis}
    \inv := \left\{(j,k,\lp,\iota) \in \Lambda : \supp \hat{\gen}_{j,k,\lp,\iota}  \cap W_{\phi}  = \emptyset\right\},
\end{equation}
and the \emph{visible indices} by $\vis := \Lambda \backslash \inv$.
\end{definition}

A visualization of the geometry described in Definition \ref{def:vis} can be found in Figure \ref{fig:tiling:3}. The notion of \emph{invisible coefficients} was originally coined by Frikel in \cite{Frikel2013b} for the case of curvelet frames, an anisotropic function system similar to shearlets, but based on rotation instead of shearing. It was proven that for invisible curvelet elements $\gen_j \in L^2(\R^2)$ it holds true that $\gen_j \in \ker \RadonLim$. This property was then used for dimension reduction of the \emph{synthesis-based} $\ell^1$-regularization
    \begin{equation*}
        z^\ast \in \argmin_z \norm{z}_{1,w} + \frac{1}{2} \norm{\RadonLim \left(\sum_{j\in \Lambda} \gen_j z_j\right)  - y}_2^2,
    \end{equation*}
where $\Lambda$ denotes the index set of the considered curvelet frame $(\gen_j)_{j \in \Lambda}$. It was shown that the coefficients associated to invisible curvelets satisfy $z^*_{j} = 0$, which can be immediately used to obtain an equivalent, smaller dimensional problem. Due to the similarities of shearlets and curvelets, these arguments directly translate to shearlet frames, as already pointed out in \cite{Frikel2013b}. 

We expect that a similar statement holds true for the analysis-based minimization of \eqref{eq:l1-ana}, i.e., for an invisible shearlet index $(j,k,\lp,\iota) \in \inv$ of a solution $\solu$ of \eqref{eq:l1-ana} it holds $\langle \solu, \gen_{j,k,\lp,\iota} \rangle \approx 0$. Although a theoretical analysis of this conjecture appears to be more complicated and therefore beyond the scope of this work, we have chosen the analysis formulation for our work for several reasons: first, it is known that analysis based methods usually yield better reconstruction quality in imaging applications. Furthermore, since the optimization is over the image domain, the resulting problem is of smaller dimension and therefore more efficient solvers exist. Also, the analysis formulation allows to naturally include the non-negativity constraints, which is fundamental in tomography applications. Finally, the analysis point of view is closer related to the characterization of wavefront sets with the continuous shearlet transform, which forms the theoretical foundation of our approach.

Before we present a numerical example that justifies the terminology of (in-)visibile shearlet elements, we first discuss a handy relaxation of Definition \ref{def:vis} in the following remark.  


\begin{remark}
For shearlets with $\supp \hat{\gen}_{j,k,\lp,\iota}  \not\subseteq W_{\phi}$ and $\supp  \hat{\gen}_{j,k,\lp,\iota} \cap W_\phi \neq \emptyset$, the \mbox{(in-)visibility} attribution can be less clear. While $\RadonLim \gen_{j,k,\lp,\iota} \neq 0$ in such a case, the contribution can still be negligible if most of the support lies outside the visible wedge $W_{\phi}$.  In our numerical experiments, we therefore relax the condition of \eqref{eq:vis} by classifying a shearlet as visible if its orientation, determined by the shearing and anisotropic scaling, corresponds to a visible direction of $\RadonLim$. This principle is visualized in Figure \ref{fig:tiling:3}, where the \emph{semi-visible} shearlet would be classified as visible since most of its support lies inside the visible wedge. 

In particular for a \emph{fanbeam} scanning geometry, the definition via the visible wedge breaks down. In this case, we propose a generalization of \eqref{eq:vis} by setting
\begin{equation*}
    \vis = \left\{ (j,k,\lp,\iota) : \norm{\RadonLim \gen_{j,k,\lp,\iota}}_2 > Q_j(\phi/\pi) \right\},
\end{equation*}
where $Q_j(p)$ denotes the $p$-quantile of all norms of the projected shearlets at scale $j$.
\end{remark}

\begin{figure}
        \centering

        \begin{subfigure}[t]{0.30\textwidth}
    		\centering
    		\includegraphics[width=\textwidth]{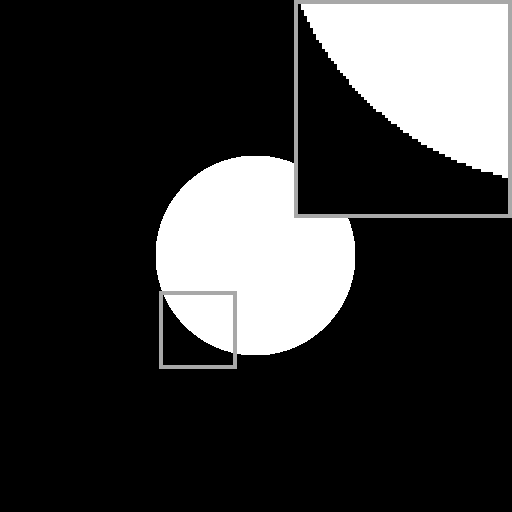}
    		\caption{}
    		\label{fig:exp_vis:1}
	    \end{subfigure}%
	    \quad
	    \begin{subfigure}[t]{0.3\textwidth}
    		\centering
    		\includegraphics[width=\textwidth]{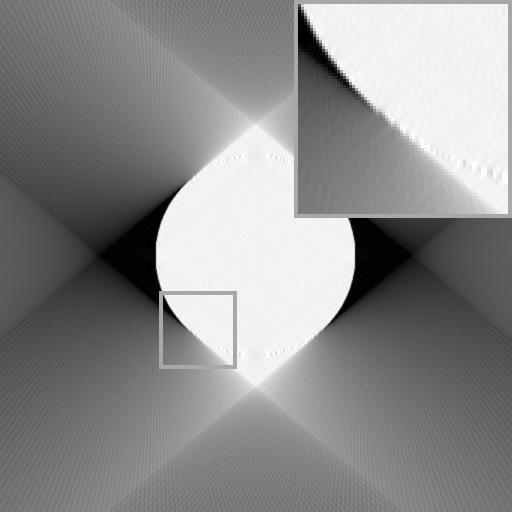}
    		\caption{}
    		\label{fig:exp_vis:2}
	    \end{subfigure}%
	    \quad
	    \begin{subfigure}[t]{0.3\textwidth}
    		\centering
    		\includegraphics[width=\textwidth]{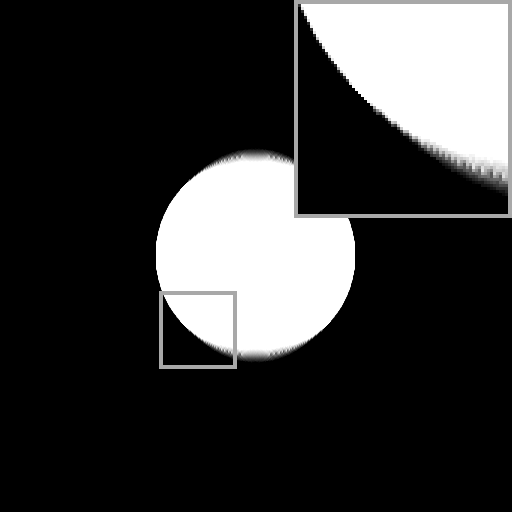}
    		\caption{}
    		\label{fig:exp_vis:3}
	    \end{subfigure}%
	    \\
	     \begin{subfigure}[t]{0.3\textwidth}
    		\centering
    		\includegraphics[width=\textwidth]{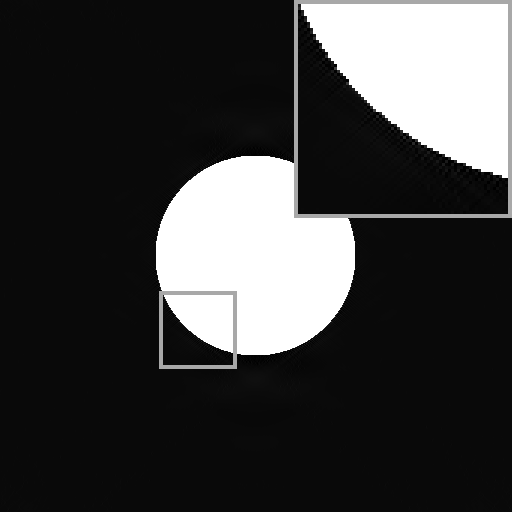}
    		\caption{}
    		\label{fig:exp_vis:4}
	    \end{subfigure}%
	    \quad
	    \begin{subfigure}[t]{0.3\textwidth}
    		\centering
    		\includegraphics[width=\textwidth]{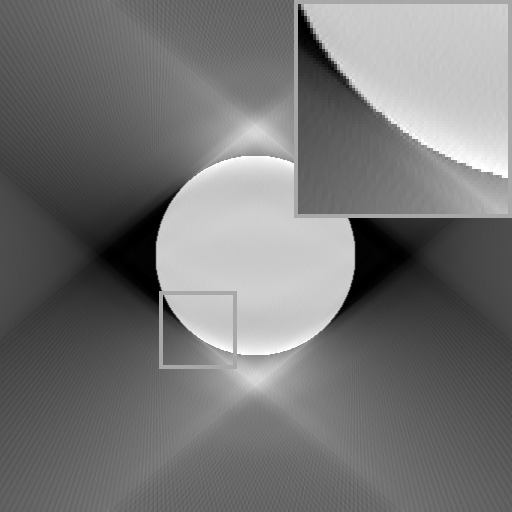}
    		\caption{}
    		\label{fig:exp_vis:5}
	    \end{subfigure}%
	    \quad
	    \begin{subfigure}[t]{0.3\textwidth}
    		\centering
    		\includegraphics[width=\textwidth]{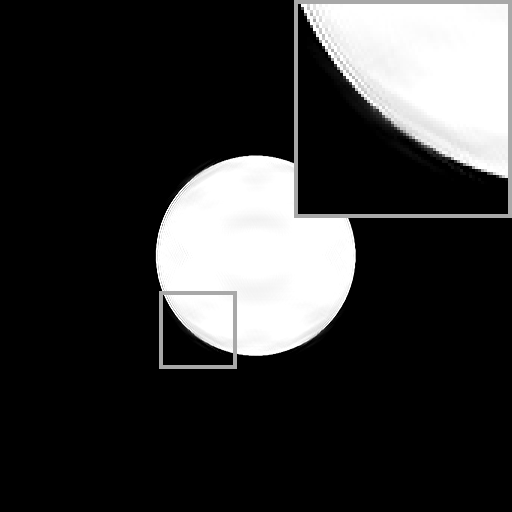}
    		\caption{}
    		\label{fig:exp_vis:6}
	    \end{subfigure}%

        \caption{A simulation visualizing the Visibility Principle \ref{princ:shear} using noisy $\Radon_{50^\circ}$ measurements, i.e.,~a missing wedge of $80^\circ$. \subref{fig:exp_vis:1} shows a simple choice for $f$ and \subref{fig:exp_vis:2} its standard FBP reconstruction. \subref{fig:exp_vis:3} displays the $\ell^1$-regularized shearlet solution  of \eqref{eq:l1-ana}.  In \subref{fig:exp_vis:4} and \subref{fig:exp_vis:5} we plot the results of the oracle replacement with perfect invisible coefficients as described in \eqref{eq:replaceInvis} and \eqref{eq:replaceFBP}, respectively. Subplot \subref{fig:exp_vis:6} shows a reconstruction, where the invisible shearlet coefficients are inferred by a neural network. Note that the dynamic range of the plots is modified for better contrast.}
        \label{fig:exp_vis}
        
    \end{figure}

In the following, we will justify the terminology of \emph{(in-)visible} shearlet indices by a simple numerical simulation. We are considering noisy Radon measurements $y = \Radon_{50^\circ}f + \eta$ of a circle $f$, which is displayed in Figure \ref{fig:exp_vis:1}\footnote{All objects of this computational example are certainly finite-dimensional, e.g., $f \in \R^{512\times 512}$. For the sake of clarity, we chose to stick to the continuous notation until introducing our digitalized reconstruction framework in the next section.}.
Figure \ref{fig:exp_vis:2} shows a standard FBP reconstruction, which suffers from strong streaking artifacts and blurry edges. Such degradations are mostly avoided in the $\ell^1$-regularized solution of \eqref{eq:l1-ana}, which is plotted in Figure \ref{fig:exp_vis:3}. Note, that the visible edges are recovered almost perfectly, however, the horizontal, invisible boundary sections cannot be retrieved.


In order to validate the concept of (in-)visible coefficients, we form the image
\begin{equation}
\label{eq:replaceInvis}
    \sh_{\gen,\phi}^{T}(\sh_{\gen,\phi}(\solu)_{\vis} + \sh_{\gen,\phi}(f)_{\inv}),
\end{equation}
i.e., the visible coefficients of the $\ell^1$-solution $\solu$ are combined with the (in practice certainly unknown) invisible coefficients of the ground truth signal $f$\footnote{For $x \in \ell^2(\Lambda)$ and a set $\mathcal{I} \subseteq \Lambda$, $x_{\mathcal{I}} \in \ell^2(\Lambda)$ shall denote the vector with $x_{\mathcal{I}}(i) = x(i)$ for $i \in \mathcal{I}$ and $x_{\mathcal{I}}(i) = 0$ otherwise.}.  From the strong resemblance of Figure~\ref{fig:exp_vis:4} with the original image, we conclude that the visible coefficients of \eqref{eq:l1-ana} resolve the visible boundary information almost perfectly, whereas the invisible coefficients do not convey any relevant information.

The replacement strategy of Equation~\eqref{eq:replaceInvis} can be interpreted as having access to an \emph{oracle} that produces invisible coefficients $\sh_{\gen,\phi}(f)_{\inv}$, given the visible ones $\sh_{\gen,\phi}(\solu)_{\vis}$ as an input. While such a perfect prediction is certainly to much to hope for, we will show in this work, that deep neural networks can be used to obtain an accurate estimation of the invisible coefficients. 
Indeed, Figure \ref{fig:exp_vis:6} shows the result obtained by combining the visible $\ell^1$-coefficients $\sh_{\gen,\phi}(\solu)_{\vis}$ with invisible coefficients that have been inferred by a CNN trained on a collection of ellipses; see Section \ref{sec:numerics} for further details.

We would like to point out that the previous visibility interpretation of the coefficients is not valid for the classical filtered backprojection solution $f_{\small \texttt{FBP}}$. The result of the analogous oracle replacement 
\begin{equation}
\label{eq:replaceFBP}
    \sh_{\gen,\phi}^{T}(\sh_{\gen,\phi}(f_{\small \texttt{FBP}})_{\vis} + \sh_{\gen,\phi}(f)_{\inv})
\end{equation}
is shown in Figure \ref{fig:exp_vis:5}. It reveals that the visible coefficients of $f_{\small \texttt{FBP}}$ are tainted with artifacts and blurry edges, making such a separation into visible and invisible coefficients impossible.

Concluding this discussion, we state the following visibility principle:

\vspace*{0.25cm}

\begin{princ}{princ:shear}
We can split the shearlet coefficients $\sh_{\gen,\phi} (\solu)$ of \eqref{eq:l1-ana} into a set of \emph{visible} and \emph{invisible} coefficients:
\begin{enumerate}
    \item The visible coefficients $\sh_{\gen,\phi} (\solu)_{\vis}$ carry reliable information about edges that are possible to reconstruct according to the visibility principle of Section \ref{sec:vis}.
    \item The invisible coefficients $\sh_{\gen,\phi} (\solu)_{\inv}$ are penalized by the $\ell^1$-norm and do not contain relevant information.
\end{enumerate}
\end{princ}

\vspace*{0.25cm}

\begin{figure}
    \centering
    \begin{subfigure}[t]{0.3\textwidth}
		\centering
		\includegraphics[trim = {4.5cm 1.5cm 4.0cm 3.0cm},clip,width=\textwidth]{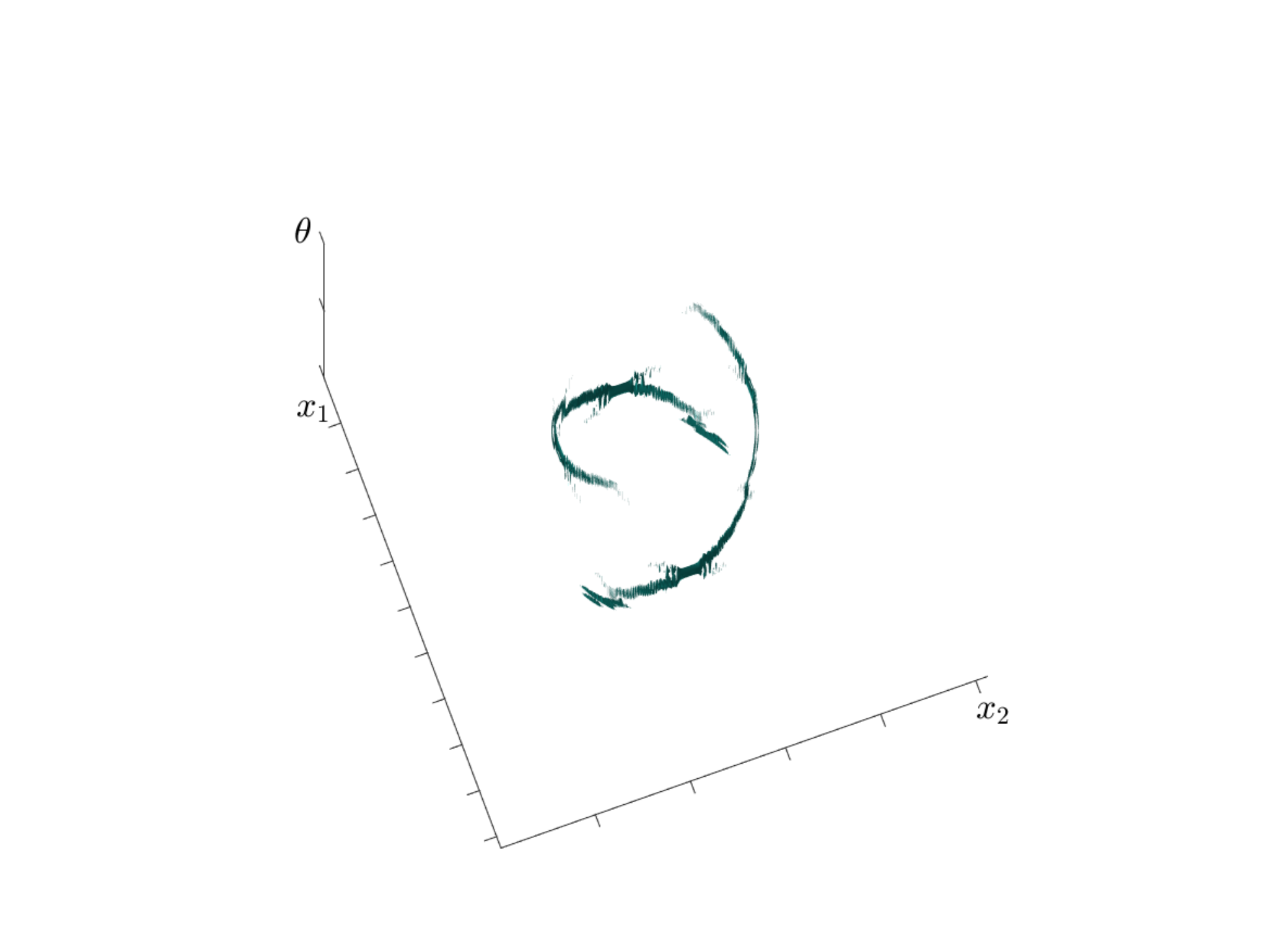}
		\caption{}
		\label{fig:candy_wrap:1}
	\end{subfigure}%
	\quad
	\begin{subfigure}[t]{0.3\textwidth}
		\centering
		\includegraphics[trim = {4.5cm 1.5cm 4.0cm 3.0cm},clip,width=\textwidth]{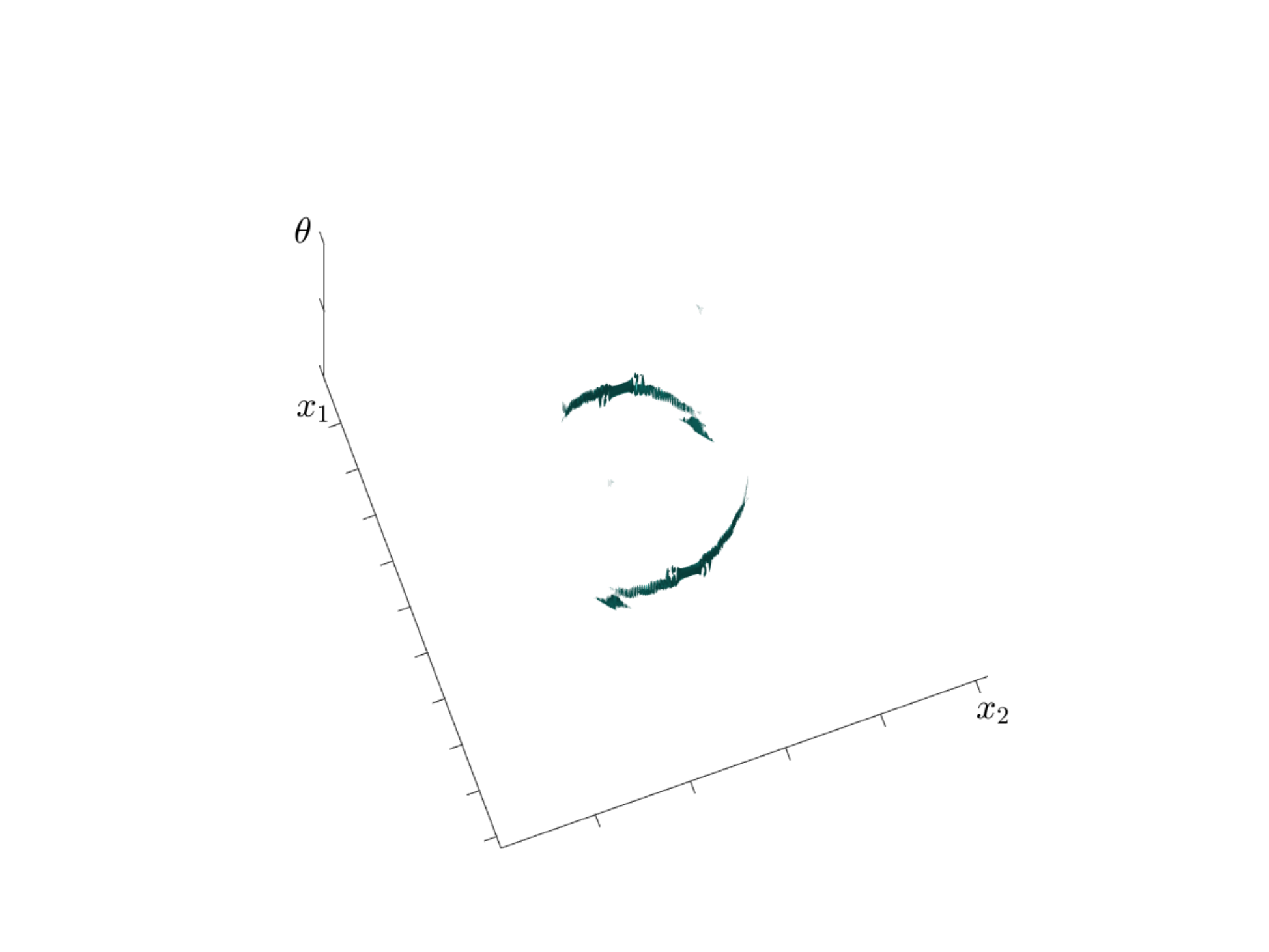}
		\caption{}
		\label{fig:candy_wrap:2}
	\end{subfigure}%
	\quad
	\begin{subfigure}[t]{0.3\textwidth}
		\centering
		\includegraphics[trim = {4.5cm 1.5cm 4.0cm 3.0cm},clip,width=\textwidth]{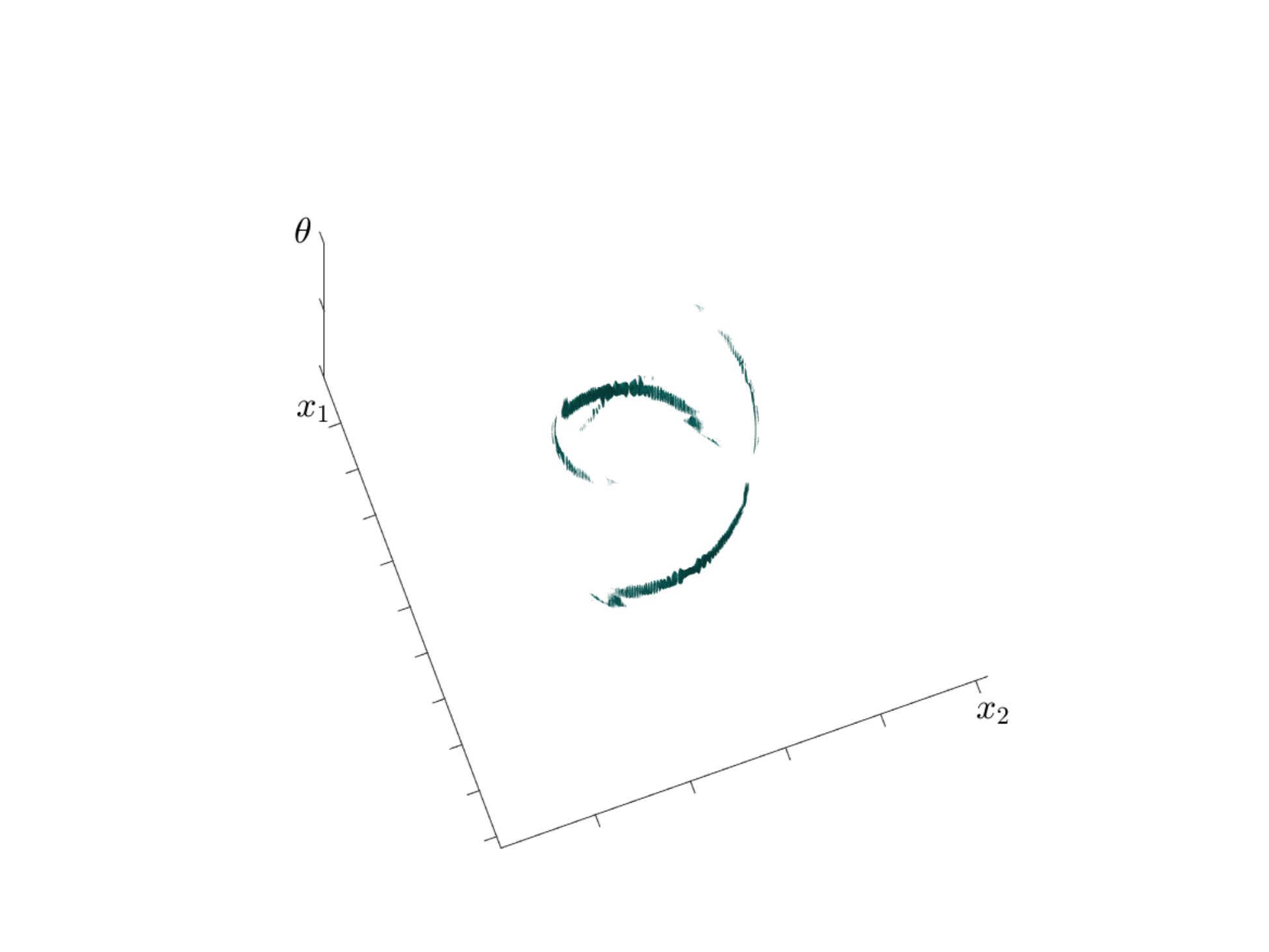}
		\caption{}
		\label{fig:candy_wrap:3}
	\end{subfigure}%
	
    \caption{ \subref{fig:candy_wrap:1} shows a stylized visualization of the shearlet coefficients of the circle in Figure \ref{fig:exp_vis:1}. \subref{fig:candy_wrap:2} displays the coefficients of the $\ell^1$-analysis solution $\solu$, revealing holes on the invisible part. \subref{fig:candy_wrap:3} shows the coefficients of the reconstruction in Figure \ref{fig:exp_vis:6}, i.e., after the invisible coefficients have been inferred by a neural network.}
    \label{fig:candy_wrap}
\end{figure}

Finally, we emphasize the close relationship between the shearlet coefficients $\sh_\gen (\solu)$ obtained by solving \eqref{eq:l1-ana} and the phase space representation of microlocal analysis:  recall that the wavefront set information can be extracted by analyzing the decay properties of the continuous shearlet transform. In terms of the discrete shearlet system a rapid decay of the continuous transform manifests as \emph{sparsity} of the associated shearlet coefficients. It is therefore quite natural to access this information via the sparsity promoting effect of $\ell^1$-minimization. As desired, the coefficients of shearlets which are not aligned with smooth directions of $f$ are \emph{``pushed to 0"} by the $\ell^1$-norm.  When sorted properly, the coefficients belonging to one particular scale are reminiscent of a discretized version of the wavefront set of $f$. In particular, they obey similar structural properties as wavefront sets in the phase space. A visualization of this observation can be found in Figure \ref{fig:candy_wrap}: a stylized plot of the finest scale coefficients of Figure \ref{fig:exp_vis}'s circle is shown in Figure~\ref{fig:candy_wrap:1}; cf.~the phase space visualization of Figure \ref{fig:wavefront:2}. As to be expected, the coefficients $\sh_\gen (\solu)$ of Figure~\ref{fig:candy_wrap:2} follow the same pattern on the visible part, however, there are holes corresponding to the invisible coefficients. Finally, Figure \ref{fig:candy_wrap:3} shows the coefficients of the deep learning based solution of Figure \ref{fig:exp_vis:6}. Indeed, the neural network seems to pick up the phase space structure and accurately estimates the invisible information.

\section{Proposed Reconstruction Method: Learning the Invisible (LtI)}
\label{sec:prop}

In this section, we define our hybrid recovery framework that makes use of an artificial neural network to learn the invisible information that cannot be retrieved from the measured data. We first introduce and discuss the general reconstruction workflow and then give more details on supervised learning of shearlet coefficients and on our particular CNN architecture.

\subsection{Algorithm}
After suitable discretization, we are given the finite-dimensional measurement vector
\begin{equation*}
    \meas = \RadonLimD \f + \noise \in \R^m,
\end{equation*}
where $\f \in \R^{n^2}$ denotes the (unknown) discrete and vectorized image, $\RadonLimD \in \R^{m  \times n^2}$ describes a discretized version of $\RadonLim$ and $\noise \in \R^m$ models the measurement noise. We propose the following recovery scheme for finding a reconstruction $\prop \in \R^{n^2}$ of $\f$:

\vspace*{0.25cm}

\begin{alg}{algo}
\begin{enumerate}
    \item[\it Step 1:] Obtain the visible coefficients $\shD (\soluD)_{\vis}$ via a nonlinear reconstruction
    \begin{equation}
    \label{eq:ana_disc}
       \soluD \in  \argmin_{\f\geq 0} \norm{\shD (\f)}_{1,\vec{w}} + \frac{1}{2} \norm{\RadonLimD \f - \meas}_2^2,
    \end{equation}
    where $\shD \in \R^{J \cdot n^2 \times n^2}$ denotes a digitalized version of $\sh_{\gen,\phi}$ with $J$ decomposition subbands.

    \item[\it Step 2:] Apply a CNN, denoted by $\NNt$, that is trained to estimate the invisible coefficients from the visible ones, i.e., determine the coefficients
    \begin{equation*}
        \vec{F} = \NNt (\shD (\soluD)) \quad (\approx \shD (\f)_{\inv}).
    \end{equation*}

    \item[\it Step 3:] Combine the visible and the learned invisible coefficients in a reliable manner by setting
    \begin{equation*}
        \prop = \shD^T(\shD (\soluD)_{\vis} + \vec{F} ).
    \end{equation*}
\end{enumerate}
\end{alg}

\vspace*{0.25cm}

A schematic workflow of the proposed method can be found in Figure \ref{fig:workflow}. For solving the optimization problem \eqref{eq:ana_disc} there is an abundance of possibilities. In this work, we are using the alternating direction method of multipliers (ADMM), as detailed in Appendix \ref{sec:solveAna}. After a discussion of our approach in the next section, we will describe the particular architecture of $\NNt$ and the procedure of learning the parameter vector $\vec{\theta}$ in the remainder of this chapter.

\subsection{Motivation and Discussion}
\label{sec:disc}
In the following, we motivate the proposed hybrid reconstruction scheme by relating it to Visibility Principle \ref{princ:shear}, and discuss some of its properties.

\subsubsection{Motivation}
The missing wedge of $\RadonLim$ results in a lack of directional information in the measured data, which is eventually responsible for artifacts and missing image features in model-based reconstruction methods. By solving the $\ell^1$-analysis minimization \eqref{eq:ana_disc}, we gain access to shearlet coefficients that correspond to reliable image features. The remaining invisible coefficients cannot be retrieved from the measured data. However, the obtained coefficient tensor $\shD (\soluD)$ somewhat resembles a discretized version of the phase space that is interspersed with holes on the invisible parts. For natural images, the visible coefficients are highly structured allowing for an inference of the invisible sections (cf.\ discussion of Section  \ref{sec:visShear} and Figure \ref{fig:wavefront}).  Thus, it appears to be a natural choice to apply machine learning techniques for the estimation in \textit{Step 2} of our proposed method.

Recently, CNNs have shown to be very effective for computer vision tasks such as image classification \cite{krizhevsky2012} or segmentation \cite{ronneberger2015u,jegou2017one}, but also in the context of inverse problems, e.g.,~\cite{burger2012,xie2012,unser2017,adler2017,adler2018,hauptmann2018}.
Based on the intuition that low-dose CT artifacts possess a directional nature, \cite{kang2017,jcy} proposed a post-processing of the FBP's directional wavelet coefficients. The subbands of the wavelet decomposition are thereby treated as different channels of the FBP image and a CNN is trained to remove their artifacts. Most, if not all, of such post-processing methods are based on so-called U-Net architectures \cite{ronneberger2015u}. For the estimation of the invisible coefficients in \textit{Step 2}, we use a modification of a similar architecture that we call \emph{\methodname{}}, see~Section \ref{sec:DLmethod}.

Our hybrid approach that combines model-based visible coefficients and learned invisible coefficients as detailed in \textit{Step 3} is accompanied by particular features that we will briefly discuss the next three paragraphs.

\subsubsection{Interpretability}


By incorporating the neural network  into a model-based approach, our proposed scheme offers a clear interpretation of its post-processing abilities in the context of limited angle CT.
 In \textit{Step 2}, the network $\NNt$ estimates the invisible coefficients from the knowledge of the previously reconstructed visible coefficients. The underlying principle of such an estimation is that the shearlet coefficients of natural images obey specific structural rules, similar to a wavefront set in the phase space. During the training over a particular class of images (cf.~Section~\ref{sec:ML}), the parameter vector $\vec{\theta}$ captures these general structural properties in the shearlet domain (cf.~Section \ref{sec:ML} and Figure \ref{fig:candy_wrap}). When applying to fresh testing data the neural network estimates the invisible coefficients according to these rules. We wish to emphasize that \textit{Step 2} can also be interpreted as a \emph{3D-inpainting problem:} the invisible parts of the reshaped shearlet coefficient tensor $\shD (\soluD) \in \R^{n \times n \times J}$ are sought to be inpainted by the neural network $\NNt$.

 Overall, the split into visible and invisible shearlet coefficients and the dedication of the CNN to solely infer the invisible ones clarifies the post-processing capabilities of neural networks. In contrast, in \cite{kang2017,jcy,unser2017}, deep learning is merely used as a black box tool for a somewhat unspecified removal of artifacts in the FBP or its coefficients.

\subsubsection{Reliability}

While our hybrid approach entrust the CNN with the transparent task of estimating invisible coefficients, it remains unclear to what extent this is actually possible. As we have argued previously, despite the success of deep learning, there is up to date no profound understanding under which assumptions on the training data, the neural network architecture, and other design choices, an accurate inference is feasible.

Our proposed hybrid reconstruction method alleviates these issues from another perspective:
by keeping the visible coefficients of the nonlinear reconstruction for the final image formation in \textit{Step 3}, we limit the influence of the neural network on the final reconstruction to a minimum. The information that we can reconstruct via the well-understood and model-based $\ell^1$-minimization of \eqref{eq:ana_disc} directly contributes to the formation of $\prop$. Only the part that is provably not contained in the measured data  is estimated by a CNN.

In particular for medical applications, it might be unsatisfactory to process an image by a CNN in order to remove its artifacts, while having no control over the applied modifications. In our approach, the impact of the \emph{``black-box CNN"} on the final reconstruction is constrained to the smallest possible extent. We believe that such an entanglement of model and data-based methods compromises between performance and reliability: the reconstructions greatly benefit from the abilities of neural networks for the estimation of the invisible part, whereas the visible boundaries are kept as reliable as possible.


\subsubsection{Performance}

While the two previous characteristics are mostly of conceptual nature, our numerical experiments reveal superior reconstruction quality when compared to other methods. In particular, we observe a remarkable generalization when our method is applied to different testing data. This is largely due to the fact that the visible coefficients are reconstructed by an advanced model-based method and therefore make for a better initialization of the input for the neural network.  Furthermore, generalization is only of relevance on the invisible part of the wavefront set, since the final image formation of \textit{Step 3} is based on the model-based reconstruction of the visible part.   

When a CNN is used for post-processing an FBP reconstruction \cite{unser2017,kang2017,jcy}, a lot of its \emph{expressiveness} is needed for removing the streaking artifacts. This is particularly important for limited angle CT, where depending on the size of the missing wedge, the streaking artifacts are severe. In contrast, an initialization with the nonlinear reconstruction of \eqref{eq:ana_disc} is contaminated far less with unwanted artifacts and the network is allowed to focus on learning the invisible edge information. Additionally, it is well known that $\ell^1$-minimization leads to sharper boundaries when compared to FBP images. This effect might be amplified by processing with a U-Net architecture \cite{han2018}. However, it is completely avoided on the visible part by the combination proposed in \textit{Step 3}.


\begin{figure} 
\begin{center}
\scalebox{0.65}{%
\makebox[0pt][c]{%
\begin{tabular}{@{}X@{\quad}Y@{\qquad}S@{\qquad}L@{\qquad}S@{\qquad}Y@{\hspace{-1cm}}Y@{}}
\hspace{1.0cm} $\meas$
	& \includegraphics[scale=0.35]{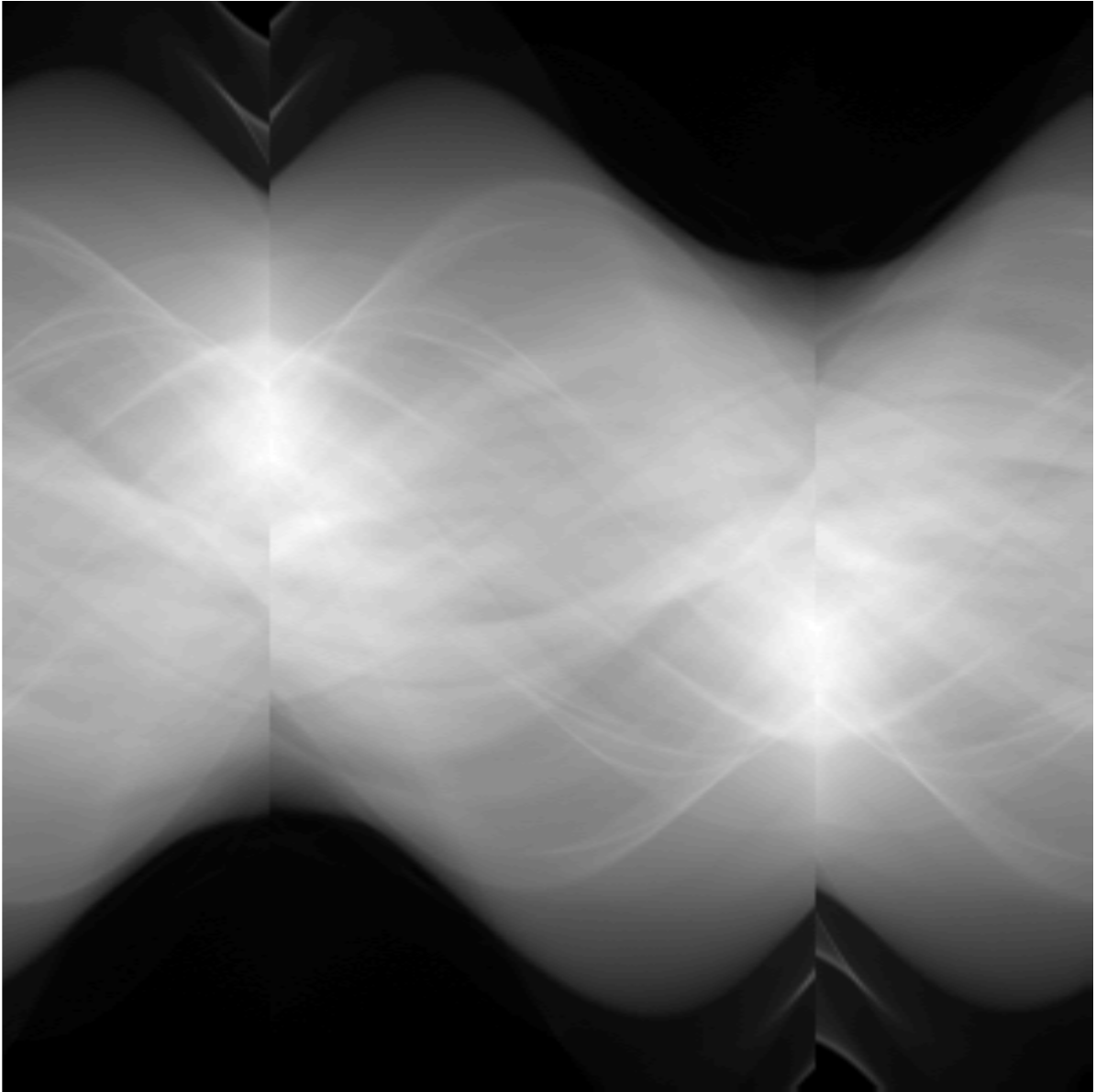}
	&
	& \vspace{3.0cm} $\prop$
	& \begin{tikzpicture}
	\draw[white] (0.5,1.5) -- (0.5,4);
	\draw[<-] (0.65,0.25) -- (1.75,1);
	\draw (0.75,1) node{$\shD^T$};
	\end{tikzpicture}
	& \tdplotsetmaincoords{70}{20}
	\begin{tikzpicture}
	[tdplot_main_coords,
	cube/.style={very thin, black}]
	
	\draw[cube] (0,0,0) -- (0,3,0) -- (3,3,0) -- (3,0,0) -- cycle;
	\draw[cube] (0,0,3) -- (0,3,3) -- (3,3,3) -- (3,0,3) -- cycle;	
	\draw[cube] (0,0,0) -- (0,0,3);
	\draw[cube,dashed] (0,3,0) -- (0,3,3);
	\draw[cube] (3,0,0) -- (3,0,3);
	\draw[cube] (3,3,0) -- (3,3,3);

	\draw[cube] (0,0,0.5) -- (0,3,0.5) -- (3,3,0.5) -- (3,0,0.5) -- cycle;
	\draw[cube] (0,0,1.25) -- (0,3,1.25) -- (3,3,1.25) -- (3,0,1.25) -- cycle;	
	
	\fill[cube,black,fill opacity = 0.1] (0,0,0) -- (0,3,0) -- (3,3,0) -- (3,0,0);
	\fill[cube,black,fill opacity = 0.1] (0,0,3) -- (0,3,3) -- (3,3,3) -- (3,0,3);
	\fill[cube,black,fill opacity = 0.1] (0,0,0) -- (0,3,0) -- (0,3,0.5) -- (0,0,0.5);
	\fill[cube,black,fill opacity = 0.1] (0,3,0) -- (3,3,0) -- (3,3,0.5) -- (0,3,0.5);
	\fill[cube,black,fill opacity = 0.1] (3,3,0) -- (3,0,0) -- (3,0,0.5) -- (3,3,0.5);
	\fill[cube,black,fill opacity = 0.1] (3,0,0) -- (0,0,0) -- (0,0,0.5) --(3,0,0.5);
	
	\fill[cube,black,fill opacity = 0.1] (0,0,1.25) -- (0,3,1.25) -- (0,3,3) -- (0,0,3);
	\fill[cube,black,fill opacity = 0.1] (0,3,1.25) -- (3,3,1.25) -- (3,3,3) -- (0,3,3);
	\fill[cube,black,fill opacity = 0.1] (3,3,1.25) -- (3,0,1.25) -- (3,0,3) -- (3,3,3);
	\fill[cube,black,fill opacity = 0.1] (3,0,1.25) -- (0,0,1.25) -- (0,0,3) --(3,0,3);

	\fill[cube,red,fill opacity = 0.5] (0,0,0.5) -- (0,3,0.5) -- (3,3,0.5) -- (3,0,0.5);
	\fill[cube,red,fill opacity = 0.5] (0,0,1.25) -- (0,3,1.25) -- (3,3,1.25) -- (3,0,1.25);	
	\fill[cube,red,fill opacity = 0.5] (0,0,0.5) -- (0,3,0.5) -- (0,3,1.25) -- (0,0,1.25);
	\fill[cube,red,fill opacity = 0.5] (0,3,0.5) -- (3,3,0.5) -- (3,3,1.25) -- (0,3,1.25);
	\fill[cube,red,fill opacity = 0.5] (3,3,0.5) -- (3,0,0.5) -- (3,0,1.25) -- (3,3,1.25);
	\fill[cube,red,fill opacity = 0.5] (3,0,0.5) -- (0,0,0.5) -- (0,0,1.25) --(3,0,1.25);

	\draw (1,2.5,-1.5) node{$\shD (\soluD)_{\vis} + \vec{F}$};
	\end{tikzpicture} & \\[-1.5em]
& \begin{tikzpicture}
	\draw[<-] (0,0) -- (0,2.5);
	\draw (-1.5,2.25) node{\emph{Step 1}};
	\draw (-1.5,1.75) node{nonlin.~rec.};
	\draw (-1.5,1.25) node{of vis.~coeff.};
	\draw (-1.5,0.7) node{via \eqref{eq:ana_disc}};
	\draw[white] (0.5,1) -- (1.75,1);
	\end{tikzpicture}
	&
	&  \includegraphics[scale=0.24]{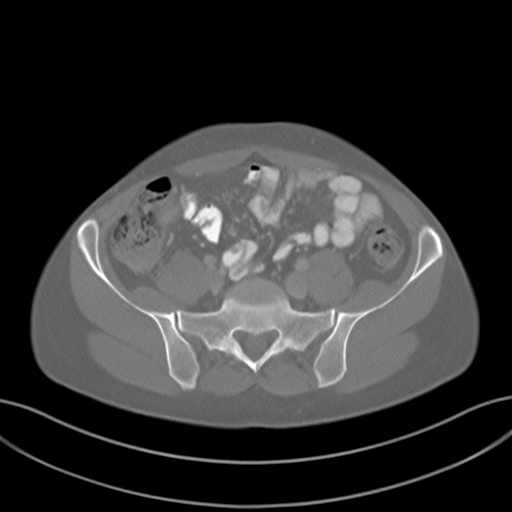}
	&
	& \begin{tikzpicture}
	\draw[->] (0,0) -- (0,2.5);
	\draw (1.5,1.25) node{combine};
	\draw (1.5,1.75) node{\textit{Step 3}};
	\draw (1.5,0.75) node{both parts};
	\draw (1.5,0.25) node{$\Sigma$};
	\end{tikzpicture} & \\[-1em]
& \tdplotsetmaincoords{70}{20}
\begin{tikzpicture}
[tdplot_main_coords,
	cube/.style={very thin, black}]
	
\draw[cube] (0,0,0) -- (0,3,0) -- (3,3,0) -- (3,0,0) -- cycle;
\draw[cube] (0,0,3) -- (0,3,3) -- (3,3,3) -- (3,0,3) -- cycle;	
\draw[cube] (0,0,0) -- (0,0,3);
\draw[cube,dashed] (0,3,0) -- (0,3,3);
\draw[cube] (3,0,0) -- (3,0,3);
\draw[cube] (3,3,0) -- (3,3,3);

\draw[cube] (0,0,0.5) -- (0,3,0.5) -- (3,3,0.5) -- (3,0,0.5) -- cycle;
\draw[cube] (0,0,1.25) -- (0,3,1.25) -- (3,3,1.25) -- (3,0,1.25) -- cycle;	

\fill[cube,black,fill opacity = 0.1] (0,0,0) -- (0,3,0) -- (3,3,0) -- (3,0,0);
	\fill[cube,black,fill opacity = 0.1] (0,0,3) -- (0,3,3) -- (3,3,3) -- (3,0,3);
	\fill[cube,black,fill opacity = 0.1] (0,0,0) -- (0,3,0) -- (0,3,0.5) -- (0,0,0.5);
	\fill[cube,black,fill opacity = 0.1] (0,3,0) -- (3,3,0) -- (3,3,0.5) -- (0,3,0.5);
	\fill[cube,black,fill opacity = 0.1] (3,3,0) -- (3,0,0) -- (3,0,0.5) -- (3,3,0.5);
	\fill[cube,black,fill opacity = 0.1] (3,0,0) -- (0,0,0) -- (0,0,0.5) --(3,0,0.5);
	
	\fill[cube,black,fill opacity = 0.1] (0,0,1.25) -- (0,3,1.25) -- (0,3,3) -- (0,0,3);
	\fill[cube,black,fill opacity = 0.1] (0,3,1.25) -- (3,3,1.25) -- (3,3,3) -- (0,3,3);
	\fill[cube,black,fill opacity = 0.1] (3,3,1.25) -- (3,0,1.25) -- (3,0,3) -- (3,3,3);
	\fill[cube,black,fill opacity = 0.1] (3,0,1.25) -- (0,0,1.25) -- (0,0,3) --(3,0,3);

\fill[cube,darkblue,fill opacity = 0.5] (0,0,0.5) -- (0,3,0.5) -- (3,3,0.5) -- (3,0,0.5);
\fill[cube,darkblue,fill opacity = 0.5] (0,0,1.25) -- (0,3,1.25) -- (3,3,1.25) -- (3,0,1.25);	
\fill[cube,darkblue,fill opacity = 0.5] (0,0,0.5) -- (0,3,0.5) -- (0,3,1.25) -- (0,0,1.25);
\fill[cube,darkblue,fill opacity = 0.5] (0,3,0.5) -- (3,3,0.5) -- (3,3,1.25) -- (0,3,1.25);
\fill[cube,darkblue,fill opacity = 0.5] (3,3,0.5) -- (3,0,0.5) -- (3,0,1.25) -- (3,3,1.25);
\fill[cube,darkblue,fill opacity = 0.5] (3,0,0.5) -- (0,0,0.5) -- (0,0,1.25) -- (3,0,1.25);

\draw (1,2.5,-1.5) node{$\shD (\soluD)$};
\end{tikzpicture} 	
	& \begin{tikzpicture}
	\draw[->] (0.5,1) -- (2,1);
	\end{tikzpicture} 	
	& \begin{tikzpicture}
	\draw (0,1.4) ellipse (2cm and 1.25cm);
	\draw (0,2.1) node{\emph{Step 2}};
	\draw (0,1.4) node{\methodname{} $\NNt$};
	\draw (0,0.8) node{learn the invisible};
	\end{tikzpicture}
	& \begin{tikzpicture}
	\draw[->,white] (0.5,1) -- (1.0,1);
	\draw[->] (1.0,1) -- (2.5,1);
	\end{tikzpicture}
	& \tdplotsetmaincoords{70}{20}
\begin{tikzpicture}
[tdplot_main_coords,
	cube/.style={very thin, black}]

\draw[cube] (0,0,0.5) -- (0,3,0.5) -- (3,3,0.5) -- (3,0,0.5) -- cycle;
\draw[cube] (0,0,1.25) -- (0,3,1.25) -- (3,3,1.25) -- (3,0,1.25) -- cycle;	

\draw[dashed] (0,3,0.5) -- (0,3,1.25);
\draw (3,3,0.5) -- (3,3,1.25);
\draw (3,0,0.5) -- (3,0,1.25);
\draw (0,0,0.5) -- (0,0,1.25);

\fill[cube,red,fill opacity = 0.5] (0,0,0.5) -- (0,3,0.5) -- (3,3,0.5) -- (3,0,0.5);
\fill[cube,red,fill opacity = 0.5] (0,0,1.25) -- (0,3,1.25) -- (3,3,1.25) -- (3,0,1.25);	
\fill[cube,red,fill opacity = 0.5] (0,0,0.5) -- (0,3,0.5) -- (0,3,1.25) -- (0,0,1.25);
\fill[cube,red,fill opacity = 0.5] (0,3,0.5) -- (3,3,0.5) -- (3,3,1.25) -- (0,3,1.25);
\fill[cube,red,fill opacity = 0.5] (3,3,0.5) -- (3,0,0.5) -- (3,0,1.25) -- (3,3,1.25);
\fill[cube,red,fill opacity = 0.5] (3,0,0.5) -- (0,0,0.5) -- (0,0,1.25) --(3,0,1.25);

\draw (1,2.5,-1.25) node{$\vec{F}$};
\end{tikzpicture} & \\[-3.5em]
& & & & & & \begin{tikzpicture}
\fill[black,fill opacity = 0.15] (0.75,0) rectangle (0.5,0.25);
\fill[darkblue,fill opacity = 0.5] (0.75,-0.5) rectangle (0.5,-0.25);
\fill[red,fill opacity = 0.5] (0.75,-1) rectangle (0.5,-0.75);

\draw[black,anchor=west] (1,0.15) node{visible coeff. $\vis$};
\draw[darkblue,anchor=west] (1,-0.35) node{invisible coeff. $\inv$};
\draw[red,anchor=west] (1,-0.85) node{learned coeff.};
\end{tikzpicture} 
\end{tabular}}}
\end{center}
\vspace{-4.0em}

\scalebox{0.8}{%
\makebox[0pt][c]{%
\begin{tikzpicture}
\draw[ultra thick,dashed,gray] (-10,-7.8) rectangle (9,0);.5
\draw (4.15,0) to  [bend left=25] (0.1,2.5);
\draw (-6.25,0) to  [bend left=-25] (-2.0,2.5);

\fill[aquamarine] (-9.5,-2.25) rectangle (-7.5,-0.25);
\fill[gray, fill opacity=0.5] (-7.3,-2.25) rectangle (-5.3,-0.25);
\draw[->,thick,dashed] (-5,-1.25) -- (4,-1.25);
\fill[aquamarine] (6.5,-0.25) rectangle (8.5,-2.25);
\fill[gray, fill opacity=0.5] (4.3,-0.25) rectangle (6.3,-2.25); 

\fill[amber] (-7.5,-4) rectangle (-6.5,-3);
\fill[gray, fill opacity=0.5] (-6.3,-4) rectangle (-5.3,-3);
\draw[->,thick,dashed] (-5,-3.5) -- (4,-3.5);
\fill[purple] (5.5,-3) rectangle (6.5,-4);
\fill[gray, fill opacity=0.5] (4.3,-3) rectangle (5.3,-4);

\fill[amber] (-5,-5) rectangle (-4.5,-5.5);
\fill[gray, fill opacity=0.5] (-4.3,-5) rectangle (-3.8,-5.5);
\draw[->,thick,dashed] (-3.5,-5.25) -- (2.5,-5.25);
\fill[purple] (3.5,-5.5) rectangle (4.0,-5);
\fill[gray, fill opacity=0.5] (2.8,-5.5) rectangle (3.3,-5);

\fill[amber] (-0.95,-6.5) rectangle (-0.7,-6.25);
\fill[gray, fill opacity=0.5] (-0.5,-6.5) rectangle (-0.25,-6.25);
\fill[purple] (-0.05,-6.5) rectangle (0.2,-6.25);

\draw (-8.5,-2.5) node{($512\times 512, 64$)};
\draw (-6.8,-4.3) node{($256 \times 256, 128$)};
\draw (-5.1,-5.8) node{($128 \times128, 256$)};
\draw (-2.0,-6.8) node{($64 \times64, 512$)};

\draw (-3.9,-0.5) node{($512 \times 512, 128$)};
\draw (-3.9,-3.2) node{($256 \times 256, 256$)};
\draw (-2.4,-5.0) node{($128 \times128, 512$)};
\draw (-0.52,-5.95) node{($64 \times 64, 512$)};

\draw (1.4,-6.8) node{($128 \times 128, 512$)};
\draw (3.6,-5.8) node{($256 \times $};
\draw (3.6,-6.2) node{$256, 256$)};
\draw (5.8,-4.3) node{($512 \times 512, 128$)};
\draw (7.5,-2.5) node{($512 \times 512, 59$)};

\draw (1.4,-5.0) node{($128 \times 128, 256$)};
\draw (2.9,-3.2) node{($256 \times 256, 128$)};
\draw (2.9,-0.5) node{($512 \times 512, 64)$};

\draw[->,thick] (-7,-2.4) -- (-6.7,-2.9);
\draw[->,thick] (5.6,-2.9) -- (5.9,-2.4);

\draw[->,thick] (-5.5,-4.2) -- (-4.8,-4.9);
\draw[->,thick]  (3.8,-4.9) -- (4.5,-4.2);

\draw[->,thick] (-3.8,-5.6) -- (-1.1,-6.4);
\draw[->,thick]  (0.3,-6.4) -- (2.8,-5.7);

\fill[aquamarine] (4.75,-5.5) rectangle (4.5,-5.25);
\fill[gray, fill opacity=0.5] (4.75,-6) rectangle (4.5,-5.75);
\fill[amber] (4.75,-6.5) rectangle (4.5,-6.25);
\fill[purple] (4.75,-7) rectangle (4.5,-6.75);
\draw[->,dashed]  (4.5,-7.35) -- (4.75,-7.35);

\draw[aquamarine] (6,-5.35) node{{\footnotesize Convolution}};
\draw[gray] (6.65,-5.85) node{{\footnotesize Trimmed-DenseBlock}};
\draw[amber] (6.3,-6.35) node{{\footnotesize Transition Down}};
\draw[purple] (6.1,-6.85) node{{\footnotesize Transition Up}};
\draw (6.7,-7.35) node{{\footnotesize Copy and Concatenate}};
\end{tikzpicture}}}
\caption{A schematic workflow of the proposed reconstruction framework \emph{LtI} (see~Algorithm \ref{algo}), which learns the invisible shearlet coefficients for limited angle tomography. The lower part depicts the architecture \methodname{}. The output shape of each layer is denoted by (size $\times$ size, channels). The input to \methodname{} is of shape ($512 \times 512, 59$). We choose $n=4$ layers in each TDB, except for the center TDB where $n=8$. Thus, the overall number of layers is $6 \times 4 + 8 + 8  = 40$.} 
    \label{fig:workflow}
\end{figure}
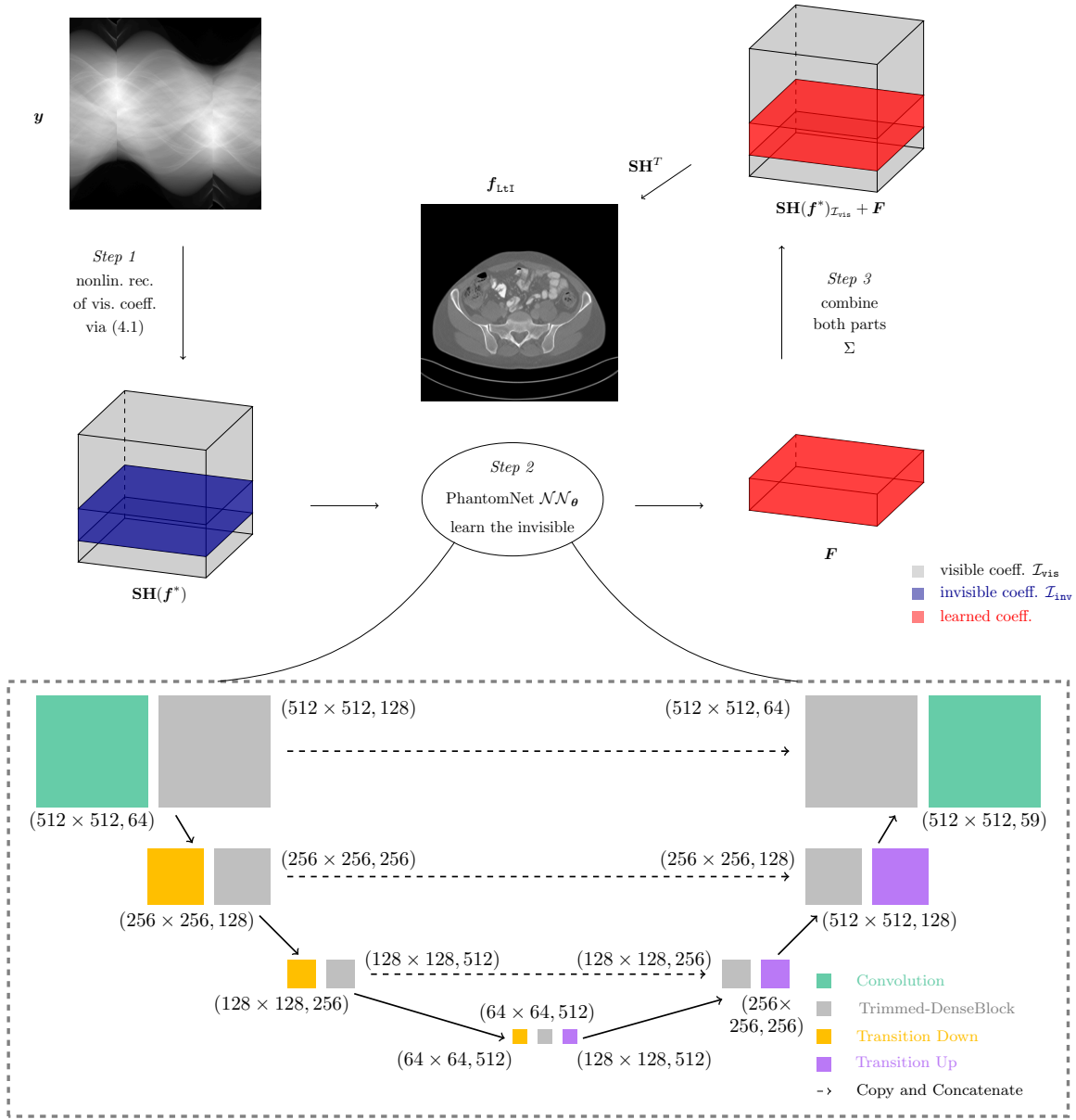



\subsection{Learning the Invisible with PhantomNet}

\enlargethispage{0.4cm}
Given the visible shearlet coefficients of the nonlinear reconstruction, we are now briefly describing a machine learning framework for the estimation of the invisible coefficients by means of a CNN.
The overall goal is to find a (non-linear) mapping ${\NNt : \R^{n \times n \times J} \rightarrow \R^{n \times n \times J}}$, parametrized by a (high-dimensional) vector $\vec{\theta}$, that ideally satisfies the relation
\begin{equation*}
    \NNt (\shD (\soluD)) \approx \shD (\f)_{\inv}.
\end{equation*}
We first give a short introduction to the general statistical learning framework and then describe the particular CNN architecture PhantomNet that we use for our experiments in Section \ref{sec:numerics}.

\subsubsection{(Supervised) Learning of Invisible Coefficients}
\label{sec:ML}
For a mathematical formalization of learning invisible coefficients, we regard the tuple $(\f, \soluD) \in \R^{n^2} \times \R^{n^2}$ as a random variable with a  joint probability distribution $p$. 
Ideally, we would like to find a parameter vector $\vec{\theta}$ that allows for an estimation of the invisible coefficients with respect to $p$. This could for instance be achieved by minimizing the expected risk
\begin{equation}
\label{eq:DLobj}
    \min_{\vec{\theta}} \left( \mathbb{E}_{(\f,\soluD) \sim p} \norm{\NNt (\shD (\soluD)) - \shD (\f)_{\inv}}_{\vec{w},2}^2 \right),
\end{equation}
where $\vec{w} \in \R^{\#\inv}$ is a vector of weights, accounting for instance for the fact that shearlet coefficients come in different orders of magnitude depending on their scale.
Note that the $\ell^2$-loss is only computed on the invisible coefficients that are sought to be learned. In principle, any other loss function or an additional regularization term, such as the sparsity promoting $\ell^1$-norm of $\NNt (\shD (\soluD))$, could be beneficial. For the sake of brevity, we will stick to the basic form of \eqref{eq:DLobj}.

In practice, computing the expectation with respect to $p$ is not possible. Instead, we are typically given a finite set of independent drawings $(\f_1,\soluD_1),\dots,(\f_N,\soluD_N)$ and consider the minimization of the empirical risk
\begin{equation}
\label{eq:DLobjective}
    \min_{\vec{\theta}} \frac{1}{N} \sum_{j=1}^N \norm{\NNt (\shD (\soluD_j)) - \shD (\f_j)_{\inv}}_{\vec{w},2}^2.
\end{equation}
Depending on the properties of $\NNt$, the optimization problem is in general non-convex. In the case of neural networks, typically some form of gradient descent is used, where the gradients are calculated via \emph{backpropagation} \cite{rumelhart1986}. Computing the gradient for the sum over the entire training set in \eqref{eq:DLobjective} is often not feasible for large-scale problems due to memory limitations. To circumvent this problem, \emph{stochastic} or \emph{minibatch gradient descent} is used, in which the gradient is approximated over smaller, randomly selected batches of training examples \cite[Chpt. 8]{Goodfellow2016}.

The final performance (i.e.,~the \emph{generalization})  of the trained neural network $\NNt$ is evaluated on a separate set of independent drawings, the so-called \emph{test set}, that were not previously used for the optimization of $\vec{\theta}$ in \eqref{eq:DLobjective}. 


\subsubsection{Convolutional Neural Networks}
The main building blocks of CNNs are convolutional layers of the following form: let $\vec{I} = [\vec{I}_1,\dots,\vec{I}_{c_1}] \in \R^{k\times k \times c_1}$ be an input array, where $k \in \N$ is referred to as the \emph{spatial dimension} and $c_1 \in \N$ as the number of input \emph{channels}. For a desired number of output channels $c_2\in \N$ and $i \in \left\{1,\dots,c_2\right\}$ let $\vec{w}^i = [\vec{w}^i_1,\dots,\vec{w}^i_{c_1}] \in \R^{s \times s \times c_1}$ denote a convolutional filter with kernel size $s \in \N$ and $b_i \in \R$ a bias. Then, the $i$-th output channel of the convolutional layer is given by
\begin{equation}
    \label{eq:conv}
    \vec{o}_i = \sigma \left( \sum_{j=1}^{c_1} \vec{I}_j \ast \vec{w}^i_{j}  + b_i \right),
\end{equation}
where $\ast$ denotes a 2D-convolution and $\sigma : \R \rightarrow \R$ is a non-linear (\emph{activation}) function. With an abuse of notation, the application of $\sigma$ and the addition is thereby applied elementwise.

State of the art CNNs typically consist of dozens or hundreds of concatenated convolutional layers which are regularly alternated with \emph{pooling}  layers \cite[Chpt.~9.3]{Goodfellow2016}. Their main features are a relatively low number of parameters (when compared with fully connected NNs) and the translation invariance due to the convolutional structure. The set of all free parameters, i.e., the convolutional weights and biases of all layers, are collected in the parameter vector $\vec{\theta}$.  There is a plethora of variations and extensions of this basic definition and we refer the interested reader to \cite{Goodfellow2016} for more details on this subject. A precise description of the \emph{PhantomNet}-architecture used for our experiments will be given in the following section.

\subsubsection{\methodname{}}

\label{sec:DLmethod}

Our architecture,  which we refer to as \textit{\methodname{}}, is largely based on U-Net - a CNN that was introduced in \cite{ronneberger2015u} for biomedical image segmentation. In general, U-Nets  consist of an encoder and a decoder similar to autoencoders  \cite[Chpt.~14]{Goodfellow2016}. The encoder takes the input and maps it to a latent compressed representation while the decoder up-samples towards the output. The U-Net architecture enables passing the high resolution feature maps directly from the encoder to the decoder. For the concatenation of the feature maps in the decoder, the architecture is kept symmetric with respect to the encoder, such that the overall network architecture resembles a 'U'; cf.~Figure \ref{fig:workflow}. 



\newpage
\noindent The encoder and the decoder of \methodname{} consist of the following \textbf{building blocks:}
\begin{itemize}
	\item Trimmed-DenseBlocks (TDB):
	the defining feature of \methodname{} are its modificated \emph{DenseBlocks}. DenseBlocks were introduced in \cite{huang2017densely} as densely connected groups of layers and can be seen as an extension of the popular \emph{residual networks (ResNets)} \cite{he2016deep}. A residual block bypasses the non-linear transformation by an identity function, i.e.,
    \begin{equation}
    \vec{x}_{n+1} = H_{n+1}(\vec{x}_{n}) + \vec{x}_{n},
    \end{equation}
    where $H_{n+1}$ represents the non-linear transformation \eqref{eq:conv} at the $n+1$-th layer and $\vec{x}_n$ denotes the output of the $n$-th layer. This shortcut connection enforces the layer to learn something new compared to the input. Additionally, the residual connections help for a faster convergence during the optimization. \cite{veit2016residual} shows that the ResNets address the problem of vanishing gradients in very deep networks by using short paths.

	DenseBlocks (DB) generalize this idea in the sense that their layers have connections from all previous layers - i.e., the input to the current layer consists of the feature maps of all the layers in the DenseBlock before it:
    \begin{equation}
    \vec{x}_{n+1} = H_{n+1}([\vec{x}_{0}, \vec{x}_{1}, ...,\vec{x}_{n}]),
\end{equation}
    where $[\cdot]$ denotes the concatenation of the arrays $\vec{x}_i$.
Each layer in the block consists of a $3 \times 3$ convolution with a stride of $1$ followed by a \emph{hyperbolic tangent (tanh)} non-linear activation $\varphi$. The outputs of the layer are $c_2$ feature maps, which is defined as the \emph{growth rate}. The output of the DB is a concatenation of the output of all the layers in it (of size $n \times c_2$), concatenated with the input $\vec{x}_0$. TDBs are DBs, in which the input feature maps to a block, i.e. $\vec{x}_0$, are not passed to the output of the block. In each TDB, we choose $n \in \left\{ 4,8\right\}$ layers and a growth rate $c_2 \in \left\{16,32, 64, 128\right\}$, such that there are fewer feature maps in the initial blocks and more of them towards the latent representation; cf.~Figure \ref{fig:workflow}. 

    \item TransitionDown (TD): the encoder of the \methodname{} compresses the input to a latent representation. To enable the contracting path, the TransitionDown consists of a $3 \times 3$ convolution with a stride of $1$ followed by \textit{max pooling} operation with a stride of $2$. This brings down the resolution of the feature maps by a factor of $2$.
    \item TransitionUp (TU): the decoder upscales the latent representation using the transpose of a $3 \times 3$ convolution with a stride of $2$.
\end{itemize}

\methodname{} consists of 4 TDBs on the encoder and 3 TDBs on the decoder, as shown in Figure \ref{fig:workflow}. Feature maps of the first 3 TDBs on the encoder are passed through the skip connections and concatenated with the respective feature maps before given as input to the respective TDB at the decoder. Following the first 3 TDBs are TDs which bring down the resolution of the feature maps by a factor of $2$. On the decoder, there are TU blocks after the TDB blocks, which make use of the convolution transpose operation to upscale the resolution of the feature maps by a factor of $2$.

The \methodname{}  architecture is inspired from \cite{jegou2017one}, albeit with differences. While \cite{jegou2017one} uses DenseBlocks with batch normalization and rectified linear units as non linear activations, \methodname{} makes use of Trimmed-DenseBlocks with just \textit{tanh} activation. The main difference between TDBs and DBs is that in TDBs the input feature maps to a block are not passed to the output of the block. Since the output feature maps from a TDB are passed through a skip connection to the decoder, there is a considerable reduction in the size of the network which is beneficial for computational expense. The residual connections in the block already force each layer to learn useful representations. The trimming of the input feature maps to the output does no harm and is found to work better for problems where sparse information has to be learned.

\methodname{} takes the entire stack of shearlet coefficients $\shD (\soluD)$ as an input tensor. Similar to \cite{kang2017,jcy}, the subbands of $\shD (\soluD)$ (corresponding to directional features at different scales) are thereby treated as different input channels. Note that \methodname{} is a \emph{fully convolutional neural network} \cite{long2015}. This means that, in contrast to classification neural networks, the last layer is not a fully connected layer but also convolutional. Therefore, \methodname{} is able to process inputs of arbitrary spatial dimension, which will be used during the training stage.



\section{Experiments and Results}
\label{sec:numerics}

In this section, we evaluate the performance of the proposed reconstruction scheme by comparing with classical and learning based reconstruction methods.  For thorough testing we consider a combination of different measurement setups and different types of simulated and measured data.

\subsection{Preliminaries}

Let us begin by describing the considered experimental scenarios and giving details on the implementation of the used operators, the training procedure and the methods that we compare our results with.

\subsubsection{Experimental Scenarios}

\begin{figure}
    \centering
    \small
            \begin{subfigure}[t]{0.3\textwidth}
    		\centering
    		\includegraphics[height=\textwidth]{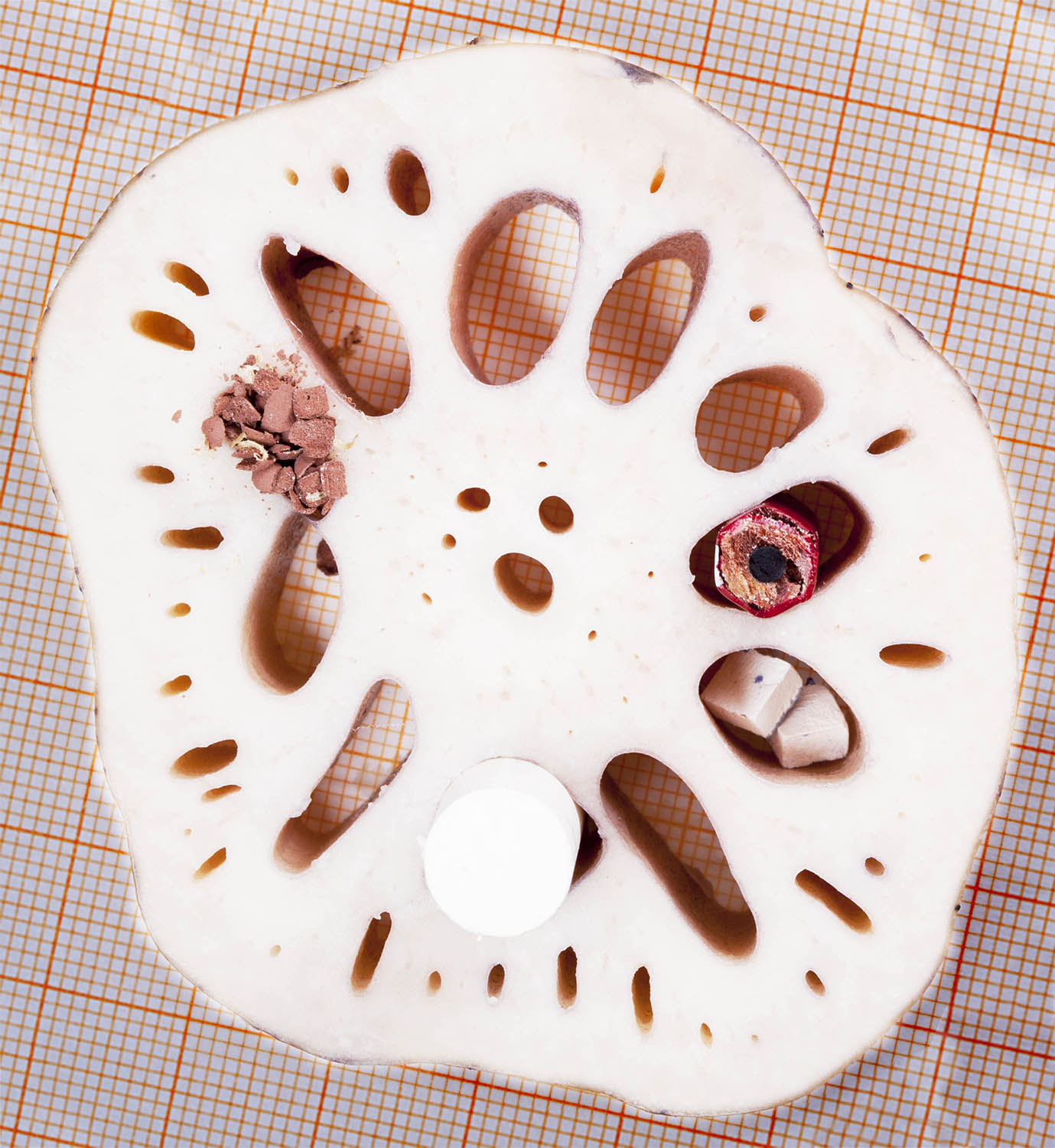}
    		\caption{\centering }
    		\label{fig:gt_lotus:1}
	    \end{subfigure}%
	    \quad
	    \begin{subfigure}[t]{0.3\textwidth}
    		\centering
    		\includegraphics[height=\textwidth]{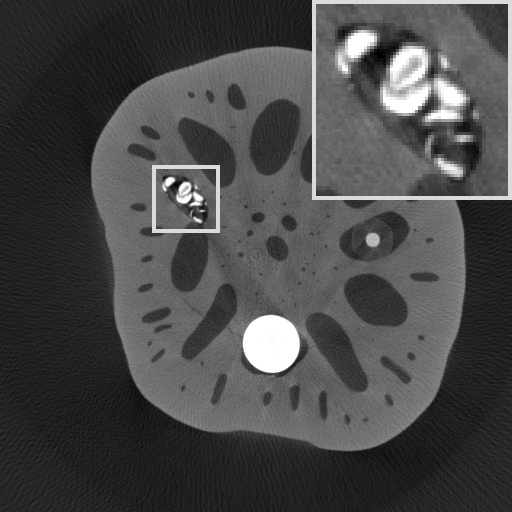}
    		\caption{\centering }
    		\label{fig:gt_lotus:2}
	    \end{subfigure}%
        \caption{The image  \subref{fig:gt_lotus:1} shows a photography of the signal used in the experiments \lotus{} and \Lotus{}. In \subref{fig:gt_lotus:2}, a reference scan with full angular measurements is displayed, in which the plotting window is slightly adapted for better contrast.}
        \label{fig:gt_lotus}
\end{figure}

We consider three different types of data for evaluating our proposed method:
\begin{enumerate}
    \item The \emph{ellipsoid dataset} consists of 2000 synthetic images of ellipses, where the number, locations, sizes and the intensity gradients of the ellipses are chosen at random. Using the Matlab function \texttt{radon}, we simulate noisy measurements for a missing wedge of $80^\circ$. To avoid an \emph{inverse crime} \cite{mueller2012} the measurements are simulated at a higher resolution and then downsampled for an image resolution of $512 \times 512$.  1600 images are used for training, 200 images for validation and 200 for testing. This experimental setup is referred to as Experiment \ellip.
    \item In a more realistic setup, we are working on human abdomen scans provided by the \emph{Mayo Clinic} for the  AAPM Low-Dose CT Grand Challenge \cite{mayo}. The data consists of 10 patients resulting in 2378 images of size $512 \times 512$ with a slice thickness of 3mm. We use 9 patients for training (2134 slices) and 1 patient for testing (244 slices). We remark that this is not completely consistent with the setup used in \cite{jcy}, where 1 training patient was used for validation (330 slices). Noisy measurements are simulated with Astra \cite{astra} using a \emph{fanbeam} geometry corresponding to missing wedges of $60^\circ$ and $30^\circ$.  We will refer to these scenarios as \mayo{} and \Mayo, respectively.
    \item For testing the generalization properties of our method, we furthermore make use of real data from a scan of a \emph{lotus root} \cite{bubba2016}.  This setup is referred to as \lotus{} and \Lotus{}, respectively. A reference image without missing wedge can be found in Figure \ref{fig:gt_lotus}. Note, that the fanbeam geometry of the Mayo data is chosen such that it matches the specifications of the lotus scan. 
\end{enumerate}

\subsubsection{Operators} For an implementation of the discrete limited angle operator $\RadonLimD$ we use the standard \texttt{radon} routine of Matlab or a fanbeam geometry of the Astra toolbox \cite{astra}, which fits to the geometry described in \cite{bubba2016}.
For all our experiments we are using a discrete, bandlimited shearlet system  generated with the toolbox \cite{dedale}. The resulting system has $J=5$ scales, resulting in a transformation ${\shD \in \R^{59\cdot 512^2\times 512^2}}$, i.e., the shearlet coefficient cube $\shD (\f)$ has 59 subbands of size $512 \times 512$.

\subsubsection{Network Training}


Training \methodname{} is performed using Tensorflow \cite{abadi2016tensorflow} with an Adam optimizer \cite{kingma2014adam} and a learning rate (step size) of $10^{-4}$. In order to converge to a \emph{good} local minimizer of \eqref{eq:DLobjective}, i.e., one with a small generalization error, neural networks typically require a large number of training samples. Since the ellipsoid and Mayo data sets only consist out of $\sim 2000$ images, we make use of additional data augmentation techniques to help the network's convergence: for each tensor of size $512 \times 512 \times 59$, a random  $320 \times 320 \times 59$-patch is sampled on-the-fly and given as input to the fully convolutional \methodname{}. We find that such an on-the-fly sampling defines an effective regularization method. For the evaluation on the test set, the full array is given as input to the network. 

Since the shearlet coefficients naturally come in different orders of magnitude on each scale, we observe that weighting higher scales with larger weights in \eqref{eq:DLobjective}  (corresponding to small images features) significantly improves the neural network's performance, cf.\ the weighting strategy for the $\ell^1$-minimization in Appendix \ref{sec:solveAna}.

\subsubsection{Compared Methods}

We compare our reconstruction results with a variety of classical and learning based methods, which we describe briefly in the following list:

\begin{enumerate}[leftmargin=!,labelindent=30pt,itemindent=0pt]
    \item[$\fbp$:] Standard filtered backprojection with a 'shepp-logan' or 'ram-lak' filter, as provided by Matlab and Astra, respectively.
    \item[$\soluD$:] The $\ell^1$-regularized shearlet solution of \eqref{eq:ana_disc}.
    \item[$\tv$:] Total variation regularized solution with non-negativity constraint, i.e., a solution of \eqref{eq:ana_disc} where $\shD$ is replaced by a discrete gradient operator.
    \item[$\unser$:] Post-processing of $\fbp$ with \methodname{}, i.e., $\NNt$ is trained to remove artifacts in $\fbp$. Besides the different CNN architecture, this method resembles the one proposed in \cite{unser2017}.
    \item[$\jcyour$:] Post-processing of FBP's shearlet coefficients with \methodname{}, i.e., $\NNt$ is trained to remove artifacts in the coefficient domain. This method is closely related to the one proposed in \cite{jcy}.
    \item[$\jcysol$:] The actual method of \cite{jcy}.
\end{enumerate}

We wish to emphasize that all deep learning based comparison methods do not distinguish between visible and invisible information. Furthermore, residual learning is deployed, meaning that the networks learn the difference between input and output. 

\subsubsection{Similarity Measures}
For an assessment of image quality, we are using several quantitative measures, such as the relative error (RE)  given by
\begin{align*}
   \norm{\prop - \f }_2/\norm{\f}_2,
\end{align*}
where $\f$ denotes the reference image and $\prop$ its reconstruction.
Furthermore, we consider the peak signal-to-noise ratio (PSNR) and the structured similarity index (SSIM) \cite{wang2004} provided by Matlab. Finally, we are reporting the Haar wavelet-based perceptual similarity index (HaarPSI) that was recently proposed in \cite{haarpsi}.

\subsection{Results}

In the following, we will report and discuss the results of our numerical experiments.

\subsubsection{\ellip}

\begin{figure}
        \centering
        \small

        \begin{subfigure}[t]{0.3\textwidth}
    		\centering
    		\includegraphics[width=\textwidth]{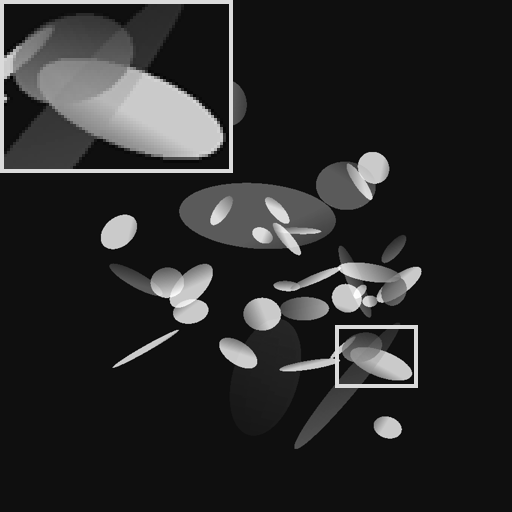}
    		\caption{\centering $\f$}
    		\label{fig:ellip:1}
	    \end{subfigure}%
	    \quad
	    \begin{subfigure}[t]{0.3\textwidth}
    		\centering
    		\includegraphics[width=\textwidth]{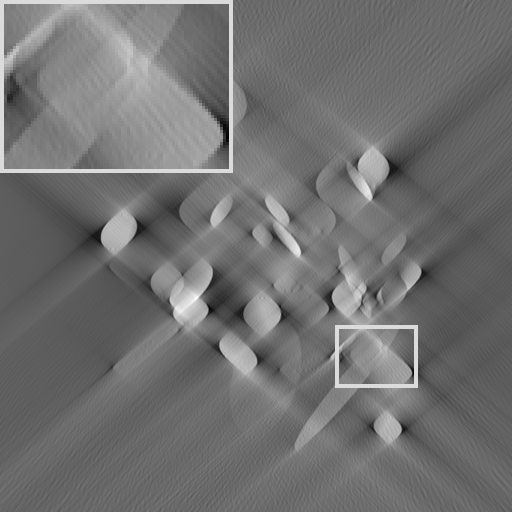}
    		\caption{\centering $\fbp$ \newline RE: 0.85, HaarPSI: 0.18}
    		\label{fig:ellip:2}
	    \end{subfigure}%
	    \quad
	    \begin{subfigure}[t]{0.3\textwidth}
    		\centering
    		\includegraphics[width=\textwidth]{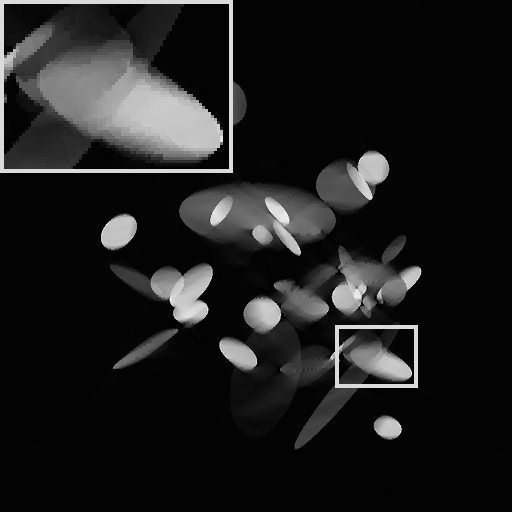}
    		\caption{\centering $\tv$ \newline RE: 0.29, HaarPSI: 0.42}
    		\label{fig:ellip:3}
	    \end{subfigure}%
	    \\
	     \begin{subfigure}[t]{0.3\textwidth}
    		\centering
    		\includegraphics[width=\textwidth]{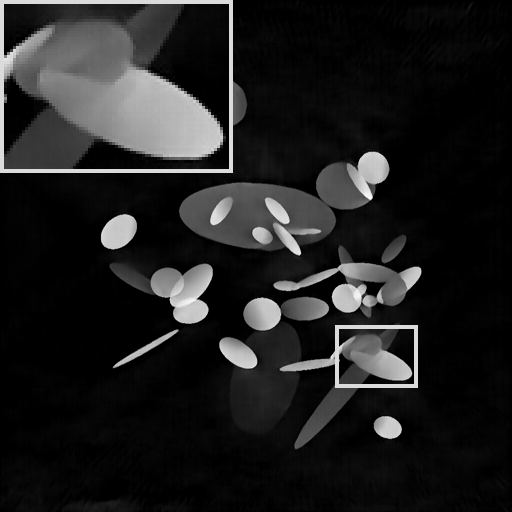}
    		\caption{\centering $\unser$ \newline RE: 0.19, HaarPSI: 0.74}
    		\label{fig:ellip:4}
	    \end{subfigure}%
	    \quad
	    \begin{subfigure}[t]{0.3\textwidth}
    		\centering
    		\includegraphics[width=\textwidth]{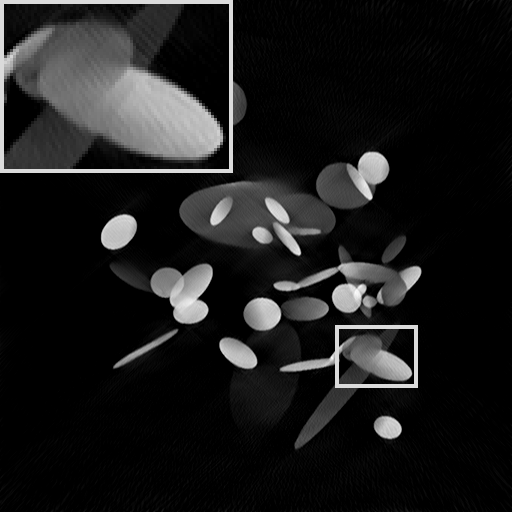}
    		\caption{\centering $\jcyour$ \newline RE: 0.22, HaarPSI: 0.67 }
    		\label{fig:ellip:5}
	    \end{subfigure}%
	    \quad
	    \begin{subfigure}[t]{0.3\textwidth}
    		\centering
    		\includegraphics[width=\textwidth]{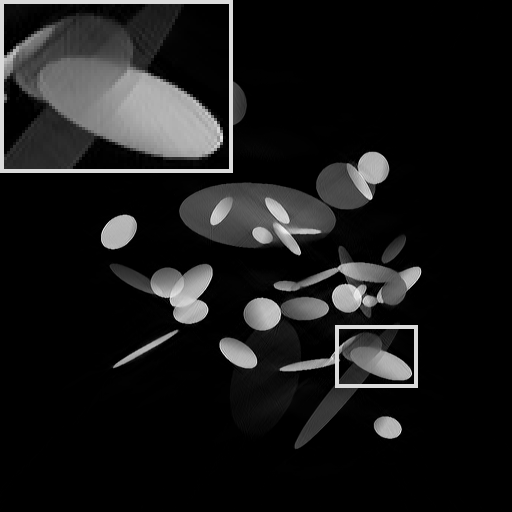}
    		\caption{\centering $\prop$ \newline RE: 0.11, HaarPSI: 0.81}
    		\label{fig:ellip:6}
	    \end{subfigure}%
	    \\
	    \begin{subfigure}[t]{0.3\textwidth}
    		\centering
    		\includegraphics[width=\textwidth]{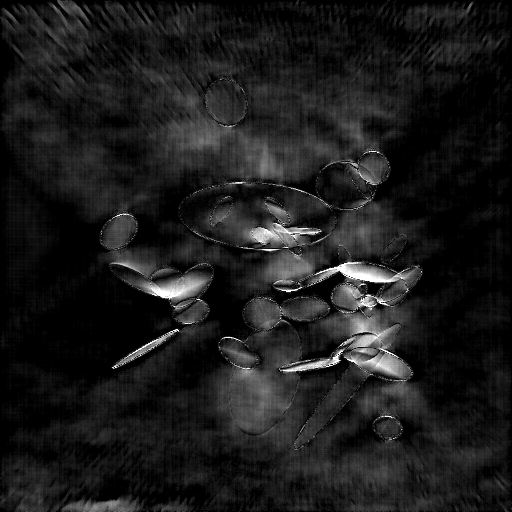}
    		\caption{$|\f - \unser|$}
    		\label{fig:ellip:7}
	    \end{subfigure}%
	    \quad
	    \begin{subfigure}[t]{0.3\textwidth}
    		\centering
    		\includegraphics[width=\textwidth]{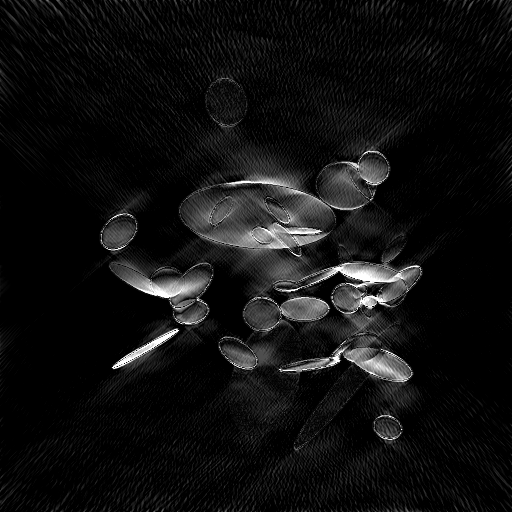}
    		\caption{$|\f - \jcyour|$}
    		\label{fig:ellip:8}
	    \end{subfigure}%
	    \quad
	    \begin{subfigure}[t]{0.3\textwidth}
    		\centering
    		\includegraphics[width=\textwidth]{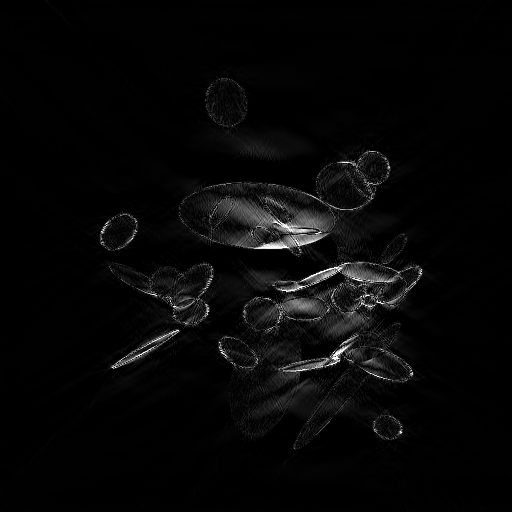}
    		\caption{$|\f - \prop|$}
    		\label{fig:ellip:9}
	    \end{subfigure}%
        \caption{Visualization of the results for one test image in \ellip. The last row shows the absolute value of the difference plots with respect to the ground truth image in the same plotting window. See Table \ref{table:ellipse} for averaged similarity measures over the test set.} 
        \label{fig:ellip}
    \end{figure}

The average image quality measures of the 200 test images are reported in Table \ref{table:ellipse}. Note that for the \ellip{} setup we did not compare with \cite{jcy}, since their networks have only been trained for a fanbeam geometry and for smaller missing wedges.
A visualization of the reconstruction quality for one of the test images is given in Figure \ref{fig:ellip}. Due to the large missing wedge of $80^\circ$, the FBP image in Figure \ref{fig:ellip:2} is heavily contaminated with streaking artifacts and contrast changes. Using $\ell^1$-minimization in Figure \ref{fig:ellip:3}, it is possible to reduce such artifacts significantly, however, the invisible boundaries are certainly not recoverable. The second row of Figure \ref{fig:ellip} reveals the effect of post-processing with CNNs: in all three methods, the invisible boundaries are well estimated by \methodname{}. Although the methods in \subref{fig:ellip:4} and \subref{fig:ellip:5} do a remarkable job, the zoomed parts of Fig \ref{fig:ellip:6} and the values in Table \ref{table:ellipse} reveal an advantage of our proposed framework. For better distinction of the learning based methods, the plots \subref{fig:ellip:7}-\subref{fig:ellip:9} show the absolute value of the difference images with respect to the ground truth. Subplot \subref{fig:ellip:7} and \subref{fig:ellip:8} reveal that $\jcyour$ and in particular $\unser$ mainly struggle with removing background fluctuations of $\fbp$. The error plot in Fig \ref{fig:ellip:9} shows that due to the $\ell^1$-minimization this issue is less prominent for our proposed solution, where mostly high frequency information of the invisible edges is missing.
For the sake of brevity we omitted plotting $\soluD$, which is very similar to $\tv$. However, since the class \ellip{} consists out of simple, almost piecewise constant images, $\tv$ is a very effective reconstruction method, even when compared CNN-based methods. A similar observation was made in \cite{unser2017} in the context of low-dose CT.

    \begin{table}
    \begin{center}
    \begin{tabular}{c|c|c|c|c}
    \hline \hline
     Method &  RE  & PSNR & SSIM & HaarPSI  \\
      \hline
    $\fbp$ & 0.84 & 17.16 & 0.12 & 0.18 \\
    $\soluD$ & 0.22 & 28.76 & 0.94 & 0.47 \\
    $\tv$ & 0.21 & 29.54 & 0.95 & 0.54 \\
    $\unser$ & 0.19 & 30.20 & 0.54 & 0.75 \\
    $\jcyour$ & 0.18 & 30.52 & 0.78 & 0.72 \\
    \hdashline
    $\prop$ & \textbf{0.09} & \textbf{36.96} & \textbf{0.96} & \textbf{0.86} \\
    \hline \hline
    \end{tabular}
    \end{center}
    \caption{Comparison of reconstruction methods for \ellip. The similarity values are averaged over the images in the test set. An example is displayed in Figure \ref{fig:ellip}.}
    \label{table:ellipse}
    \end{table}

\subsubsection{\mayo and \Mayo}

\begin{figure}
        \centering
        \small

        \begin{subfigure}[t]{0.3\textwidth}
    		\centering
    		\includegraphics[width=\textwidth]{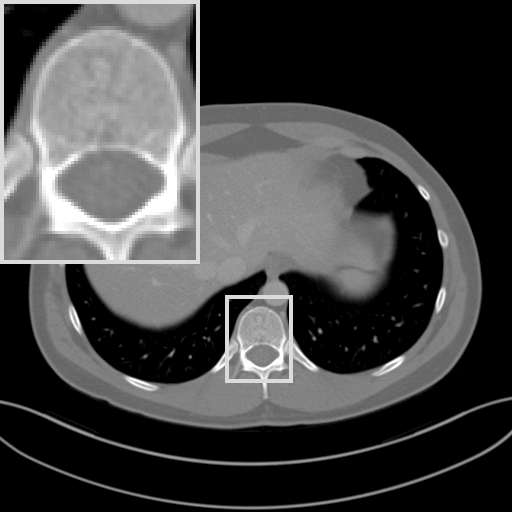}
    		\caption{\centering ground truth $\f$ }
    		\label{fig:mayo:1}
	    \end{subfigure}%
	    \quad
	    \begin{subfigure}[t]{0.3\textwidth}
    		\centering
    		\includegraphics[width=\textwidth]{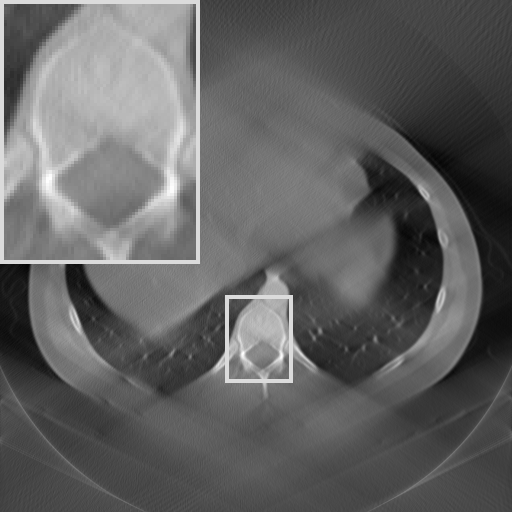}
    		\caption{\centering $\fbp$ \newline RE: 0.50, HaarPSI: 0.35}
    		\label{fig:mayo:2}
	    \end{subfigure}%
	    \quad
	     \begin{subfigure}[t]{0.3\textwidth}
    		\centering
    		\includegraphics[width=\textwidth]{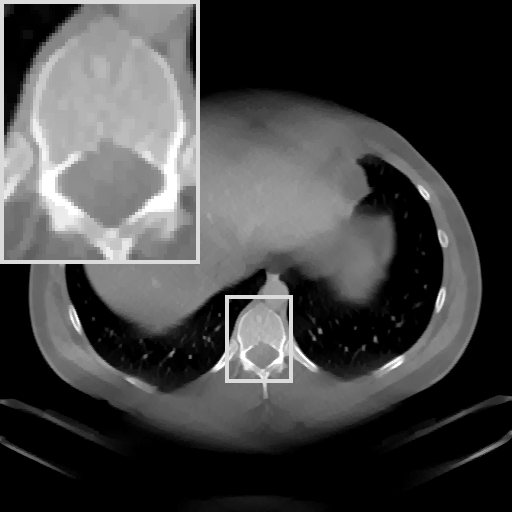}
    		\caption{\centering $\tv$ \newline RE: 0.21, HaarPSI: 0.41}
    		\label{fig:mayo:3}
	    \end{subfigure}%
	    \\
        \begin{subfigure}[t]{0.3\textwidth}
    		\centering
    		\includegraphics[width=\textwidth]{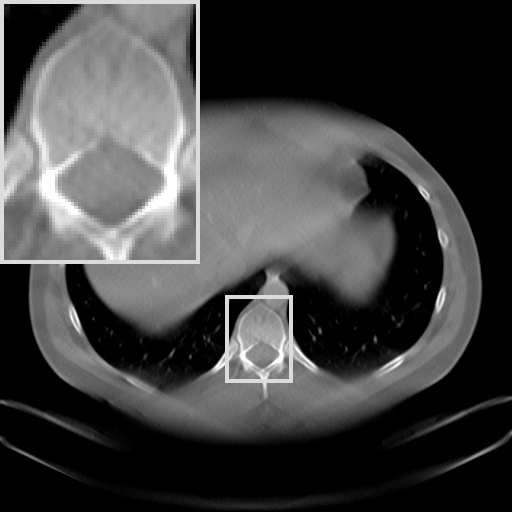}
    		\caption{\centering $\soluD$ \newline RE: 0.19, HaarPSI: 0.43}
    		\label{fig:mayo:4}
	    \end{subfigure}%
	    \quad
	    \begin{subfigure}[t]{0.3\textwidth}
    		\centering
    		\includegraphics[width=\textwidth]{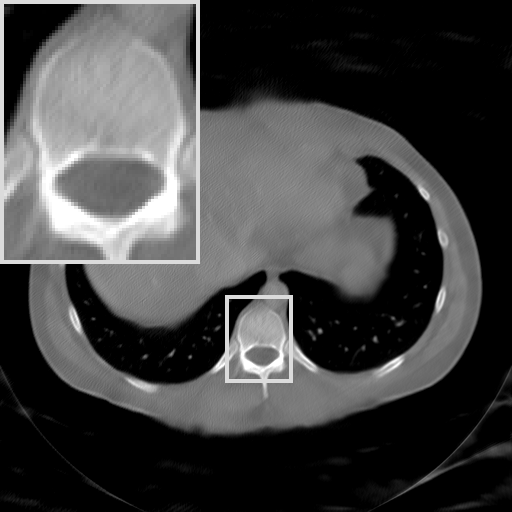}
    		\caption{\centering $\jcysol$ \newline RE: 0.22, HaarPSI: 0.40 }
    		\label{fig:mayo:5}
	    \end{subfigure}%
	    \quad
	    \begin{subfigure}[t]{0.3\textwidth}
    		\centering
    		\includegraphics[width=\textwidth]{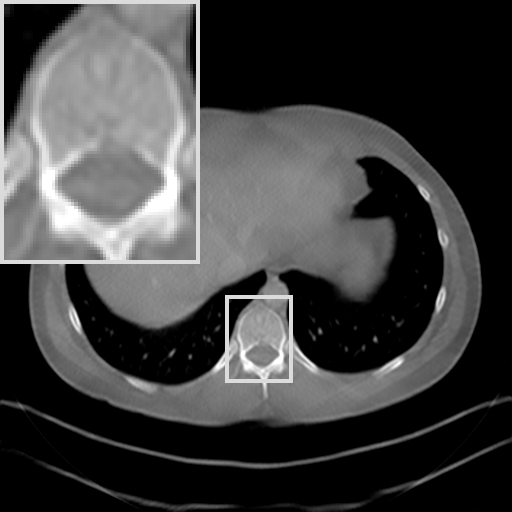}
    		\caption{\centering $\unser$ \newline RE: 0.16, HaarPSI: 0.53}
    		\label{fig:mayo:6}
	    \end{subfigure}%
	    \quad
	    \begin{subfigure}[t]{0.3\textwidth}
    		\centering
    		\includegraphics[width=\textwidth]{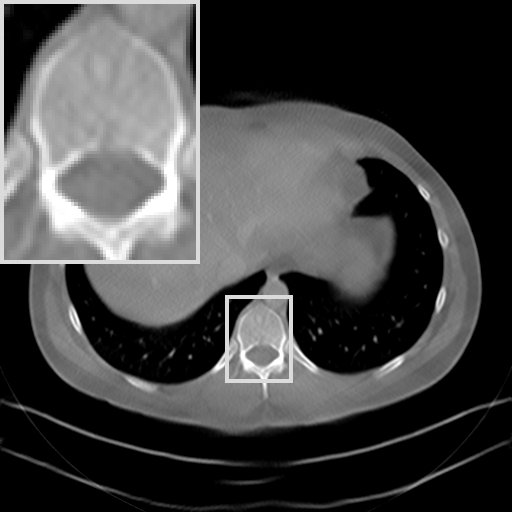}
    		\caption{\centering $\jcyour$ \newline RE: 0.16, HaarPSI: 0.58}
    		\label{fig:mayo:7}
	    \end{subfigure}%
	    \quad
	    \begin{subfigure}[t]{0.3\textwidth}
    		\centering
    		\includegraphics[width=\textwidth]{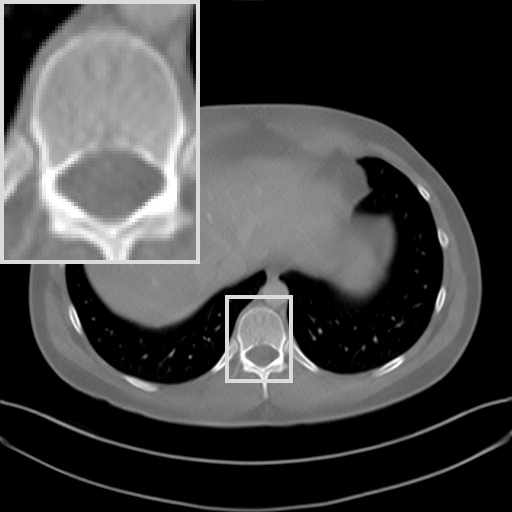}
    		\caption{\centering $\prop$ \newline RE: 0.09, HaarPSI: 0.76}
    		\label{fig:mayo:8}
	    \end{subfigure}%
        \caption{Comparison for one slice of the test patient of \mayo, i.e., coresponding to a missing wedge of $60^\circ$. The plotting window is slightly adapted for better contrast. See Table \ref{table:mayo} for averaged similarity measures across all slices of the test patient.}
        \label{fig:mayo}
    \end{figure}

\begin{figure}
        \centering
        \small

        \begin{subfigure}[t]{0.3\textwidth}
    		\centering
    		\includegraphics[width=\textwidth]{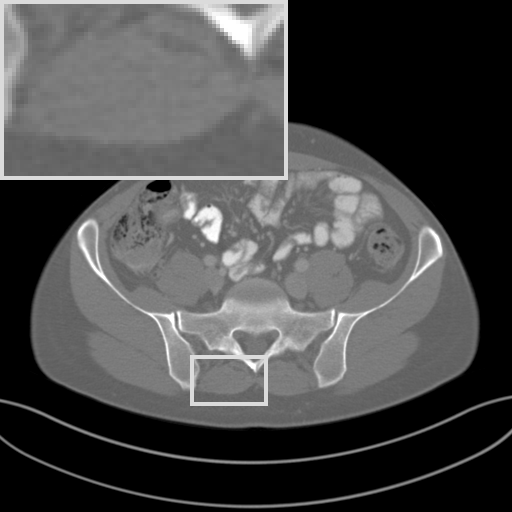}
    		\caption{\centering ground truth $\f$ }
    		\label{fig:Mayo:1}
	    \end{subfigure}%
	    \quad
	    \begin{subfigure}[t]{0.3\textwidth}
    		\centering
    		\includegraphics[width=\textwidth]{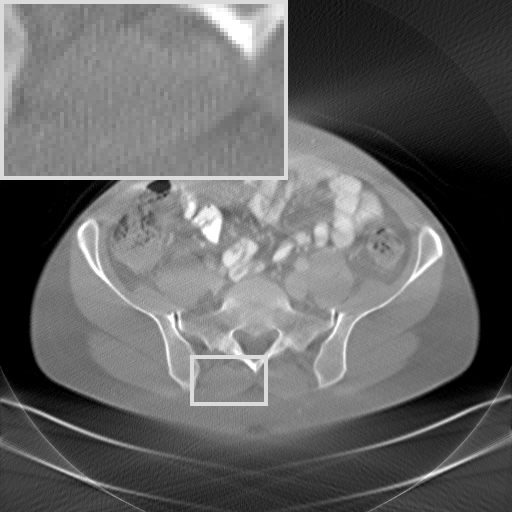}
    		\caption{\centering $\fbp$ \newline RE: 0.30, HaarPSI: 0.46}
    		\label{fig:Mayo:2}
	    \end{subfigure}%
	    \quad
	     \begin{subfigure}[t]{0.3\textwidth}
    		\centering
    		\includegraphics[width=\textwidth]{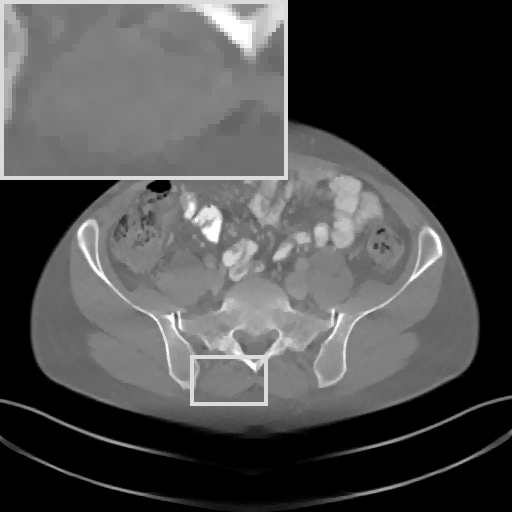}
    		\caption{\centering $\tv$ \newline RE: 0.10, HaarPSI: 0.63}
    		\label{fig:Mayo:3}
	    \end{subfigure}%
	    \\
	    \begin{subfigure}[t]{0.3\textwidth}
    		\centering
    		\includegraphics[width=\textwidth]{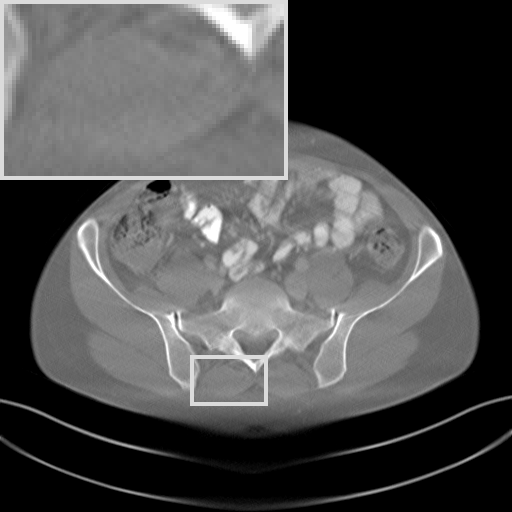}
    		\caption{\centering $\soluD$ \newline RE: 0.09, HaarPSI: 0.64}
    		\label{fig:Mayo:4}
	    \end{subfigure}%
	    \quad
	    \begin{subfigure}[t]{0.3\textwidth}
    		\centering
    		\includegraphics[width=\textwidth]{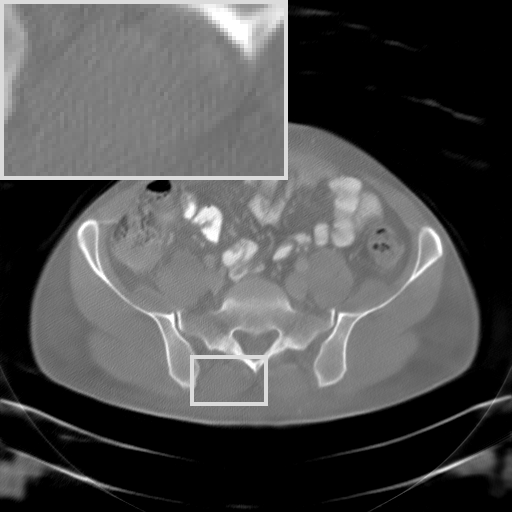}
    		\caption{\centering $\jcysol$ \newline RE: 0.21, HaarPSI: 0.43 }
    		\label{fig:Mayo:5}
	    \end{subfigure}%
	    \quad
	    \begin{subfigure}[t]{0.3\textwidth}
    		\centering
    		\includegraphics[width=\textwidth]{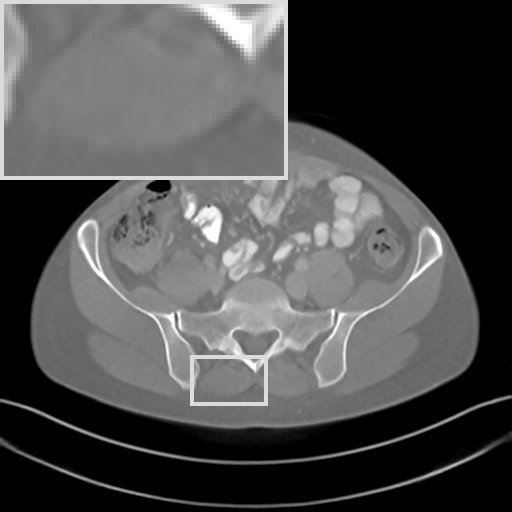}
    		\caption{\centering $\unser$ \newline RE: 0.09, HaarPSI: 0.82}
    		\label{fig:Mayo:6}
	    \end{subfigure}%
	    \quad
	    \begin{subfigure}[t]{0.3\textwidth}
    		\centering
    		\includegraphics[width=\textwidth]{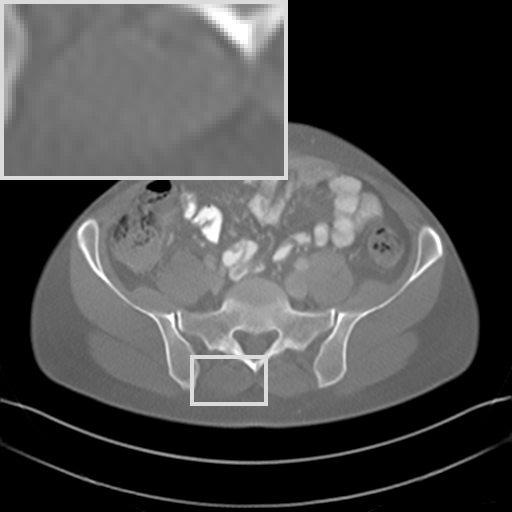}
    		\caption{\centering $\jcyour$ \newline RE: 0.06, HaarPSI: 0.82}
    		\label{fig:Mayo:7}
	    \end{subfigure}%
	    \quad
	    \begin{subfigure}[t]{0.3\textwidth}
    		\centering
    		\includegraphics[width=\textwidth]{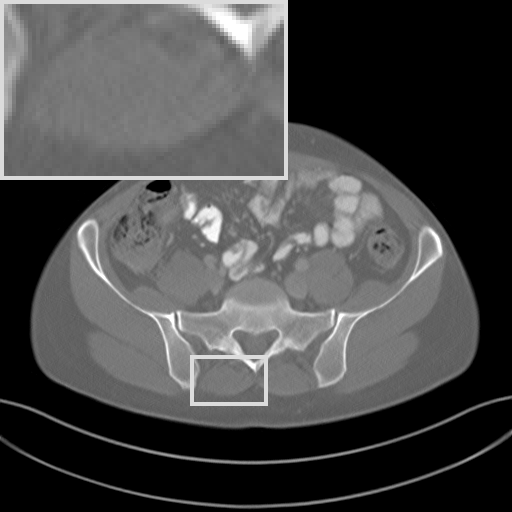}
    		\caption{\centering $\prop$ \newline RE: 0.03, HaarPSI: 0.92}
    		\label{fig:Mayo:8}
	    \end{subfigure}%
        \caption{Comparison for one slice of the test patient of \Mayo, i.e., coresponding to a missing wedge of $30^\circ$. The plotting window is slightly adapted for better contrast. See Table \ref{table:mayo} for averaged similarity measures across all slices of the test patient.}
        \label{fig:Mayo}
    \end{figure}

Table \ref{table:mayo} shows the average quality measures on the test patient of \mayo{} and \Mayo{}, respectively.
A visualization of two different slices of the test patient can be found in Figure \ref{fig:mayo} and Figure \ref{fig:Mayo}.

In the case of a missing wedge of $60^\circ$, the overall body shape shows clear differences for all considered methods. When compared to $\fbp$, the regularization methods in Figure \ref{fig:mayo:3} and \ref{fig:mayo:4} succeed in removing artifacts, however, the invisible boundaries are certainly not reconstructable. The learning-based methods displayed in the images \subref{fig:mayo:5}-\subref{fig:mayo:7} of Figure \ref{fig:mayo} estimate the invisible boundaries quite well, yet, there are unwanted fluctuations visible. Our proposed scheme reconstructs the invisible boundary almost perfectly. In particular, it is the only method that finds the correct round shape in the upper zoomed section.

    \begin{table}
    \begin{center}
    \begin{tabular}{c|c|c|c|c||c|c|c|c}
    & \multicolumn{4}{c||}{\mayo{}} & \multicolumn{4}{c}{\Mayo{}} \\
    \hline \hline
    Method &  RE  & PSNR & SSIM & HaarPSI  & RE  & PSNR & SSIM & HaarPSI \\
      \hline
    $\fbp$ & 0.47 & 17.16 & 0.40 & 0.32 & 0.31 & 21.23 & 0.48 & 0.46 \\
    $\tv$ & 0.18 & 25.88 & 0.85 & 0.37 & 0.10 & 30.99 & 0.85 & 0.64 \\
    $\soluD$ & 0.17 & 26.34 & 0.85 & 0.40 & 0.09 & 31.58 & 0.92 & 0.65 \\
    $\jcysol$ & 0.25 & 23.06 & 0.61 & 0.34 & 0.22 & 23.92 & 0.69 & 0.44 \\       
    $\unser$ & 0.15 & 27.40 & 0.78 & 0.52 & 0.10 & 31.22 & 0.81 & 0.81 \\
    $\jcyour$ & 0.16 & 26.80 & 0.74 & 0.52 & 0.06 & 35.190 & 0.90 & 0.82 \\
    \hdashline
    $\prop$ & \textbf{0.08} & \textbf{32.77} & \textbf{0.93} & \textbf{0.73} & \textbf{0.04} & \textbf{39.77} & \textbf{0.96} & \textbf{0.90} \\
    \hline \hline
    \end{tabular}
    \end{center}
    \caption{Comparison of reconstruction methods for \mayo and \Mayo. The values are averaged over all slices of the test patient. Examples are shown in Figure \ref{fig:mayo} and Figure \ref{fig:Mayo}, respectively.}
    \label{table:mayo}
    \end{table}

For a smaller missing wedge of $30^\circ$, Figure \ref{fig:Mayo} and Table \ref{table:mayo} (right-hand side) show that the differences between the compared methods are becoming less prominent. The learning-based methods of the images \subref{fig:Mayo:6}-\subref{fig:Mayo:8} of Figure \ref{fig:Mayo} are visually almost indistinguishable on mid- and large-scale features. However, for small details the zoomed part reveals a clear  advantage of our proposed method. While surprisingly $\unser$ has a small edge over $\jcyour$ on \mayo{}, here, it is indeed the other way round as reported in \cite{jcy}.  Finally, we remark that in this example, the staircasing effect of TV-regularization is visible on the zoomed part of Figure \ref{fig:Mayo:3} and avoided by relaying on shearlets as in Figure \ref{fig:Mayo:4}.

\subsubsection{\lotus and \Lotus}

In the experiments of this section, we evaluate the generalization performance by applying the  neural networks trained on the Mayo data to the scan of the lotus root. 
Let us explicitly point out that the neural networks have only been trained on the Mayo data and have \emph{not} been retrained on any other data set.  
Although there is a vague similarity between human abdomen scans and a sliced lotus root, this constitutes a difficult task for a neural network since it cannot rely on the strong correlations between scans of human patients.

\begin{figure}
        \centering
        \small

        \begin{subfigure}[t]{0.31\textwidth}
    		\centering
    		\includegraphics[width=\textwidth]{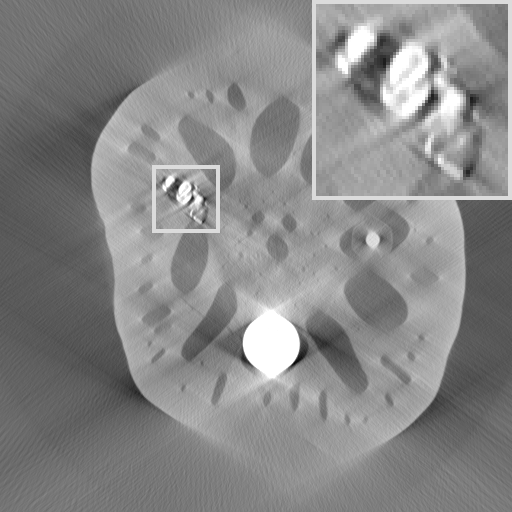}
    		\caption{\centering $\fbp$ \newline RE: 0.50, HaarPSI: 0.47}
    		\label{fig:lotus:1}
	    \end{subfigure}%
	    \quad
	    \begin{subfigure}[t]{0.31\textwidth}
    		\centering
    		\includegraphics[width=\textwidth]{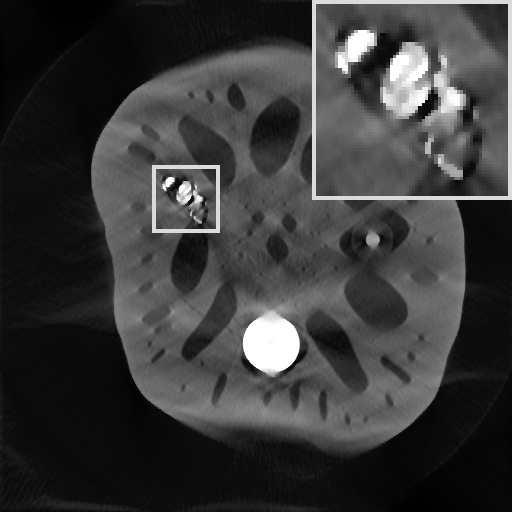}
    		\caption{\centering $\tv$ \newline RE: 0.21, HaarPSI: 0.57}
    		\label{fig:lotus:2}
	    \end{subfigure}%
	    \quad
	    \begin{subfigure}[t]{0.31\textwidth}
    		\centering
    		\includegraphics[width=\textwidth]{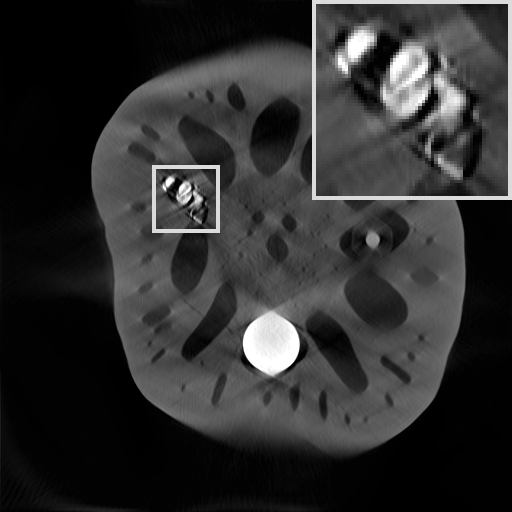}
    		\caption{\centering $\soluD$ \newline RE: 0.19, HaarPSI: 0.60}
    		\label{fig:lotus:3}
	    \end{subfigure}%
	    \\
	     \begin{subfigure}[t]{0.3\textwidth}
    		\centering
    		\includegraphics[width=\textwidth]{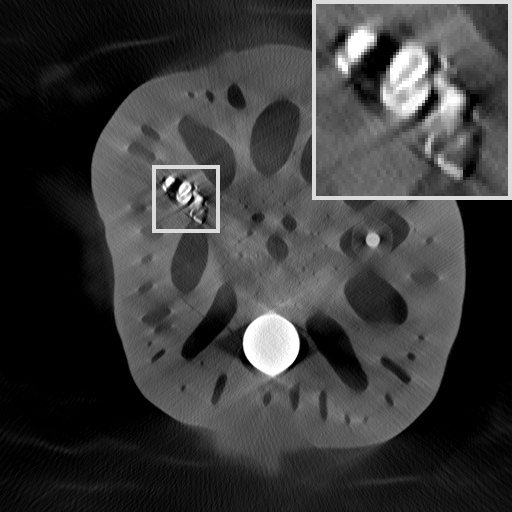}
    		\caption{\centering $\jcysol$ \newline RE: 0.43, HaarPSI: 0.54}
    		\label{fig:lotus:4}
	    \end{subfigure}%
	    \quad
	    \begin{subfigure}[t]{0.3\textwidth}
    		\centering
    		\includegraphics[width=\textwidth]{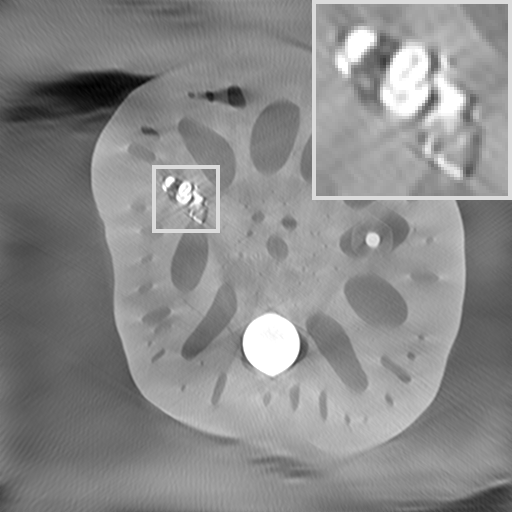}
    		\caption{\centering $\jcyour$ \newline RE: 0.55, HaarPSI: 0.46 }
    		\label{fig:lotus:5}
	    \end{subfigure}%
	    \quad
	    \begin{subfigure}[t]{0.3\textwidth}
    		\centering
    		\includegraphics[width=\textwidth]{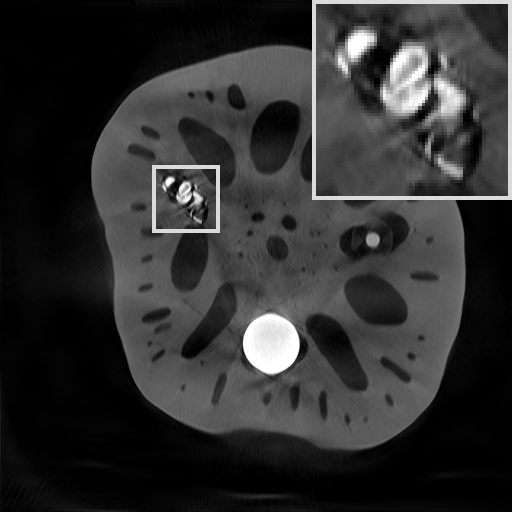}
    		\caption{\centering $\prop$ \newline RE: 0.17, HaarPSI: 0.70}
    		\label{fig:lotus:6}
	    \end{subfigure}%
        \caption{Evaluation of the generalization properties on \lotus. The neural networks have been trained on \mayo and are then tested on the \lotus measurements. A reference image can be found in Figure \ref{fig:gt_lotus}. The plotting window is slightly adapted for better contrast.}
        \label{fig:lotus}
    \end{figure}

In Figure \ref{fig:lotus}, we have visualized the reconstruction results for \lotus. Note that a reference scan can be found in Figure \ref{fig:gt_lotus} and that for the sake of brevity we have omitted the result of $\unser$, which was very similar to $\jcyour$. The model-based reconstruction methods of the first row are no surprise: non-linear $\ell^1$-regularization yields considerably better reconstructions than $\fbp$, with $\tv$ and $\soluD$ being of similar quality. The learning based methods of the images \subref{fig:lotus:4} and \subref{fig:lotus:5} reveal that both CNNs do not generalize well across different data sets. The network of $\jcysol$ suffers from contrast changes, does not remove more streaking artifacts than $\ell^1$-regularization and shows fluctuations at the invisible boundaries. The \methodname{} architecture of Figure \ref{fig:lotus:5} seems to cope even worse when working on $\fbp$. In contrast, our proposed scheme reconstructs the invisible outer shape of the lotus almost perfectly and improves on the streaking artifacts of $\soluD$. Such a remarkable generalization reveals the power of our method: relying on the well reconstructed visible coefficients of $\shD (\soluD)$ allows for a more precise estimation of the invisible ones. The reliable combination of the visible and inferred invisible coefficients ensures that the CNN $\NNt$ does not ``waste" its expressiveness on denoising the visible coefficients.

\begin{figure}
        \centering
        \small

        \begin{subfigure}[t]{0.31\textwidth}
    		\centering
    		\includegraphics[width=\textwidth]{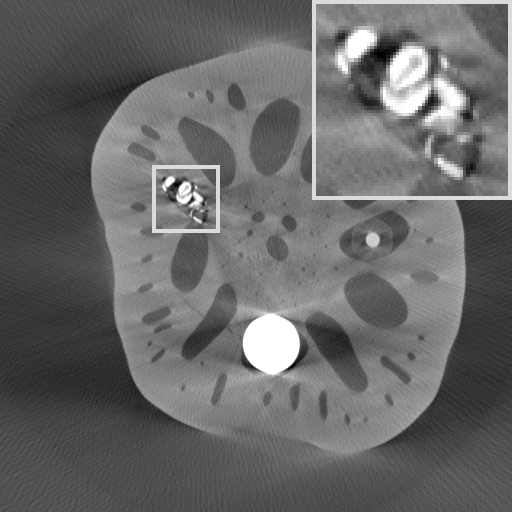}
    		\caption{\centering $\fbp$ \newline RE: 0.31, HaarPSI: 0.61}
    		\label{fig:Lotus:1}
	    \end{subfigure}%
	    \quad
	    \begin{subfigure}[t]{0.31\textwidth}
    		\centering
    		\includegraphics[width=\textwidth]{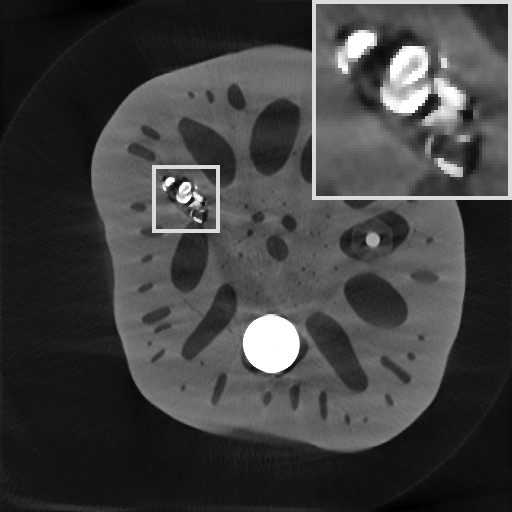}
    		\caption{\centering $\tv$ \newline RE: 0.12, HaarPSI: 0.74}
    		\label{fig:Lotus:2}
	    \end{subfigure}%
	    \quad
	    \begin{subfigure}[t]{0.31\textwidth}
    		\centering
    		\includegraphics[width=\textwidth]{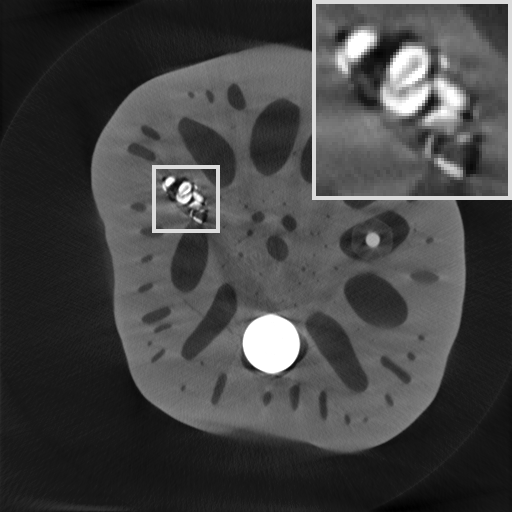}
    		\caption{\centering $\soluD$ \newline RE: 0.11, HaarPSI: 0.75}
    		\label{fig:Lotus:3}
	    \end{subfigure}%
	    \\
	     \begin{subfigure}[t]{0.3\textwidth}
    		\centering
    		\includegraphics[width=\textwidth]{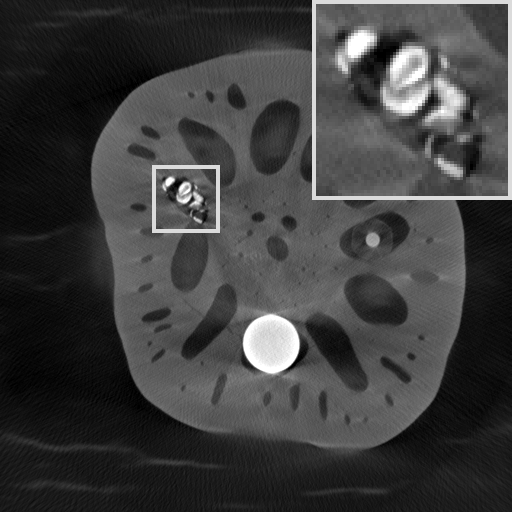}
    		\caption{\centering $\jcysol$ \newline RE: 0.25, HaarPSI: 0.62}
    		\label{fig:Lotus:4}
	    \end{subfigure}%
	    \quad
	    \begin{subfigure}[t]{0.3\textwidth}
    		\centering
    		\includegraphics[width=\textwidth]{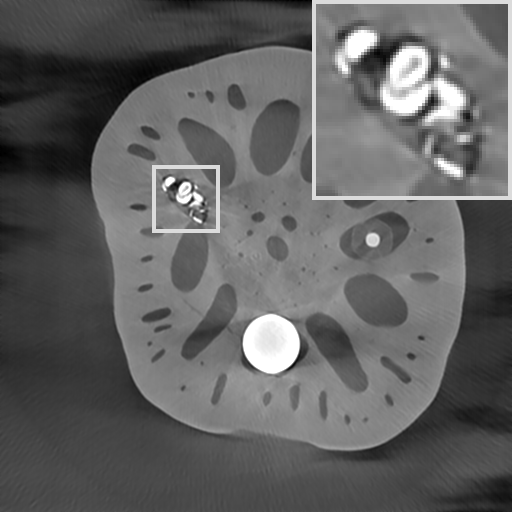}
    		\caption{\centering $\unser$ \newline RE: 0.32, HaarPSI: 0.66 }
    		\label{fig:Lotus:5}
	    \end{subfigure}%
	    \quad
	    \begin{subfigure}[t]{0.3\textwidth}
    		\centering
    		\includegraphics[width=\textwidth]{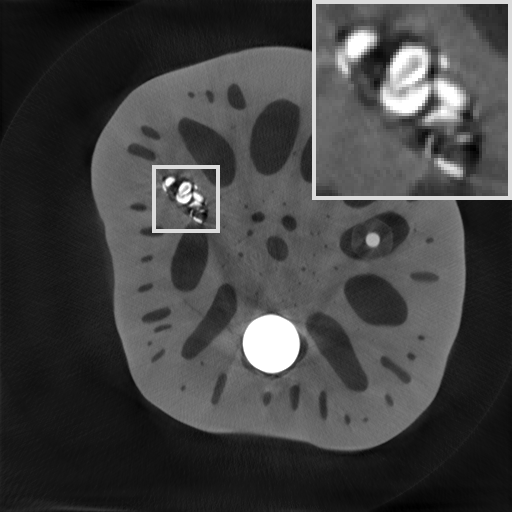}
    		\caption{\centering $\prop$ \newline RE: 0.11, HaarPSI: 0.83}
    		\label{fig:Lotus:6}
	    \end{subfigure}%
        \caption{Evaluation of the generalization properties on \Lotus. The neural networks have been trained on \mayo and are then tested on the \Lotus data.  A reference image can be found in Figure \ref{fig:gt_lotus}. The plotting window is slightly adapted for better contrast. }
        \label{fig:Lotus}
    \end{figure}

Similar, but less drastic observations hold true for the experiment \Lotus{}, displayed in Figure \ref{fig:Lotus}. The methods in \subref{fig:lotus:4} and \subref{fig:lotus:5} mostly succeed in estimating the invisible coefficients, however, there are still severe fluctuations visible. When comparing the result of Figure \ref{fig:Lotus:6} with the quality achieved on the test patient in  Figure \ref{fig:Mayo:6}, a decrease in performance is visible. Nonetheless, keeping in mind that $\sim 16\%$ of the measurements are missing and that $\NNt$ has never ``seen" a lotus root before, the reconstruction $\prop$ remains noteworthy.

\section{Conclusion}

In the present paper, we have introduced a reconstruction method for limited angle CT where the missing gaps in the wavefront set are closed by means of a deep neural network. We have shown that the visible boundary parts can be accessed by $\ell^1$-regularization with shearlets, a directional sensitive function  system that is proven to resolve wavefront sets. Based on such an accurate reconstruction of the visible shearlet coefficients, we have trained a U-Net like architecture \methodname{} for an estimation of the unknown, invisible coefficients. 

The close coupling of an advanced model-based method and a custom-tailored learning task implies an increased reliability of the final results and a better understanding of the post-processing capabilities of deep neural networks in the context of limited angle CT (cf.~the discussion in Section \ref{sec:disc}). We have furthermore shown superior reconstruction quality when compared with classical methods and less model-oriented deep learning algorithms. 

We hope that our framework for limited angle CT will spark further interest in applying similar ideas to other inverse problems. We anticipate that this might be particularly fruitful for problems, where larger amounts of data are not acquired in the measurements, as it is for instance the case in region of interest and exterior tomography \cite{Borg2018}. Furthermore, we have observed that we could improve the learning phase by optimizing over a more sophisticated loss function defined over the image domain or by adding additional regularizers. It might be worthwhile to further pursue this line of thought in the future. 

\section*{Acknowledgments}
 T.A.B., M.L. and S.S. acknowledge support by the Academy of Finland through the Finnish Centre of Excellence in Inverse Modelling and Imaging 2018-2025, decision number 312339, and Academy Project 310822.
 G.K.\ acknowledges partial support by the Bundesministerium f\"ur Bildung und Forschung (BMBF) through the Berliner Zentrum for Machine Learning (BZML), Project AP4, by the Deutsche Forschungsgemeinschaft (DFG) through grants CRC 1114 ``Scaling Cascades in Complex Systems", Project B07, CRC/TR 109 ``Discretization in Geometry and Dynamics', Projects C02 and C03, RTG DAEDALUS (RTG 2433), Projects P1 and P3, RTG BIOQIC (RTG 2260), Projects P4 and P9, and SPP 1798 ``Compressed Sensing in Information Processing", Coordination Project and Project Massive MIMO-I/II, by the
Einstein Foundation Berlin, and by the Einstein Center for Mathematics Berlin (ECMath), Project CH14.
M.M. acknowledges support by the DFG through the SPP 1798 ``Compressed Sensing in Information Processing" Coordination Project.
W.S. and V.S. acknowledge partial support by the Bundesministerium f\"ur Bildung und Forschung (BMBF) through the Berlin Big Data Center under Grant 01IS14013A and the Berlin Center for Machine Learning under Grant 01IS180371, as well as support by the Fraunhofer Society through the MPI-FhG collaboration project ``Theory \& Practice for Reduced Learning Machines''.
All authors  would like to acknowledge  Dr.   Cynthia   McCollough,  the
Mayo  Clinic,  and  the  American  Association  of  Physicists  in
Medicine as well as the  grants  EB017095
and  EB017185  from  the  National  Institute  of  Biomedical
Imaging and Bioengineering for providing the AAPM Low-Dose Grand Challenge data. Furthermore, we wish to warmly thank Jong Chul Ye for providing us with an implementation of his reconstruction method. 


\bibliographystyle{amsplain}

\bibliography{references}
\bibliographystyle{abbrv}

\begin{appendix}
\section{Solving the Analysis Formulation}
\label{sec:solveAna}

We solve the minimization problem \eqref{eq:ana_disc} by applying the alternating direction method of multipliers (ADMM) \cite{douglas1956,boyd2011}. First, we rewrite \eqref{eq:ana_disc} in the equivalent form
\begin{equation*}
    \min_{\vec{f},\vec{z}} \rho_0 \cdot \left( \frac{1}{2} \norm{\RadonLimD \f - \meas}_2^2   + \norm{ \Pi_1 \vec{z}}_{1,\vec{w}} + \iota_{\geq 0} (\Pi_2 \vec{z}) \right) \quad \mbox{ s.t. } \quad \vec{A} \f + \vec{B} \vec{z} = \vec{0},
\end{equation*}
where $\rho_0> 0$, $\vec{A}^T = (\rho_1 \shD^T, \rho_2 \vec{I}_{n^2}) \in \R^{n^2 \times (J+1)n^2}$, $\vec{B} = \diag (-\rho_1 \mathds{1}_{Jn^2}, -\rho_2 \mathds{1}_{n^2}) \in  \R^{(J+1)n^2 \times (J+1)n^2}$ for $\rho_1,\rho_2>0$, $\mathds{1}_k \in \R^k$ is the vector with all components being $1$, and $\Pi_1, \Pi_2$ denote the projections onto the first $J n^2$ and the last $n^2$ entries, respectively. Introducing $\rho_0$ might seem superfluous, but it serves as a conditioning parameter later on. Then, the scaled form of \cite{boyd2011} for $F(\f) = \rho_0/2 \norm{\RadonLim \f -\meas}_2^2$ and $G(\vec{z}) = \rho_0( \norm{ \Pi_1 \vec{z}}_{1,\vec{w}} + \iota_{\geq 0} (\Pi_2 \vec{z}))$ results in the following iterates
\begin{align}
    \f^{k+1} & := \argmin_{\f} \left(F(\f) + \rho/2 \norm{\vec{A}\f + \vec{B} \vec{z}^k +\vec{u}^k}_2^2 \right) \label{eq:fup}\\
    \vec{z}^{k+1} & := \argmin_{\vec{z}} \left(G(\vec{z}) + \rho/2 \norm{\vec{A} \f^{k+1} + \vec{B} \vec{z} + \vec{u}^{k}}_2^2  \right) \label{eq:zup}\\
    \vec{u}^{k+1} & := \vec{u}^k + \vec{A}\f^{k+1} + \vec{B} \vec{z}^{k+1}, \nonumber
\end{align}
where $\rho>0$.
Step \eqref{eq:fup} results in solving the linear system
\begin{multline*}
    \left(\rho_0 \RadonLimD^T \RadonLimD + \rho \rho_1^2 \shD^T \shD + \rho \rho_2^2 \vec{I}_{n^2} \right) \f \\ = \rho_0 \RadonLimD^T \meas +  \rho \rho_1^2 \shD^T(\Pi_1 \vec{z}^k - \Pi_1 \vec{u}^k/\rho_1) +  \rho \rho_2^2 (\Pi_2 \vec{z}^k - \Pi_2 \vec{u}^k/\rho_2).
\end{multline*}
The proximal step of \eqref{eq:zup} decouples into the following two seperate operations
\begin{equation*}
    \Pi_1 \vec{z}^{k+1} = \mbox{shrink}\left(\shD (\f^{k+1}) + \Pi_1 \vec{u}^k/\rho_1,\frac{\rho_0 \vec{w}}{\rho \rho_1^2}\right)
\end{equation*}
and
\begin{equation*}
    \Pi_2 \vec{z}^{k+1} = \max (\f^{k+1} + \Pi_2 \vec{u}^k/\rho_2 ,\vec{0}),
\end{equation*}
where \emph{shrink} denotes to element-wise \emph{soft-thresholding}, i.e., for $\vec{a} \in \R^n$ and $\vec{b}\in \R^n_+$ we set
\begin{equation*}
    \left( \mbox{shrink} (\vec{a},\vec{b}) \right)_i = \begin{cases} \max (|a_i| - b_i,0) \frac{a_i}{|a_i|}, & \mbox{ if } a_i\neq 0 \\ 0, & \mbox{ else. } \end{cases}
\end{equation*}
After substituting $\rho \rho_i^2$  by $\rho_i$, and $\Pi_i \vec{u}^{k}/\rho_i$ by $\Pi_i \vec{u}^{k}$ and using that $\shD^T \shD = \vec{I}_{n^2}$, we obtain the overall algorithm
\begin{align}
     \f^{k+1} & := \left(\rho_0 \RadonLimD^T \RadonLimD + (\rho_1 + \rho_2) \vec{I}_{n^2} \right)^{-1} \label{eq:lin_sys}\\&   \left(\rho_0 \RadonLimD^T \meas +  \rho_1 \shD^T(\Pi_1 \vec{z}^k - \Pi_1 \vec{u}^k) +   \rho_2 (\Pi_2 \vec{z}^k - \Pi_2 \vec{u}^k) \right)  \nonumber  \\
     \Pi_1 \vec{z}^{k+1} & := \mbox{shrink}\left(\shD (\f^{k+1}) + \Pi_1 \vec{u}^k,\frac{\rho_0 \vec{w}}{\rho_1}\right)  \nonumber \\
     \Pi_2 \vec{z}^{k+1} & := \max (\f^{k+1} + \Pi_2 \vec{u}^k ,\vec{0}) \nonumber \\
     \Pi_1 \vec{u}^{k+1} & := \Pi_1 \vec{u}^k + \shD (\f^{k+1}) - \Pi_1 \vec{z}^{k+1} \nonumber \\
     \Pi_2 \vec{u}^{k+1} & := \Pi_2 \vec{u}^k + \f^{k+1} - \Pi_2 \vec{z}^{k+1}.\nonumber
\end{align}
The overparametrization by $\rho_i$ is used to balance out data fidelity and regularization for solving the linear system \eqref{eq:lin_sys}. Although the algorithm converges independently of the choice of these parameters, setting them properly may lead to significant speed-up of convergence. Note, that it is not necessary to solve \eqref{eq:lin_sys} up to full precision \cite{boyd2011}. In all our experiments, we are using the conjugate gradient method with the previous iterate $\vec{f}^k$ as a warm start for finding an approximate solution. The ADMM is known to converge to modest precision within a few tens of iterations, which is usually enough for imaging purposes \cite{boyd2011}. The speed of the overall algorithm is mostly dominated by cost of applying $\RadonLimD^T \RadonLimD$  for solving \eqref{eq:lin_sys}.

In all our experiments, we fix $\rho_2 = 1$, such that only $\rho_0,\rho_1$ and the weight $\vec{w}$ are left as hyperparameters. We initialize the algorithm with $\f^0:= \RadonLimD^T \meas$, $\vec{z}^0 := \vec{0}$ and $\vec{u}^{0}:= \vec{0}$ and stop after 50 iterations. For the different experiments we are choosing the following parameter setup, which were found by manual tuning:
\begin{itemize}
\item $\rho_0 = 0.02, \rho_1 = 0.1$ and $w_j  =  3^j/400$, (\ellip)
\item $\rho_0= 0.50, \rho_1 = 0.1$ and $w_j = 2^j/400$, (\mayo)
\item $\rho_0 = 0.08, \rho_1 = 0.5$ and $w_j = 2^j/72$, (\Mayo + \Lotus)
\item $\rho_0 = 0.01, \rho_1 = 0.1$ and $w_j = 2^j/40$, (\lotus)
\end{itemize}
where $j$ denotes the $j$-th shearlet scale. A reconstruction of a $512 \times 512$ image with a missing wedge of $60^\circ$ takes about 2.5 minutes using an Intel i7 processor and 16GB RAM.

\end{appendix}


\end{document}